\documentclass[lettersize,journal]{IEEEtran}
\usepackage{amsfonts, amsmath, amssymb, amsthm}
\usepackage{algorithmic}
\usepackage{algorithm}
\usepackage{array}
\usepackage{booktabs}
\usepackage{threeparttable}
\usepackage{textcomp}
\usepackage{stfloats}
\usepackage{url}
\usepackage{verbatim}
\usepackage{graphicx}
\usepackage{tikz}
\usepackage{subcaption}
\usepackage{cite}
\usepackage{cancel}
\usepackage{mathtools}
\usepackage{marvosym}
\usepackage{circledsteps}
\usepackage{multirow}
\usepackage{threeparttable}
\theoremstyle{remark}

\theoremstyle{plain}

\begin{document}
	
\title{Liquid Gated Attention}

\author{Yiheng Jiang, Yuanbo Xu, and Yongjian Yang%
	\thanks{All authors are with the Mobile Intelligent Computing Laboratory,
	College of Computer Science and Technology, Jilin University, China
	(e-mail: jiangyh22@mails.jlu.edu.cn; yuanbox@jlu.edu.cn;
	yyj@jlu.edu.cn).}%
	\thanks{Corresponding author: Yuanbo Xu.}}
	
	\markboth{IEEE TRANSACTIONS ON PATTERN ANALYSIS AND MACHINE INTELLIGENCE,~Vol.~XX, No.~XX, Month~202X}{}
	
	
\maketitle
	
\begin{abstract}
Real-world time series often exhibit irregular sampling and extended temporal horizons, requiring models to capture continuous-time dynamics across arbitrary time intervals without prohibitive scaling costs. Conventional discrete-time methods typically project observations onto uniformly spaced indices, collapsing variable time intervals into static positional steps and discarding temporal distances between observations. Conversely, solver-dependent continuous-time models preserve temporal structure but rely on sequential numerical integration, precluding parallelization. While solver-free approximations bypass this computational overhead, no parallel architecture explicitly couples observed time intervals with input-driven state modulation, limiting their ability to distinguish genuine temporal dynamics from observational noise. To bridge this gap, we propose Liquid Gated Attention (LGA), a solver-free parallel temporal operator. By parameterizing an input-driven gating mechanism with observed time intervals, LGA introduces a continuous-time inductive bias and formulates hidden state evolution as a fast-weight associative memory, enabling parallel computation across the temporal dimension. By leveraging matrix associativity in non‑causal encoding and a prefix scan in causal encoding, LGA achieves linear temporal complexity with respect to sequence length in both modes. To stabilize long-horizon optimization, we employ a sequence-level normalization that bounds the cumulative temporal decay. Building on this operator, we instantiate LFormer, a modular backbone for continuous-time representation learning. Evaluations across six tasks and sixteen datasets, spanning up to 17,984 steps, demonstrate LFormer's capabilities in long-range dependency modeling, fine-grained state tracking, and trajectory reconstruction from sparse and noisy observations. LFormer delivers competitive performance against state-of-the-art discrete-time and continuous-time baselines, alongside linear scaling efficiency confirmed by empirical benchmarks. These results illustrate how LGA embeds continuous-time dynamical principles into a parallel architecture, positioning LFormer as a scalable and versatile backbone for time series representation learning. Code is available at \url{https://anonymous.4open.science/r/Liquid-Gated-Attention-6B55}.
\end{abstract}
	
\begin{IEEEkeywords}
	Continuous-time Model, Linear Attention, Time Series Analysis, Numerical Stability
\end{IEEEkeywords}

\section{Introduction \label{sec: introduction}}
\IEEEPARstart{M}{onitoring} real-world processes, such as physiological states~\cite{DBLP:journals/pami/QuinnWM09}, financial fluctuations~\cite{DBLP:journals/pami/LiCFHZ24}, and biological signaling~\cite{DBLP:journals/pami/ElnaggarHDRWJGF22}, typically yields complex time series that exhibit irregular sampling and can span extended temporal horizons. These properties present significant challenges for temporal modeling. Specifically, irregular sampling requires models to characterize continuous-time dynamics across arbitrary observed time intervals~\cite{DBLP:journals/corr/abs-2012-00168, DBLP:conf/nips/RubanovaCD19}. Simultaneously, extended sequence lengths demand capturing long-range dependencies without prohibitive computational costs~\cite{DBLP:journals/corr/abs-2312-00752, DBLP:conf/aaai/ZhouZPZLXZ21}. Furthermore, realistic observational constraints call for algorithms that extract the underlying continuous-time process from sparse and noisy measurements~\cite{kidger2022neuraldifferentialequations, DBLP:conf/icml/SchirmerELR22}.
	
Most existing approaches operate within the discrete-time paradigm that projects continuous temporal trajectories onto uniformly spaced indices~\cite{DBLP:conf/nips/ChenRWFSL23}. Despite excelling at handling regularly sampled data, this paradigm collapses variable intervals into static positional steps and discards the temporal distances between observations, which are necessary for accurate continuous-time dynamics modeling~\cite{DBLP:journals/corr/abs-2012-00168, DBLP:conf/nips/RubanovaCD19, DBLP:conf/nips/ChenRWFSL23}. Recurrent neural networks (RNNs)~\cite{DBLP:conf/emnlp/ChoMGBBSB14, DBLP:journals/neco/HochreiterS97} and Transformers~\cite{DBLP:conf/nips/VaswaniSPUJGKP17} represent two dominant discrete-time models. Standard RNNs compress temporal dependencies into sequential hidden states. To accommodate irregular observations, they typically rely on heuristic preprocessing strategies, such as binning~\cite{DBLP:conf/mlhc/LiptonKW16} and imputation~\cite{DBLP:conf/nips/SmiejaST0S18}, which inject spurious states into the timeline and distorts the underlying continuous dynamics~\cite{DBLP:conf/nips/RubanovaCD19}. Moreover, their recurrence precludes parallelization along the temporal dimension. Transformers eliminate this sequential constraint through parallel self-attention. Yet, even when augmented with explicit temporal encoding~\cite{DBLP:conf/icde/WangJXWY22, DBLP:journals/tkde/JiangYXW24, xu2019self}, they collapse time intervals into static similarity weights between discrete tokens, lacking a dedicated mathematical mechanism to model continuous-time dynamics between observations~\cite{DBLP:conf/nips/ChenRWFSL23}. Furthermore, their quadratic complexity limits scalability to extended temporal horizons.
	
To preserve temporal structure, continuous-time approaches formulate the state evolution as a differential equation parameterized by a neural network. Neural ordinary differential equations pioneered this formulation~\cite{DBLP:conf/nips/ChenRBD18}. Subsequent extensions, including neural controlled and rough differential equations, accommodate irregularly sampled time series through continuous path integration~\cite{DBLP:conf/nips/KidgerMFL20, DBLP:conf/icml/MorrillSKF21}. Rooted in differential calculus, these methods explicitly incorporate temporal distances between observations~\cite{DBLP:conf/nips/RubanovaCD19} and can naturally smooth high-frequency noise~\cite{kidger2022neuraldifferentialequations}. However, their forward pass relies on numerical solvers for sequential integration. This solver-dependent design introduces two bottlenecks. First, numerical solvers require multiple function evaluations per time step, and their sequential execution resists parallelization~\cite{DBLP:journals/natmi/HasaniLALRTTR22}. This computational overhead challenges these methods in large-scale or real-time deployments. Second, truncation errors can accumulate over extended integration horizons~\cite{kuleshov2026denots}, progressively degrading the fidelity of learned dynamics.
	
These bottlenecks motivate solver-free alternatives that retain continuous-time dynamics while bypassing iterative integration. To render continuous state evolution directly evaluable, three principal strategies have emerged: closed-form approximations~\cite{DBLP:journals/natmi/HasaniLALRTTR22}, linear state-space formulations~\cite{DBLP:conf/iclr/GuGR22}, and truncated algebraic path signatures~\cite{10.5555/3737916.3741287}. Despite eliminating the solver-induced computational overhead, these methods commit continuous-time dynamics to fixed mathematical forms. These compromises introduce specific trade-offs documented in their respective literature, including potential gradient decay in closed-form recurrence~\cite{DBLP:journals/natmi/HasaniLALRTTR22}, constrained expressivity for nonlinear dynamics in linear state transitions~\cite{DBLP:journals/corr/abs-2312-00752}, and information loss from truncated signatures~\cite{10.5555/3737916.3741287}. Furthermore, within this paradigm, no parallel architecture explicitly couples observed time intervals with input-driven state modulation. Without this linkage, they struggle to distinguish noise from genuine state transitions~\cite{DBLP:conf/nips/RubanovaCD19, DBLP:conf/icml/SchirmerELR22}, which limits their robustness under sparse and noisy conditions.

Across these three paradigms, a critical gap emerges: no single approach simultaneously incorporates observed time intervals, enables parallel computation, and maintains observational noise robustness. Discrete-time models discard the underlying temporal metric. Solver-dependent continuous-time methods retain temporal fidelity and noise robustness but sacrifice parallelism. Solver-free approaches recover parallel scalability yet lack an explicit temporal-state transition link, compromising robustness to sparse and noisy measurements.

In this work, we address this trilemma via \textbf{L}iquid \textbf{G}ated \textbf{A}ttention (LGA), a solver-free temporal operator reconciling observed time intervals, parallel computation, and observational noise robustness across four design steps:
\begin{itemize}
	\item[\textbf{(1)}] \textbf{Continuous-time Gating.} LGA derives its gating mechanism from the closed-form structure of a one-dimensional liquid time-constant (1D LTC) equation~\cite{DBLP:conf/aaai/HasaniLARG21}. The resulting \textbf{liquid gate} jointly encodes temporal decay and input-driven modulation, providing a continuous-time inductive bias that attenuates high-frequency noise perturbations.
	
	\item[\textbf{(2)}] \textbf{Computational Efficiency.} To avoid the computational overhead of numerical integration, we approximate the liquid gate via a learnable endpoint interpolation inspired by the trapezoidal rule. Fixing the interpolation weight at $0.5$ recovers the classical trapezoidal rule with second-order local accuracy, whereas the generalized learnable configuration serves as a trainable numerical surrogate. This step parallelizes the gating calculation while preserving the continuous-time inductive bias.
	
	\item[\textbf{(3)}] \textbf{Expressive Capability.} To capture complex long-range dependencies, we elevate the scalar state in 1D LTCs to a matrix-valued associative memory. Adopting a fast-weight programming framework~\cite{DBLP:journals/neco/Schmidhuber92}, we translate the liquid gate into explicit modulation coefficients for attention features. This design preserves the continuous-time gating dynamics while yielding a parallel sequence operator that maintains linear temporal complexity with respect to sequence length in both non-causal and causal modes, the latter realized via a prefix scan.
	
	\item[\textbf{(4)}] \textbf{Numerical Safety.} To stabilize long-horizon optimization, we introduce a sequence-level normalization that bounds the cumulative decay coefficients, preventing exponential underflow. This normalization operates as a regularized surrogate rather than an algebraically exact equivalent of the original LTC decay: it bounds the total decay budget while preserving the relative temporal distance allocation across the sequence.
\end{itemize}
	
Building upon the LGA operator, we instantiate \textbf{LFormer}, a modular backbone for continuous-time representation learning. LFormer constructs its deep temporal representations by interleaving decoupled multi-head LGA layers with channel-mixing blocks, connected via residual connections with layer normalization. LFormer directly processes discrete observations, while inheriting the continuous-time inductive bias, noise robustness, linear temporal complexity, and training stability properties from the underlying LGA operator.

This work delivers three primary contributions:
\begin{itemize}
	\item \textbf{A parallel continuous-time temporal operator.} LGA unifies the input-driven and time-interval-coupled liquid gate, a trapezoidal-rule-inspired numerical integral surrogate, fast-weight associative memory, and sequence-level normalization into a single solver-free formulation.
	
	\item \textbf{A modular backbone for continuous-time representation learning.} LFormer instantiates LGA in a decoupled multi-head form within a residual architecture. This design enables linear-complexity parallel sequence modeling directly over discrete observations, while inheriting LGA's continuous-time inductive bias, noise robustness, and training stability properties.
	
	\item \textbf{A comprehensive empirical validation across diverse temporal scenarios.} Extensive evaluations across six distinct tasks and sixteen datasets demonstrate that LFormer achieves competitive performance compared to state-of-the-art discrete and continuous baselines. These outcomes highlight its capabilities in long-range dependency modeling, fine-grained state tracking, and trajectory reconstruction from sparse and noisy observations, while maintaining linear scaling efficiency and empirical training stability over extended temporal horizons.
	\end{itemize}

Ultimately, these results illustrate how LGA embeds continuous-time dynamical principles into a parallel architecture, positioning LFormer as a scalable and versatile backbone for time series representation learning.

\section{Related Work \label{sec: related work}}
This section reviews prior literature along two axes. We first examine three temporal modeling paradigms distinguished by their mathematical treatment of time. We then discuss scalable parallelization mechanisms, focusing on fast-weight programmers and linear attention variants, which establish the computational foundation of the proposed architecture.
	
\subsection{Paradigms in Temporal Modeling}
\subsubsection{Discrete-time Models} 
This paradigm projects continuous temporal trajectories onto uniformly spaced discrete indices. Early approaches partition timeline into consecutive uniform bins~\cite{10.5555/2986916.2986950, pmlr-v56-Choi16, DBLP:conf/ihi/MarlinKKW12, DBLP:conf/mlhc/LiptonKW16}, and subsequently leverage imputation strategies~\cite{little2014statistical, 2018Recurrent, DBLP:conf/icml/YoonJS18, DBLP:conf/nips/SmiejaST0S18} to fill the resulting grids, yielding regularly sampled inputs on which standard architectures like RNNs~\cite{DBLP:conf/emnlp/ChoMGBBSB14, DBLP:journals/neco/HochreiterS97} and Transformers~\cite{DBLP:conf/nips/VaswaniSPUJGKP17} achieve strong performance. However, these heuristic preprocessing strategies introduce pseudo-observations that can alter the underlying continuous dynamics~\cite{cheng2025comprehensive, DBLP:journals/corr/abs-2012-00168}. Recent methods shift this discrete-time paradigm to the sequence level by grouping adjacent time steps into rigid patches~\cite{DBLP:conf/iclr/NieNSK23, DBLP:conf/aaai/HuangSZCDZW24, pmlr-v238-zhang24l, DBLP:conf/kdd/EkambaramJNSK23, DBLP:conf/iclr/LuoW24}. Although patching optimizes effective sequence lengths and mitigates the  self-attention quadratic complexity, these rigid boundary partitions compromise fine-grained temporal structure. Alternatively, several approaches enrich token representations with explicit temporal encoding, including sinusoidal functions~\cite{DBLP:conf/icde/WangJXWY22, DBLP:journals/tkde/JiangYXW24}, kernel-based methods~\cite{DBLP:conf/kdd/Ma0ZTDZG22}, and learnable embeddings~\cite{DBLP:conf/iclr/ShuklaM21, xu2019self}. Nevertheless, these encodings reduce time to an auxiliary feature attached to discrete tokens, rather than leveraging it as a continuous variable that could inform state transitions. Consequently, they offer limited capacity for capturing input-dependent continuous-time dynamics~\cite{DBLP:conf/nips/ChenRWFSL23}.

\subsubsection{Solver-dependent Continuous-time Models}
In contrast to discrete-time approaches, this paradigm parameterizes the derivative of continuous states via neural networks, solving the resulting differential equations through numerical methods~\cite{kidger2022neuraldifferentialequations}. This formulation affords a mathematically natural description of continuous-time dynamics. Pioneered by neural ordinary differential equations~\cite{DBLP:conf/nips/ChenRBD18}, this formulation underpins a broad lineage of architectures. Latent ODEs~\cite{DBLP:conf/nips/RubanovaCD19} extend this framework to generative modeling for irregularly sampled time series. Neural controlled differential equations~\cite{DBLP:conf/nips/KidgerMFL20} and neural rough differential equations~\cite{DBLP:conf/icml/MorrillSKF21} construct continuous input paths to drive continuous-time state evolution. Stochastic extensions~\cite{DBLP:conf/aistats/LiWCD20} model uncertainty within these latent processes, while continuous-time attention variants~\cite{DBLP:journals/pami/ChienC23, DBLP:conf/nips/ChenRWFSL23} integrate differential equations with the self-attention mechanism. Although mathematically principled, this solver-dependent paradigm incurs substantial computational overhead, particularly when integrating stiff equations~\cite{DBLP:journals/natmi/HasaniLALRTTR22}. Furthermore, sequential numerical integration inherently limits parallelization~\cite{lim2024parallelizing}. Moreover, truncation errors can accumulate over extended integration horizons~\cite{kuleshov2026denots}, progressively degrading the fidelity of the learned continuous-time dynamics.

\subsubsection{Solver-free Continuous-time Models}
To circumvent the computational bottlenecks of numerical integration, this closely related paradigm transforms continuous-time dynamics into directly evaluable algebraic forms. Three principal strategies are prevalent. First, closed-form approximations solve specific differential equations analytically, as exemplified by closed-form continuous-time (CfC) models~\cite{DBLP:journals/natmi/HasaniLALRTTR22} for liquid time-constant networks~\cite{DBLP:conf/aaai/HasaniLARG21}. Second, exact linear discretization rules from control theory govern state-space models~\cite{DBLP:conf/iclr/GuGR22, DBLP:journals/corr/abs-2312-00752} and their stochastic variants~\cite{park2025amortized}. Third, truncated algebraic path signatures from rough path theory encode continuous trajectories without invoking differential solvers~\cite{10.5555/3737916.3741287}. Despite eliminating solver-induced overhead, these methods introduce specific compromises. As documented in their respective literature, these trade-offs include potential gradient decay in closed-form recurrences~\cite{DBLP:journals/natmi/HasaniLALRTTR22}, constrained expressivity for highly nonlinear dynamics in linear transitions~\cite{DBLP:journals/corr/abs-2312-00752}, and structural information loss due to signature truncation~\cite{10.5555/3737916.3741287}.

While each paradigm offers distinct advantages, none simultaneously satisfies the three requirements identified in Section~\ref{sec: introduction}. This gap motivates a hybrid design that combines the parallel scalability of fast-weight attention with a continuous-time inductive bias derived from the LTC network---a synthesis absent from prior work and central to our proposed LGA.

\subsection{Parallelization via Fast Weights and Linear Attention}
The self-attention mechanism underpinning Transformers~\cite{DBLP:conf/nips/VaswaniSPUJGKP17} has become a dominant architecture in modern time series modeling~\cite{DBLP:journals/csr/ZhaoCXCWB26}. While explicit pairwise interactions afford this mechanism exceptional expressive power, its quadratic complexity in sequence length precludes scaling to the extended horizons typical of complex dynamical systems.

To mitigate this bottleneck, linear attention~\cite{DBLP:conf/icml/KatharopoulosV020} restructures the attention computation. By replacing the softmax operator with kernel feature maps, this formulation exploits matrix associativity to achieve linear complexity with respect to sequence length. Interpreting the attention state as a matrix-valued memory updated via outer products establishes a formal equivalence with fast-weight programmers~\cite{DBLP:journals/neco/Schmidhuber92, DBLP:conf/icml/SchlagIS21}, providing a principled framework for parallelizing recurrent architectures. Subsequent advances have enriched this framework with data-dependent gating mechanisms, including forget gates in RetNet~\cite{DBLP:journals/corr/abs-2307-08621}, gated linear attention~\cite{DBLP:conf/icml/YangWSPK24}, and the RWKV series~\cite{DBLP:conf/emnlp/PengAAAABCCCDDG23, DBLP:journals/corr/abs-2404-05892, DBLP:journals/corr/abs-2503-14456}. Concurrently, state-space models such as Mamba~\cite{DBLP:journals/corr/abs-2312-00752, DBLP:conf/icml/DaoG24} and DeltaNet~\cite{DBLP:conf/nips/YangWZSK24} achieve linear complexity through input-dependent transitions and hardware-aware associative scans. Despite their remarkable computational efficiency, these modern variants are predominantly designed for the discrete-time domain~\cite{zhangsurvey}, where gating operators and state transitions are parameterized by integer step indices.

This historical focus leaves a key open question: can the parallelization principles of fast-weight attention be integrated with continuous-time, input-driven gating dynamics? Addressing this question requires embedding observed temporal distances into fast-weight memory, enabling the architecture to process irregularly sampled time series---a capability absent from existing parallel architectures. This requirement underpins the design of LGA.

\section{Background: Liquid Time-constant Gating \label{sec: background}}
Liquid time-constant networks~\cite{DBLP:conf/aaai/HasaniLARG21} are continuous-time recurrent neural models. At time $t$, an LTC network determines the $d$-dimensional continuous state vector $\mathbf{s}(t) \in \mathbb{R}^{d}$ via the following initial value problem (IVP):
\begin{equation}
	\frac{\mathrm{d}\mathbf{s}(t)}{\mathrm{d}t}
	=-\Bigl[\mathbf{w}_\tau+f\bigl(\mathbf{s}(t), \mathbf{x}(t)\bigr)\Bigr]
	\odot
	\mathbf{s}(t)+\mathbf{b}_\tau \odot f\bigl(\mathbf{s}(t), \mathbf{x}(t)\bigr),
	\label{eq: ltc ivp}
\end{equation}
where $\mathbf{x}(t) \in \mathbb{R}^{m}$ denotes the exogenous input to the system, and $\mathbf{w}_\tau, \mathbf{b}_\tau \in \mathbb{R}^{d}$ represent the intrinsic inverse time-constant weight and bias, respectively. The function $f(\cdot)$ is a strictly positive, bounded, continuous, and monotonically increasing nonlinearity, while $\odot$ denotes the element-wise product. 

\noindent\emph{Discussion.} The ``liquid'' property stems from the input-driven modulation of the system's effective inverse time constants, defined as $\mathbf{w}_\tau + f\bigl(\mathbf{s}(t), \mathbf{x}(t)\bigr)$. This mechanism allows individual units to adjust their intrinsic time constants based on the incoming signal, enabling the network to adaptively characterize varying temporal dynamics.

To achieve analytical tractability, we first introduce the one-dimensional LTC configuration which isolates a single scalar state unit $s(t) \in \mathbb{R}$ by removing self-connections~\cite{DBLP:journals/natmi/HasaniLALRTTR22}. Driven by a scalar exogenous input $x(t) \in \mathbb{R}$, this continuous state follows a linear non-autonomous variant of Eq.~\eqref{eq: ltc ivp}:
\begin{equation}
	\frac{\mathrm{d}s(t)}{\mathrm{d}t}
	=
	-\bigl[w_\tau + f\bigl(x(t)\bigr)\bigr]s(t)
	+ b_\tau f\bigl(x(t)\bigr),
	\label{eq: 1d ltc ivp}
\end{equation}
where $w_\tau, b_\tau\in\mathbb{R}$ are scalar parameters. To derive the solvable form, we follow the structural symmetry assumption established in~\cite{DBLP:journals/natmi/HasaniLALRTTR22}, which introduces a balancing inverse time-constant parameter into the bias compartment of Eq.~\eqref{eq: 1d ltc ivp}:
\begin{equation}
	\frac{\mathrm{d}s(t)}{\mathrm{d}t} = -\bigl[w_\tau+f(x(t))\bigr]s(t) + \bigl[w_\tau+f(x(t))\bigr]b_\tau.
	\label{eq:symmetric_ode}
\end{equation}
Under this symmetric condition, applying the linear ODE theory~\cite{1991Differential} yields the integral solution:
\begin{equation}
	s(t) = \bigl(s(0)-b_\tau\bigr) \exp\left[-w_\tau t - \int_{0}^{t} f\bigl(x(u)\bigr)\,\mathrm{d}u\right] + b_\tau,
	\label{eq: 1d ltc closed-form integral solution}
\end{equation}
where $s(0) \in \mathbb{R}$ denotes the initial state boundary at $t=0$. Section S1 in the Supplementary Material details the algebraic verification under this symmetric condition.

Let $t_{n-1}$ and $t_n$ denote two consecutive, irregularly spaced observation timestamps where $t_{n-1}<t_n$. Defining the local temporal interval as $\delta_n = t_n - t_{n-1}$, evaluating the continuous integral solution in Eq.~\eqref{eq: 1d ltc closed-form integral solution} across this boundary yields the following state transition:
\begin{equation}
	\frac{s(t_n)-b_\tau}{s(t_{n-1})-b_\tau}
	= \exp\left[-w_\tau \delta_n - \int_{t_{n-1}}^{t_n} f\bigl(x(u)\bigr)\,\mathrm{d}u\right].
	\label{eq: 1d ltc state transition}
\end{equation}
Algebraic rearrangement of Eq.~\eqref{eq: 1d ltc state transition} produces a recurrent state update rule modulated by a continuous-time, input-driven gating mechanism:
\begin{align}
	s(t_n) &= g(t_n)s(t_{n-1}) + \bigl[1 - g(t_n)\bigr]b_\tau,
	\label{eq: 1d ltc rnn}
	\\
	g(t_n) &= \exp\left[-w_\tau \delta_n - \int_{t_{n-1}}^{t_n} f\bigl(x(u)\bigr)\,\mathrm{d}u\right].
	\label{eq: 1d ltc gate}
\end{align}
\noindent\emph{Discussion.} We refer to Eq.~\eqref{eq: 1d ltc gate} as the liquid gate which structurally encapsulates two mathematical properties:
\begin{itemize}
	\item \textbf{Temporal Decay.} The component $-w_\tau \delta_n$ characterizes the intrinsic continuous decay of the state trajectory over the observed temporal distance $\delta_n$.
	
	\item \textbf{Input Adaptability.} The integral $-\int_{t_{n-1}}^{t_n} f(x(u))\mathrm{d}u$ aggregates the cumulative modulation of exogenous inputs across the same localized time horizon.
\end{itemize}
The exponential operator maps these properties into a unified and bounded gate $g(t_n)\in(0, 1)$. Derived from LTC dynamics, this formulation endows the liquid gate with continuous-time inductive bias and noise-attenuation properties. This closed-form relation provides the continuous-time, input-driven gating mechanism that underpins LGA Step~(1).
	
\section{Methodology: Liquid Gated Attention \label{sec: methodology}}
In this section, we formulate Liquid Gated Attention, a solver-free parallel operator for continuous-time representation learning. Building upon the continuous-time inductive bias established in Section~\ref{sec: background}, the subsequent derivation proceeds through three conceptual layers: (1) a learnable endpoint interpolation that approximates the liquid gate while circumventing numerical integration; (2) an expansion of the scalar state to a matrix-valued associative memory, enabling expressive sequence modeling; and (3) a sequence-level normalization that stabilizes optimization over extended sequences. Consequently, LGA functions as a continuous-time-inspired parallel operator rather than a numerically exact solver translating from the original LTC dynamics. 

\subsection{Deriving the Recurrent Form of LGA \label{subsec: recurrent LGA}}
The RNN-like hidden state evolution in the 1D LTC system exposes two limitations: the computational overhead of the integral-based liquid gate and the limited capacity of the scalar hidden state. The following subsections elaborate on the solutions tailored to these two issues, respectively.
	
\subsubsection{A Learnable Integral Approximation}
We approximate the integral-based liquid gate in Eq.~\eqref{eq: 1d ltc gate} via a learnable endpoint interpolation inspired by the trapezoidal rule. Given two consecutive input representations $\mathbf{x}_{n-1}, \mathbf{x}_n \in \mathbb{R}^{1 \times d}$, a shared nonlinearity first projects each token into a scalar variable:
\begin{equation}
	\begin{aligned}
		\tilde{x}_{n-1} &= \sigma\left(\mathbf{x}_{n-1}\mathbf{w}_f+b_f\right), \\ \tilde{x}_n &= \sigma\left(\mathbf{x}_n\mathbf{w}_f+b_f\right),
		\label{eq: non-linear}
	\end{aligned}
\end{equation}
where $\mathbf{w}_f \in \mathbb{R}^{d \times 1}$ and $b_f \in \mathbb{R}$ are learnable parameters, and $\sigma(\cdot)$ denotes the sigmoid activation function.

We then approximate the integral via a learnable weighted average across the boundary endpoints:
\begin{equation}
	\int_{t_{n-1}}^{t_n}\tilde{x}(u)\,\mathrm{d}u
	\approx
	\delta_n
	\underbrace{
		\left[\mu\tilde{x}_{n-1}+(1-\mu)\tilde{x}_{n}\right]}_{\text{defined as }\overline{x}_n}
	=
	\delta_n\overline{x}_n,
\end{equation}
where $\delta_n=t_n-t_{n-1}$ denotes the time interval. We parameterize the interpolation coefficient as $\mu=\sigma(\theta)\in(0, 1)$ with learnable $\theta\in\mathbb{R}$ to balance each endpoint's contribution. When $\mu = 0.5$, this interpolation recovers the classical trapezoidal rule with second-order local accuracy. In the general learnable case, it serves as a trainable numerical surrogate rather than a fixed quadrature rule. Section~S2 of the Supplementary Material provides the proof.

For implementation simplicity, we merge the intrinsic inverse time-constant $w_\tau$ and the input-dependent term into a single decay surrogate, simplifying Eq.~\eqref{eq: 1d ltc gate} to the following efficient liquid gate:
\begin{equation}
	g_n = \exp\left[-\delta_n \overline{x}_n\right].
	\label{eq: liquid_gating_mechanism}
\end{equation}
This formulation preserves the temporal decay and input adaptability functionalities of LTC while eliminating the need for numerical integration. This completes LGA Step~(2) Computational Efficiency.

\begin{figure}[t]
	\centering
	\includegraphics[scale=0.25]{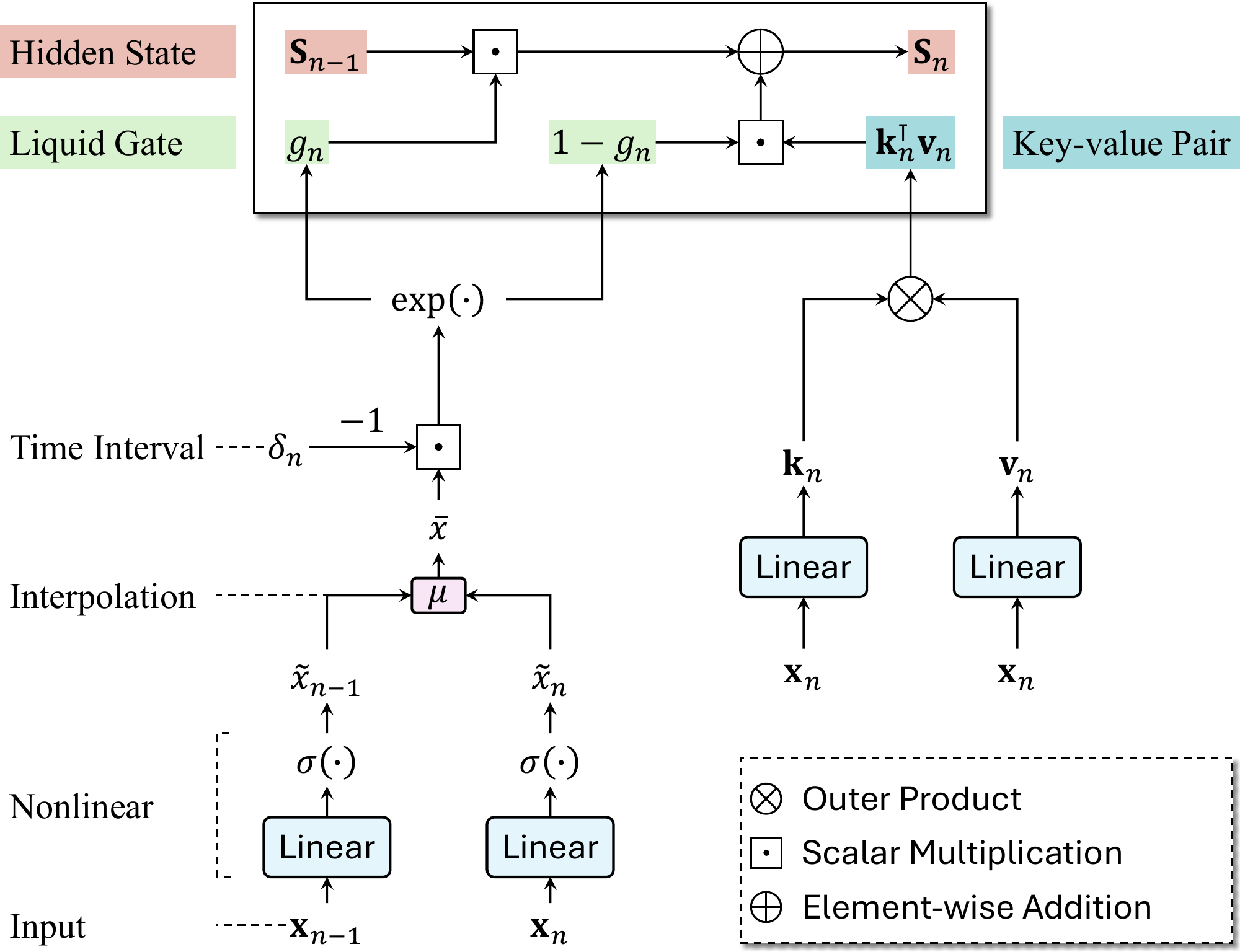}
	\caption{LGA hidden state update mechanics. The \textbf{upper block} depicts the recurrent state transition where the liquid gates modulate the combination of historical memory and associative update matrix. The \textbf{lower pipeline} details the liquid gate evaluation workflow incorporating nonlinear projection, learnable interpolation, temporal scaling, and exponential mapping.}
	\label{recurrent form figure}
\end{figure}

\subsubsection{An Expressive Hidden State}
We elevate the hidden state from a scalar to a matrix $\mathbf{S}_n\in\mathbb{R}^{d\times d}$ that expands the memory storage capacity for long-range modeling. Furthermore, a dynamic, input-dependent associative memory term substitutes the static bias $b_\tau$ in Eq.~\eqref{eq: 1d ltc rnn}, improving model flexibility in tracking instantaneous feature variations. Following the linear attention framework~\cite{DBLP:conf/icml/KatharopoulosV020}, the input vector $\mathbf{x}_n$ is projected into key and value vectors $\mathbf{k}_n, \mathbf{v}_n \in \mathbb{R}^{1 \times d}$:
\begin{equation}
	\begin{aligned}
		\mathbf{k}_n&=\mathbf{x}_n\mathbf{W}_k/\sqrt{d}+\mathbf{b}_k,\\
		\mathbf{v}_n&=\mathbf{x}_n\mathbf{W}_v+\mathbf{b}_v,
	\end{aligned}
\end{equation}
where $\mathbf{W}_k, \mathbf{W}_v \in \mathbb{R}^{d \times d}$ and $\mathbf{b}_k, \mathbf{b}_v \in \mathbb{R}^{1 \times d}$ are learnable parameters, and $\sqrt{d}$ represents the scaling factor. As visualized in Fig.~\ref{recurrent form figure}, the recurrent hidden state update rule in LGA is:
\begin{equation}
	\mathbf{S}_n = g_n \mathbf{S}_{n-1} + (1 - g_n)\mathbf{k}_n^\top \mathbf{v}_n.
	\label{eq: lga_recurrent_update}
\end{equation}
		
\noindent\emph{Discussion.} Eq.~\eqref{eq: lga_recurrent_update} implements the fast-weight programming scheme~\cite{DBLP:journals/neco/Schmidhuber92}, where $\mathbf{S}_n$ acts as an associative memory updated via vector outer products. This elevation expands the representational capacity into a $d \times d$ interaction matrix~\cite{DBLP:conf/nips/BeckPSAP0KBH24, DBLP:journals/corr/abs-2404-05892, DBLP:conf/icml/DaoG24}, fulfilling LGA Step~(3) Expressive Capability.

The output retrieval in LGA handles the feature extraction and state readout from associative memory matrices. At time $t_n$, the input representation $\mathbf{x}_n$ is mapped to the query vector $\mathbf{q}_n \in \mathbb{R}^{1 \times d}$ and the output gate $\mathbf{o}_n \in \mathbb{R}^{1 \times d}$ via:
\begin{align}
	\mathbf{q}_n &= \mathbf{x}_n \mathbf{W}_q + \mathbf{b}_q, \label{eq: query_proj} \\
	\mathbf{o}_n &= \sigma\left(\mathbf{x}_n \mathbf{W}_o + \mathbf{b}_o\right), \label{eq: og_proj}
\end{align}
where $\mathbf{W}_q, \mathbf{W}_o \in \mathbb{R}^{d \times d}$ and $\mathbf{b}_q, \mathbf{b}_o \in \mathbb{R}^{1 \times d}$ are learnable weights and biases, respectively. The output gate then modulates the retrieved context within the hidden state matrix $\mathbf{S}_n$ to produce the final output vector $\mathbf{y}_n \in \mathbb{R}^{1 \times d}$ via:
\begin{equation}
	\mathbf{y}_n = \mathbf{o}_n \odot \left(\mathbf{q}_n \mathbf{S}_n\right).
	\label{eq: output_retrieval}
\end{equation}
The output gate allows the model to suppress irrelevant or noisy retrieved channels, improving robustness against high-frequency observational perturbations.	

\subsection{Deriving the Parallel Former of LGA \label{subsec: parallel LGA}}
Assuming a zero-initialized hidden state $\mathbf{S}_0 = \mathbf{0}$, unrolling the recurrent update rule in Eq.~\eqref{eq: lga_recurrent_update} across the temporal horizon yields:
\begin{equation}
	\mathbf{S}_n = \sum_{i=1}^{n} \left[ \left( \prod_{j=i+1}^{n} g_j \right) \left(1 - g_i\right) \mathbf{k}_i \right]^\top \mathbf{v}_i.
	\label{eq: unrolled_hidden_state}
\end{equation}
Substituting Eq.~\eqref{eq: unrolled_hidden_state} into the output retrieval expression in Eq.~\eqref{eq: output_retrieval} and factoring out the $n$-dependent terms from the temporal summation yields:
\begin{equation}
	\mathbf{y}_n = \mathbf{o}_n \odot \left[ \left(\dot{g}_n \mathbf{q}_n\right) \sum_{i=1}^{n} \left[\left(1 - g_i\right) \mathbf{k}_i\right]^\top \left(\frac{\mathbf{v}_i}{\dot{g}_i}\right) \right],
	\label{eq: lga_parallel_unroll}
\end{equation}
where $\dot{g}_n = \prod_{i=1}^{n} g_i \in \mathbb{R}$ represents the cumulative liquid gate product. Under zero initialization, Eq.~\eqref{eq: lga_parallel_unroll} remains algebraically equivalent to the recurrent updates in Eq.~\eqref{eq: lga_recurrent_update}. Temporal index alignment enables these liquid gates to directly scale token representations, transforming the continuous-time inductive bias into explicit modulation coefficients $g_n$ and $\dot{g}_n$ for the query, key, and value vectors. 

Building upon Eq.~\eqref{eq: lga_parallel_unroll}, exploiting matrix associativity within the fast-weight programming framework~\cite{DBLP:journals/neco/Schmidhuber92} can eliminate sequential dependencies in the recurrent state. This yields two linearly scaling parallel modes for causal and non-causal computation, depending on the temporal aggregation boundaries. Given the modulated features $\dot{\mathbf{Q}}, \dot{\mathbf{K}}, \dot{\mathbf{V}}\in\mathbb{R}^{n\times d}$ and the element-wise output gate $\mathbf{O}\in\mathbb{R}^{n\times d}$, the parallel matrix transformations in LGA that map these inputs to the output matrix $\mathbf{Y} \in \mathbb{R}^{n \times d}$ are instantiated as:
\begin{itemize}
	\item \textbf{Causal Mode.} Online streaming and autoregressive generation tasks strictly prohibit the future information leakage. Leveraging a parallel prefix scan, the causal LGA operator is formulated as:
	\begin{equation}
		\mathbf{Y} = \mathbf{O} \odot \left[ \dot{\mathbf{Q}} \circledast \mathrm{CumSum}\left( \dot{\mathbf{K}} \otimes \dot{\mathbf{V}} \right) \right],
		\label{eq: lga_causal_linear_form}
	\end{equation}
	where $\otimes$ and $\circledast$ denote the row-wise outer product and vector-matrix multiplication, respectively. $\mathrm{CumSum}(\cdot)$ computes the associative cumulative sum along the temporal dimension. This implementation maintains causal semantics while reducing both computational and memory complexity to $\mathcal{O}(nd^2)$.
	
	\item\textbf{Non-causal Mode.} For offline representation learning tasks where the strict causality is unnecessary, the temporal aggregation directly encompasses the full sequence. Replacing the prefix scan with a global temporal summation, LGA functions as a bidirectional encoder that
	\begin{equation}
		\mathbf{Y} = \mathbf{O} \odot \left[ \dot{\mathbf{Q}} \left(\dot{\mathbf{K}}^\top \dot{\mathbf{V}}\right) \right],
		\label{eq: lga_parallel_matrix_form}
	\end{equation}
	which prioritizes the matrix product between the key and value matrices to preserve the linear complexity $\mathcal{O}\left(nd^2\right)$.
\end{itemize}

\begin{figure}[t]
	\centering
	\includegraphics[scale=0.25]{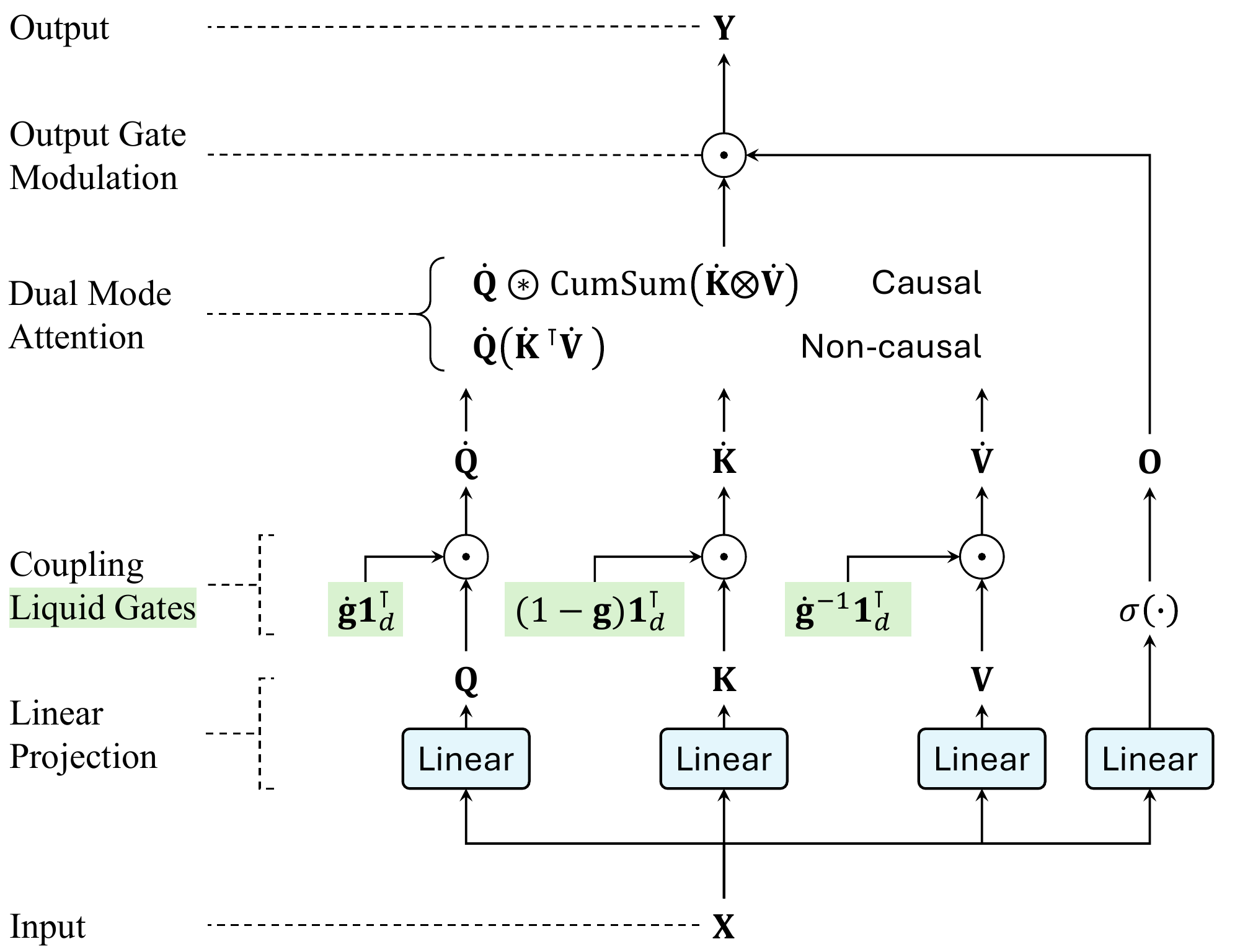}
	\caption{Parallel LGA pipeline. The input representation $\mathbf{X}$ is first projected and scaled by vectorized liquid gates to embed continuous-time gating dynamics, yielding modulated representations. The dual mode attention block then executes either a causal prefix scan or a global matrix multiplication with linear complexity. The output gate modulates the aggregated state to produce the final representation. $\odot$, $\circledast$, and $\otimes$ denote the element-wise product, row-wise vector-matrix multiplication, and row-wise matrix outer product, respectively. $\mathrm{CumSum}(\cdot)$ computes the cumulative sum along the temporal dimension.}
	\label{parallel form figure}
\end{figure}

We summarize Eq.~\eqref{eq: lga_causal_linear_form} and Eq.~\eqref{eq: lga_parallel_matrix_form} into a unified form:
\begin{equation}
	\mathbf{Y}=\mathrm{LGA}(\mathbf{X}, \mathbf{\Delta}),
\end{equation}
since liquid gates, modulated features and output gate are calculated based on the input sequence representation matrix $\mathbf{X}\in\mathbb{R}^{n\times d}$ and the corresponding time interval vector $\mathbf{\Delta}\in\mathbb{R}^{n\times 1}$. Figure~\ref{parallel form figure} visualizes the LGA pipeline. 

The base features $\mathbf{Q}, \mathbf{K}, \mathbf{V} \in \mathbb{R}^{n \times d}$ and the output gate $\mathbf{O}$ are obtained via linear projections of the input matrix $\mathbf{X}$:
\begin{align}
	\mathbf{Q} &= \mathbf{X}\mathbf{W}_q+\mathbf{b}_q,\\
	\mathbf{K} &= \mathbf{X}\mathbf{W}_k/\sqrt{d}+\mathbf{b}_k,\\
	\mathbf{V} &= \mathbf{X}\mathbf{W}_v+\mathbf{b}_v,\\
	\mathbf{O} &= \sigma\left(\mathbf{X}\mathbf{W}_o+\mathbf{b}_o\right).
\end{align}
The modulated features $\dot{\mathbf{Q}}, \dot{\mathbf{K}}, \dot{\mathbf{V}} \in \mathbb{R}^{n \times d}$ are obtained by scaling with the vectorized liquid gates $\mathbf{g}, \dot{\mathbf{g}} \in \mathbb{R}^{n \times 1}$:
\begin{align}
	\dot{\mathbf{Q}} &= \mathbf{Q} \odot \left(\dot{\mathbf{g}}\mathbf{1}_d^\top\right), \label{eq: q_modulation} \\
	\dot{\mathbf{K}} &= \mathbf{K} \odot \left[\left(1 - \mathbf{g}\right)\mathbf{1}_d^\top\right], \label{eq: k_modulation} \\
	\dot{\mathbf{V}} &= \mathbf{V} \odot \left(\dot{\mathbf{g}}^{-1}\mathbf{1}_d^\top\right). \label{eq: v_modulation}
\end{align}
where the scalar $1$ performs element-wise broadcasting and the vector $\mathbf{1}_d^\top \in \mathbb{R}^{1 \times d}$ denotes the channel-wise broadcasting.

The parallel computation of the liquid gate vector $\mathbf{g}$ is performed by applying the operations in Eqs.~\eqref{eq: non-linear}---\eqref{eq: liquid_gating_mechanism} simultaneously across the temporal dimension, as formalized in Algorithm~\ref{alg:lga_gate_computation}. To evaluate the cumulative liquid gate $\dot{\mathbf{g}}$, the sequential multiplication chain is mapped into log-space which reformulates the recurrent dependency into the cumulative summation, yielding
\begin{equation}
	\dot{\mathbf{g}} = \exp\left[-\mathrm{CumSum}\left(\mathbf{\Delta} \odot \overline{\mathbf{x}}\right)\right]
	\label{eq: lga_cumulative_gate}
\end{equation}
where $\overline{\mathbf{x}}\in\mathbb{R}^{n\times 1}$ represents the interpolated nonlinear vector.

\subsubsection{Sequence-level Normalization} 
Due to the cumulative term $\dot{\mathbf{g}}$, the vanilla parallel execution is prone to numerical instability over long sequences. Since the inputs are non-negative, the unconstrained prefix sum in Eq.~\eqref{eq: lga_cumulative_gate} grows monotonically with the sequence length, leading to exponential underflow and vanishing gradients.
	
To stabilize long-horizon optimization, we introduce a sequence-level normalization mechanism. Let $\mathbf{u} = \mathbf{\Delta} \odot \overline{\mathbf{x}} \in \mathbb{R}^{n \times 1}$. The scaling vector $\mathbf{u}$ is normalized by its summation along the sequence dimension:
\begin{equation}
	\hat{\mathbf{u}} = \frac{\mathbf{u}}{\sum_{i=1}^{n} u_i + \epsilon},
	\label{eq: lga_normalization_operator}
\end{equation}
where $u_i=\delta_i\overline{x}_i$ is the $i$-th element of $\mathbf{u}$, and $\epsilon$ is a small constant added to prevent division by zero. Substituting $\mathbf{u}$ with $\hat{\mathbf{u}}$ bounds the cumulative sum within $(0, 1]$. Consequently, the exponential scaling coefficients remain constrained to $(e^{-1}, 1]$, as proven in Section S3 of the Supplementary Material. Thus, the stable local and cumulative liquid gates are formulated as:
\begin{align}
	\mathbf{g}_{\text{stable}} &= \exp\left(-\hat{\mathbf{u}}\right),
	\label{eq: lga_stable_local_gate}
	\\
	\dot{\mathbf{g}}_{\text{stable}} &= \exp\left[-\mathrm{CumSum}\left(\hat{\mathbf{u}}\right)\right].
	\label{eq: lga_stable_cumulative_gate}
\end{align}
Algorithm~\ref{alg:lga_gate_computation} formalizes the complete parallel evaluation flow of these liquid gate vectors, including the alternative numerical stabilization path.

\begin{algorithm}[t]
	\caption{Parallel Computation of Liquid Gates under Vanilla and Stabilized Modes}
	\label{alg:lga_gate_computation}
	\makeatletter
	\renewcommand{\algorithmicrequire}{\textbf{Input:}}
	\renewcommand{\algorithmicensure}{\textbf{Output:}}
	\makeatother
	\begin{algorithmic}[1]	
		\REQUIRE Input matrix $\mathbf{X} \in \mathbb{R}^{n \times d}$, time interval vector $\mathbf{\Delta} \in \mathbb{R}^{n\times 1}$, weight vector $\mathbf{w}_f \in \mathbb{R}^{d \times 1}$, bias scalar $b_f \in \mathbb{R}$, unconstrained parameter $\theta \in \mathbb{R}$, stability constant $\epsilon > 0$, mode selector $\mathcal{M} \in \{\text{Vanilla}, \text{Stabilized}\}$.
		\ENSURE Local liquid gate vector $\mathbf{g} \in \mathbb{R}^{n \times 1}$ and cumulative liquid gate vector $\dot{\mathbf{g}} \in \mathbb{R}^{n \times 1}$.
		
		\STATE $\mu \leftarrow \sigma(\theta)$ \hfill $\triangleright$ Bounded interpolation coefficient
		
		\STATE $\tilde{\mathbf{x}} \leftarrow \sigma(\mathbf{X}\mathbf{w}_f + b_f)$ \hfill $\triangleright$ Nonlinear projection
		
		\STATE $\tilde{\mathbf{x}}_{\text{prev}} \leftarrow [0; \tilde{x}_1; \dots; \tilde{x}_{n-1}]$ \hfill $\triangleright$ Temporal shift
		
		\STATE $\overline{\mathbf{x}} \leftarrow \mu\tilde{\mathbf{x}}_{\text{prev}} + (1 - \mu)\tilde{\mathbf{x}}$ \hfill $\triangleright$ Interpolation
		
		\STATE $\mathbf{u} \leftarrow \mathbf{\Delta} \odot \overline{\mathbf{x}}$ \hfill $\triangleright$ Base decay vector
		
		\IF{$\mathcal{M} = \text{Stabilized}$}
		\STATE $\hat{\mathbf{u}} \leftarrow \mathbf{u} \,/\, (\sum_{i=1}^{n} u_i + \epsilon)$ \hfill $\triangleright$ Sequence-level normalization
		\STATE $\mathbf{g} \leftarrow \exp(-\hat{\mathbf{u}})$ \hfill $\triangleright$ Stabilized local gate
		\STATE $\dot{\mathbf{g}} \leftarrow \exp(-\mathrm{CumSum}(\hat{\mathbf{u}}))$ \hfill $\triangleright$ Stabilized cumulative gate
		\ELSE
		\STATE $\mathbf{g} \leftarrow \exp(-\mathbf{u})$ \hfill $\triangleright$ Vanilla local gate
		\STATE $\dot{\mathbf{g}} \leftarrow \exp(-\mathrm{CumSum}(\mathbf{u}))$ \hfill $\triangleright$ Vanilla cumulative gate
		\ENDIF
		
		\RETURN $\mathbf{g}$, $\dot{\mathbf{g}}$
	\end{algorithmic}
\end{algorithm}

\noindent\emph{Discussion.} The normalized gate is not algebraically equivalent to the original LTC state transition. Instead, it operates as a sequence-level regularized surrogate that preserves relative temporal distance allocation while bounding the total decay budget. Moreover, this mechanism reduces the gradient decay from exponential to polynomial as the sequence length grows (see Section S3 of the Supplementary Material). This enables stable optimization over extended sequences, fulfilling Step~(4) Numerical Safety. This design targets offline full-sequence representation learning; online streaming or causal inference requires a prefix-normalized variant to eliminate dependence on future observations.

\subsubsection{A Decoupled Multi-Head Extension}
To scale the continuous-time gating dynamics to high-dimensional spaces, we derive a decoupled multi-head LGA mechanism. 

Specifically, the $d$-dimensional latent space is partitioned into $H$ distinct subspaces, where each head $h \in \left\{1, \dots, H\right\}$ operates on a subspace of dimension $d_h = d/H$. Let $\mathbf{W}^{(h)}$, $\mathbf{b}^{(h)}$, and $\mu^{(h)}$ denote the head-specific learnable projection weights, biases, and adaptive interpolation coefficients for the $h$-th subspace, respectively. Each partitioned head executes the linear projection, adaptive interpolation, and sequence-level normalization procedures independently, yielding the decoupled subspace representations:
\begin{equation}
	\mathbf{Y}^{(h)} = \mathrm{LGA}^{(h)}\left(\mathbf{X}, \mathbf{\Delta}\right), \quad h \in \left\{1, \dots, H\right\}.
	\label{eq: lga_head_h_output}
\end{equation}
Subsequently, the outputs from all $H$ heads are concatenated along the feature dimension and linearly projected to construct the final multi-head representation:
\begin{equation}
	\mathbf{Y} = \mathrm{MHLGA}\left(\mathbf{X}, \mathbf{\Delta}\right) = \mathrm{Concat}\left(\mathbf{Y}^{(1)}, \dots, \mathbf{Y}^{(H)}\right)\mathbf{W}_{m},
	\label{eq: mhlga_final_projection}
\end{equation}
where $\mathrm{Concat}\left(\cdot\right)$ denotes the concatenation operator and $\mathbf{W}_{m} \in \mathbb{R}^{d \times d}$ represents the terminal projection weight matrix.

\noindent\emph{Discussion.} Unlike standard multi-head attention that primarily captures heterogeneous subspace features~\cite{DBLP:conf/nips/VaswaniSPUJGKP17}, the key architectural advancement lies in the head-wise decoupling of continuous-time gating dynamics. Equipping each head with independent interpolation parameters and liquid gates enhances the model's expressive capability at the system level, enabling collaborative modeling of complex temporal patterns.

\begin{figure}[t]
	\centering
	\includegraphics[width=\linewidth]{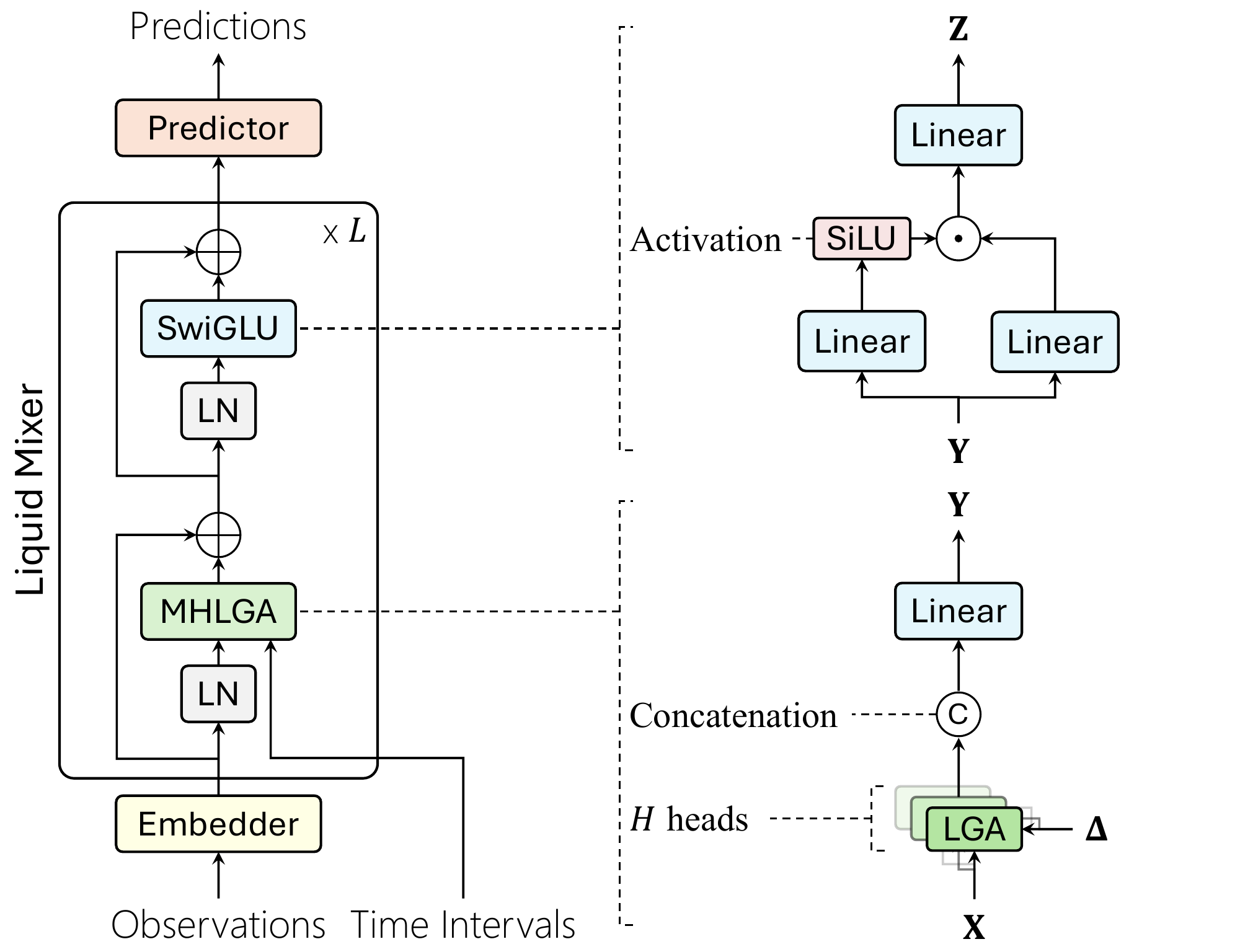}
	\caption{LFormer architecture. The \textbf{left block} depicts the macro design, where $L$ stacked Liquid Mixers connect a task-specific Embedder and Predictor. The \textbf{right panels} detail the MHLGA (bottom) and SwiGLU (upper) workflows.}
	\label{lformer figure}
\end{figure}

\subsection{LFormer: An Instantiation of LGA \label{subsec: LFormer}}
Building upon the LGA operator, we establish a modular backbone for continuous-time representation learning system, designated as LFormer. As illustrated in Fig.~\ref{lformer figure}, the architecture employs a three-stage processing pipeline consisting of a task-specific Embedder, $L$ stacked Liquid Mixers, and a task-oriented Predictor.

\subsubsection{Embedder}
The Embedder projects raw input observations into a $d$-dimensional representation space. To accommodate variable-length sequences, we define a canonical sequence length $n$. The longer sequences are partitioned into non-overlapping sub-sequences of length $n$, while shorter sequences are extended to length $n$ via a zero-padding scheme. The padding tokens are excluded from the gradient optimization. Formally, given a raw or pre-processed time series with $m$ features $\mathbf{X}_{\text{raw}}\in\mathbb{R}^{n\times m}$, the Embedder operates via:
\begin{equation}
	\mathbf{X} = \mathrm{Embed}\left(\mathbf{X}_{\text{raw}}\right),
	\label{eq: lformer_embedder}
\end{equation}
where $\mathrm{Embed}\left(\cdot\right): \mathbb{R}^m \rightarrow \mathbb{R}^d$ denotes a task-specific transformation mapping raw observations to the latent sequence representation matrix $\mathbf{X} \in \mathbb{R}^{n \times d}$.

\subsubsection{Liquid Mixer}
The Liquid Mixer serves as the core block of LFormer, designed to refine latent representations through interleaved temporal and channel-wise transformations. The framework stacks $L$ identical mixer layers, where the $l$-th mixer ($l \in \left\{1, \dots, L\right\}$) processes the representations generated by its predecessor or the initialization from the Embedder.
	
Specifically, each mixer layer consists of a multi-head liquid gated attention ($\mathrm{MHLGA}$) module and a swish-gated linear unit ($\mathrm{SwiGLU}$), organized in a residual connection structure with pre-layer normalization ($\mathrm{LN}$):
\begin{align}
		\mathbf{Y} &= \mathbf{X} + \mathrm{MHLGA}\left(\mathrm{LN}\left(\mathbf{X}\right), \mathbf{\Delta}\right),\\
		\mathbf{Z} &= \mathbf{Y} + \mathrm{SwiGLU}\left(\mathrm{LN}\left(\mathbf{Y}\right)\right),
\end{align}
where $\mathbf{Z} \in \mathbb{R}^{n \times d}$ denotes the output of the current mixer layer. The $\mathrm{MHLGA}$ component operates along the temporal sequence dimension to inject continuous-time inductive bias via decoupled liquid gates, whereas the $\mathrm{SwiGLU}$ block modulates cross-feature channel interactions to enhance non-linear expressivity. For any hidden representation $\mathbf{H}\in\mathbb{R}^{n\times d}$, the operation in $\mathrm{SwiGLU}$ is:
\begin{equation}
	\mathrm{SwiGLU}\left(\mathbf{H}\right) = \left[ \mathrm{SiLU}\left(\mathbf{H}\mathbf{W}_1\right) \odot \left(\mathbf{H}\mathbf{W}_2\right) \right] \mathbf{W}_3,
\end{equation}
where $\mathbf{W}_1, \mathbf{W}_2 \in \mathbb{R}^{d \times d_{ff}}$ and $\mathbf{W}_3 \in \mathbb{R}^{d_{ff} \times d}$ represent learnable weight matrices, and $\mathrm{SiLU}\left(x\right) = x \cdot \sigma\left(x\right)$ represents the sigmoid linear unit.

\subsubsection{Predictor}
The final output of the terminal Liquid Mixer layer, denoted as $\mathbf{Z}^{(L)} \in \mathbb{R}^{n \times d}$, serves as the high-level semantic representation of the input sequence. The Predictor maps this representation into the target task space:
\begin{equation}
	\mathbf{P}=\mathrm{Pred}(\mathbf{Z}^{(L)}),
\end{equation}
where $\mathbf{P}$ represents the task-oriented target output (e.g., forecasted future trajectories or classification probabilities). Depending on downstream constraints, the operator $\mathrm{Pred}\left(\cdot\right)$ can be instantiated as a linear projection head, a global pooling operator, or a multilayer perceptron (MLP). 
	
\section{Experiments and Discussion \label{sec: experiments and discussion}}
This section empirically evaluates LFormer around three core research questions (RQs):
\begin{itemize}
	\item \textbf{RQ1 (Comparative Performance):} How does LFormer perform against representative state-of-the-art discrete-time and continuous-time baselines across diverse temporal tasks, regarding long-range dependency modeling, fine-grained state tracking, and trajectory reconstruction from sparse and noisy observations?
	
	\item \textbf{RQ2 (Ablation Analysis):} How do the individual components of LFormer, including the continuous-time liquid gate dynamics, sequence-level normalization, and decoupled multi-head mechanism, contribute to its representational capacity and numerical stability?
	
	\item \textbf{RQ3 (Computational Efficiency):} Does LFormer achieve linear computational and memory complexity under both non-causal and causal settings while maintaining competitive accuracy over extended sequence horizons?
\end{itemize}

\subsection{Experimental Setup \label{sec: exp_setup}}
\begin{table}[t]
	\centering
	\caption{A Summary of 16 Datasets across 6 Tasks}
	\label{tab:dataset_statistics}
	\begin{minipage}{0.48\textwidth}
		\resizebox{\linewidth}{!}{
			\begin{tabular}{l c c r c c r}
				\toprule
				\textbf{Task} 
				& \textbf{Dataset}   
				& \textbf{Irregular} 
				& \textbf{Length} 
				& \textbf{Dimension} 
				& \textbf{Target} 
				& \textbf{Size} \\
				\midrule
				\multirow{6}{*}{LTS-C}  
				& \texttt{EC}        & \multirow{6}{*}{No}  & 1,751  & 3  & 4 & 524 \\
				& \texttt{EW}        &                      & 17,984 & 6  & 5 & 236 \\
				& \texttt{HB}        &                      & 405    & 61 & 2 & 409 \\
				& \texttt{MI}        &                      & 3,000  & 64 & 2 & 378 \\
				& \texttt{SCP1}      &                      & 896    & 6  & 2 & 561 \\
				& \texttt{SCP2}      &                      & 1,152  & 7  & 2 & 380 \\
				\midrule
				\multirow{3}{*}{LTS-R}
				& \texttt{HR}        & \multirow{3}{*}{No}  & \multirow{3}{*}{4,000} & \multirow{3}{*}{2} & \multirow{3}{*}{1} & 7,949 \\
				& \texttt{RR}        &                      &                        &                    &                    & 7,870 \\
				& \texttt{SpO$_2$}   &                      &                        &                    &                    & 7,949 \\
				\midrule
				PTS-C    & \texttt{HA}    & Yes & 50      & 12                  & 7 & 6,554 \\
				\midrule
				PTS-R    & \texttt{PDL}   & Yes & 50      & $24\times24$        & 2 & 4,000 \\
				\midrule
				\multirow{3}{*}{\shortstack[l]{TS-I \\ TS-E}}
				& \texttt{2DS}$_\alpha$ & \multirow{3}{*}{Yes} & 150 & 2 & 2 & 300 \\
				& \texttt{USH}                                &                      & 730 & 5 & 5 & 1,192 \\
				& \texttt{PHY}                            &                      & 72.16 & 37 & 37 & 8,000 \\
				\bottomrule
			\end{tabular}
		}
		\vspace{1pt} 
		\par 
		\footnotesize 		
		Note: \textbf{Irregular} specifies the presence of irregular sampling; \textbf{Dimension} and \textbf{Size} denote the feature count and total instances, while \textbf{Target} indicates the number of classes or continuous values to be predicted. \texttt{2DS}$_\alpha$ refers to the 2D Spiral dataset variants with $\alpha \in \{0.02, 0.2, 2\}$. Due to the variable-length nature of \texttt{PHY}, its reported length reflects the sample mean.
	\end{minipage}
\end{table}

\subsubsection{Tasks and Datasets \label{sec: datasets}}
We adopt a benchmark across six temporal modeling tasks spanning sixteen datasets to evaluate the three core capabilities defined in RQ1:
\begin{itemize}
	\item \textbf{Long-term Time Series Classification and Regression (LTS-C and LTS-R)} evaluate the capacity to model long-range dependencies by mapping global trajectories to categorical labels or continuous scalars. The classification setup incorporates six datasets \{\texttt{EC}, \texttt{EW}, \texttt{HB}, \texttt{MI}, \texttt{SCP1}, \texttt{SCP2}\} from the UEA\footnote{\url{https://timeseriesclassification.com}} archive~\cite{DBLP:journals/datamine/BagnallLBLK17} whereas the regression configuration utilizes three datasets \{\texttt{HR}, \texttt{RR}, \texttt{SpO}$_\texttt{2}$\} from the BIDMC\footnote{\url{https://physionet.org/content/bidmc/1.0.0/}} database~\cite{DBLP:journals/tbe/PimentelJCBWTC17} to infer patient vital signs from dense, 4,000-step physiological signals.
	
	\item \textbf{Per-point Time Series Classification and Regression (PTS-C and PTS-R)} assess the capability for fine-grained state tracking by decoding latent states at arbitrary, irregularly sampled timestamps. For point-wise classification, the \texttt{Human Activity} (\texttt{HA}) dataset\footnote{\url{https://doi.org/10.24432/C57G8X}}~\cite{DBLP:conf/nips/RubanovaCD19} is employed to classify localized actions at each time point, using 50 irregular observations per sequence. For point-wise regression, the synthetic \texttt{Pendulum} (\texttt{PDL}) dataset~\cite{DBLP:conf/icml/BeckerPGZTN19} is adopted to infer underlying kinematic angles from noisy image sequences spanning 50 irregularly sampled frames~\cite{DBLP:conf/icml/SchirmerELR22}.
	
	\item \textbf{Time Series Interpolation and Extrapolation (TS-I and TS-E)} evaluate trajectory reconstruction from sparse and noisy observations by recovering continuous-time dynamics under partial observation constraints. We adopt three datasets. The synthetic 2D spiral datasets (\texttt{2DS}$_\alpha$)~\cite{DBLP:conf/nips/ChenRWFSL23} generate 150-step spiral trajectories with shape variance controlled by $\alpha \in \{2, 0.2, 0.02\}$; models observe 30 randomly sampled points from the first half and reconstruct the full trajectory. The \texttt{USHCN} (\texttt{USH}) dataset\footnote{\url{https://data.ess-dive.lbl.gov/view/doi:10.3334/CDIAC/CLI.NDP019}}~\cite{osti_1394920} comprises daily climate measurement (5 variables) from 1,192 weather stations across the United States over a four-year period. Following~\cite{DBLP:conf/icml/SchirmerELR22}, we subsample 50\% of the time points and randomly drop 20\% of the remaining observations to obtain irregularly sampled sequences. The \texttt{Physionet} (\texttt{PHY}) dataset\footnote{\url{https://physionet.org/content/challenge-2012/1.0.0/}}~\cite{Silva2012PredictingIM} contains 8,000 multivariate clinical time series from intensive care unit patients, each recording 37 time-varying features over 48 hours. For both \texttt{USH} and \texttt{PHY}, we evaluate preprocessing, interpolation and extrapolation following the protocols in~\cite{park2025amortized}.
\end{itemize}
	
Table~\ref{tab:dataset_statistics} summarizes the structural characteristics of these datasets. Section S4 of the Supplementary Material provides comprehensive preprocessing details, including irregular sampling protocols and train/validation/test partitioning schemas.

\subsubsection{Baselines \label{sec: baselines}}
We benchmark LFormer against state-of-the-art baselines from three paradigms:
\begin{itemize}
	\item \textbf{Discrete-time Models.} This paradigm encompasses architectures engineered for uniformly indexed sequences without explicit physical time tracking. Representatives include GRU~\cite{DBLP:conf/emnlp/ChoMGBBSB14} and  Transformer~\cite{DBLP:conf/nips/VaswaniSPUJGKP17}. Modern linear recurrent and discrete state-space models are incorporated, including LRU~\cite{10.5555/3618408.3619518} and Mamba (S6)~\cite{DBLP:journals/corr/abs-2312-00752}. We also include baselines that augment discrete-time models with time-interval inputs or decay mechanisms: GRU-$\Delta_t$~\cite{DBLP:journals/corr/ChungGCB14}, GRU-D~\cite{2018Recurrent}, and RKN-$\Delta_t$~\cite{DBLP:conf/icml/BeckerPGZTN19}.
		
	\item \textbf{Solver-dependent Continuous-time Models.} This paradigm comprises architectures that parameterize latent state derivatives with neural networks and evaluate hidden trajectories via numerical differential solvers.
	This setup features vector-field equations ( NODE~\cite{DBLP:conf/nips/ChenRBD18}, R-ODE~\cite{DBLP:conf/nips/ChenRBD18}, O-ODE~\cite{DBLP:conf/nips/RubanovaCD19}, LSDE~\cite{DBLP:conf/nips/ZengGK23}), continuous path integrals (NCDE~\cite{DBLP:conf/nips/KidgerMFL20}, NRDE~\cite{DBLP:conf/icml/MorrillSKF21}, LogNCDE~\cite{DBLP:conf/icml/0001MQCLL24}),  solver-driven continuous-time attention ContiFormer~\cite{DBLP:conf/nips/ChenRWFSL23}, stochastic filtering CRU~\cite{DBLP:conf/icml/SchirmerELR22} and Bayesian extension GRU-ODE-B~\cite{DBLP:conf/nips/BrouwerSAM19}.
	
	\item\textbf{Solver-free Continuous-time Models:} This paradigm groups architectures that evaluate trajectory representations analytically or via exact linear discretization rules, bypassing iterative integration overhead. This setup incorporates closed-form liquid dynamics CfC~\cite{DBLP:journals/natmi/HasaniLALRTTR22}, truncated algebraic path signatures RFormer~\cite{10.5555/3737916.3741287}, multi-time embedding kernels (mTAND~\cite{shukla2021multitime}, and time-varying continuous state-space formulations (S5~\cite{smith2023simplified}, ACSSM~\cite{park2025amortized}).
\end{itemize}
	
Detailed description of all baselines is provided in Section S5 of the Supplementary Material.

\begin{table*}[t]
	\centering
	\caption{Long-term Time Series Classification Accuracy (\%, mean$_{\pm}$\textsubscript{std}) on Six UEA Datasets}
	\label{tab:ltsc_accuracy}
	\begin{minipage}{\textwidth}
		\resizebox{\linewidth}{!}{
			\begin{tabular}{l cccccccccc r}
				\toprule
				\multirow{2}[2]{*}{\textbf{Dataset}} 
				& \multicolumn{10}{c}{\textbf{Method}} 
				& \multicolumn{1}{c}{\multirow{2}[2]{*}{$\Delta_\uparrow$}} \\
				\cmidrule(lr){2-11}                 
				& LRU$^\dagger$~\cite{10.5555/3618408.3619518} 
				& S5$^\dagger$~\cite{smith2023simplified}
				& S6$^\dagger$~\cite{DBLP:journals/corr/abs-2312-00752}
				& Mamba$^\dagger$~\cite{DBLP:journals/corr/abs-2312-00752}
				& NCDE$^\dagger$~\cite{DBLP:conf/nips/KidgerMFL20}
				& NRDE$^\dagger$~\cite{DBLP:conf/icml/MorrillSKF21}
				& LogNCDE$^\dagger$~\cite{DBLP:conf/icml/0001MQCLL24}
				& Transformer$^\dagger$~\cite{DBLP:conf/nips/VaswaniSPUJGKP17}
				& RFormer$^\dagger$~\cite{10.5555/3737916.3741287}
				& LFormer (ours) &  \\
				\midrule
				\texttt{EC}   
				& 21.5$_{\pm}$\textsubscript{2.1} 
				& 24.1$_{\pm}$\textsubscript{4.3} 
				& 26.4$_{\pm}$\textsubscript{6.4} 
				& 27.9$_{\pm}$\textsubscript{4.5} 
				& 29.9$_{\pm}$\textsubscript{6.5} 
				& 25.3$_{\pm}$\textsubscript{1.8} 
				& 34.4$_{\pm}$\textsubscript{6.4} 
				& \underline{40.5}$_{\pm}$\textsubscript{6.3} 
				& 34.7$_{\pm}$\textsubscript{4.1} 
				& \textbf{45.3}$_{\pm}$\textsubscript{3.5} 
				& +11.9\% \\
				\texttt{EW}   
				& 87.8$_{\pm}$\textsubscript{2.8} 
				& 81.1$_{\pm}$\textsubscript{3.7} 
				& 85.0$_{\pm}$\textsubscript{16.1} 
				& 70.9$_{\pm}$\textsubscript{15.8} 
				& 75.0$_{\pm}$\textsubscript{3.9} 
				& 83.9$_{\pm}$\textsubscript{7.3} 
				& 85.6$_{\pm}$\textsubscript{5.1} 
				& OOM 
				& \textbf{90.3}$_{\pm}$\textsubscript{0.1} 
				& \underline{89.4}$_{\pm}$\textsubscript{3.2} 
				& -1.0\% \\
				\texttt{HB}   
				& \underline{78.4}$_{\pm}$\textsubscript{6.7} 
				& 77.7$_{\pm}$\textsubscript{5.5} 
				& 76.5$_{\pm}$\textsubscript{8.3} 
				& 76.2$_{\pm}$\textsubscript{3.8} 
				& 73.9$_{\pm}$\textsubscript{2.6} 
				& 72.9$_{\pm}$\textsubscript{4.8} 
				& 75.2$_{\pm}$\textsubscript{4.6} 
				& 70.5$_{\pm}$\textsubscript{0.1} 
				& 72.5$_{\pm}$\textsubscript{0.1} 
				& \textbf{81.6}$_{\pm}$\textsubscript{1.6} 
				& +4.1\% \\
				\texttt{MI}   
				& 48.4$_{\pm}$\textsubscript{5.0} 
				& 47.7$_{\pm}$\textsubscript{5.5} 
				& 51.3$_{\pm}$\textsubscript{4.7} 
				& 47.7$_{\pm}$\textsubscript{4.5} 
				& 49.5$_{\pm}$\textsubscript{2.8} 
				& 47.0$_{\pm}$\textsubscript{5.7} 
				& 53.7$_{\pm}$\textsubscript{5.3} 
				& 50.5$_{\pm}$\textsubscript{3.0} 
				& \underline{55.8}$_{\pm}$\textsubscript{6.6} 
				& \textbf{65.3}$_{\pm}$\textsubscript{4.1} 
				& +17.0\% \\
				\texttt{SCP1} 
				& 82.6$_{\pm}$\textsubscript{3.4} 
				& \textbf{89.9}$_{\pm}$\textsubscript{4.6} 
				& 82.8$_{\pm}$\textsubscript{2.7} 
				& 80.7$_{\pm}$\textsubscript{1.4} 
				& 79.8$_{\pm}$\textsubscript{5.6} 
				& 80.9$_{\pm}$\textsubscript{2.5} 
				& 83.1$_{\pm}$\textsubscript{2.8} 
				& 84.3$_{\pm}$\textsubscript{6.3} 
				& 81.2$_{\pm}$\textsubscript{2.8} 
				& \underline{84.5}$_{\pm}$\textsubscript{1.7} 
				& -6.0\% \\
				\texttt{SCP2} 
				& 51.2$_{\pm}$\textsubscript{3.6} 
				& 50.5$_{\pm}$\textsubscript{2.6} 
				& 49.9$_{\pm}$\textsubscript{9.5} 
				& 48.2$_{\pm}$\textsubscript{3.9} 
				& 53.0$_{\pm}$\textsubscript{2.8} 
				& \underline{53.7}$_{\pm}$\textsubscript{6.9} 
				& \underline{53.7}$_{\pm}$\textsubscript{4.1} 
				& 49.1$_{\pm}$\textsubscript{2.5} 
				& 52.3$_{\pm}$\textsubscript{3.7} 
				& \textbf{61.8}$_{\pm}$\textsubscript{3.0} 
				& +15.1\% \\
				\midrule
				\textbf{Avg. Acc.}$_\uparrow$ & 61.7 & 61.8 & 62.0 & 58.6 & 60.2 & 60.6 & 64.3 & 49.2 & \underline{64.5} & \textbf{71.3} & +10.5\% \\
				\textbf{Avg. Rank}$_\downarrow$ & 5.67 & 5.92 & 5.50 & 7.92 & 6.67 & 7.08 & \underline{3.92} & 6.50 & 4.50 & \textbf{1.33} & - \\
				\bottomrule
			\end{tabular}
		}
		\vspace{1pt}
		\par
		\footnotesize
		Note: The best results are highlighted in \textbf{bold}, and the second best are \underline{underlined}. $\Delta$ denotes the relative performance gain of LFormer over the best competitor. ``Avg. Acc.'' denotes the average accuracy over six datasets, while ``Avg. Rank'' denotes the average ranking across all datasets (fractional ranking is used for ties; OOM is ranked last). The arrows $\uparrow$ and $\downarrow$ indicate whether a higher or lower value is better. OOM indicates Out-Of-Memory. Results marked with $^\dagger$ are taken from~\cite{10.5555/3737916.3741287} under identical dataset splits and evaluation metrics.
	\end{minipage}
\end{table*}

\begin{table*}[t]
	\centering
	\caption{Long-term Time Series Regression MAE and RMSE (mean$_{\pm}$\textsubscript{std}) on Three BIDMC Datasets}
	\label{tab:ltser_results}
	\begin{minipage}{\textwidth}
		\resizebox{\linewidth}{!}{
			\begin{tabular}{l r ccccccccc c}
				\toprule
				\multirow{2}[2]{*}{\textbf{Dataset}} & \multirow{2}[2]{*}{\textbf{Metric}} 
				& \multicolumn{9}{c}{\textbf{Method}} 
				& \multirow{2}[2]{*}{$\Delta_\uparrow$} \\
				\cmidrule(lr){3-11}
				& 
				& GRU~\cite{DBLP:conf/emnlp/ChoMGBBSB14}
				& Mamba~\cite{DBLP:journals/corr/abs-2312-00752}
				& Transformer~\cite{DBLP:conf/nips/VaswaniSPUJGKP17}
				& R-ODE~\cite{DBLP:conf/nips/RubanovaCD19}
				& NCDE~\cite{DBLP:conf/nips/KidgerMFL20}
				& NRDE~\cite{DBLP:conf/icml/MorrillSKF21}
				& CfC~\cite{DBLP:journals/natmi/HasaniLALRTTR22}
				& RFormer~\cite{10.5555/3737916.3741287}
				& LFormer (ours) & \\
				\midrule
				\multirow{2}{*}{\texttt{HR}} 
				& MAE$_\downarrow$   
				& 9.64$_{\pm}$\textsubscript{0.61} & \underline{1.97}$_{\pm}$\textsubscript{0.39} & 9.82$_{\pm}$\textsubscript{0.19} & 12.11$_{\pm}$\textsubscript{0.00} & 9.20$_{\pm}$\textsubscript{0.08} & 2.74$_{\pm}$\textsubscript{0.13} & 9.77$_{\pm}$\textsubscript{0.31} & 2.37$_{\pm}$\textsubscript{0.17} & \textbf{1.03}$_{\pm}$\textsubscript{0.04} & +47.72\% \\
				& RMSE$_\downarrow$ 
				& 13.01$_{\pm}$\textsubscript{0.69} & \underline{2.75}$_{\pm}$\textsubscript{0.70} & 13.22$_{\pm}$\textsubscript{0.16} & 15.17$_{\pm}$\textsubscript{0.00} & 11.41$_{\pm}$\textsubscript{0.12} & 3.45$_{\pm}$\textsubscript{0.22} & 13.12$_{\pm}$\textsubscript{0.39} & 3.09$_{\pm}$\textsubscript{0.23} & \textbf{1.52}$_{\pm}$\textsubscript{0.07} & +44.73\% \\
				\midrule
				\multirow{2}{*}{\texttt{RR}} 
				& MAE$_\downarrow$   
				& 1.79$_{\pm}$\textsubscript{0.36} & \underline{0.93}$_{\pm}$\textsubscript{0.60} 
				& 1.45$_{\pm}$\textsubscript{0.22} & 1.85$_{\pm}$\textsubscript{0.00} & 2.27$_{\pm}$\textsubscript{0.03} & 1.20$_{\pm}$\textsubscript{0.04} & 2.15$_{\pm}$\textsubscript{0.12} & 1.06$_{\pm}$\textsubscript{0.04} & \textbf{0.74}$_{\pm}$\textsubscript{0.04} & +20.43\% \\
				& RMSE$_\downarrow$ 
				& 2.51$_{\pm}$\textsubscript{0.47} & \underline{1.32}$_{\pm}$\textsubscript{0.10} & 1.94$_{\pm}$\textsubscript{0.25} & 2.32$_{\pm}$\textsubscript{0.35} & 2.84$_{\pm}$\textsubscript{0.04} & 1.51$_{\pm}$\textsubscript{0.06} & 2.98$_{\pm}$\textsubscript{0.16} & 1.46$_{\pm}$\textsubscript{0.05} & \textbf{0.97}$_{\pm}$\textsubscript{0.02} & +26.52\% \\
				\midrule
				\multirow{2}{*}{\texttt{SpO$_2$}} 
				& MAE$_\downarrow$   
				& 2.46$_{\pm}$\textsubscript{0.03} & 1.08$_{\pm}$\textsubscript{0.49} & 2.45$_{\pm}$\textsubscript{0.03} & 2.71$_{\pm}$\textsubscript{0.00} & 2.46$_{\pm}$\textsubscript{0.12} & \underline{1.02}$_{\pm}$\textsubscript{0.09} & 2.46$_{\pm}$\textsubscript{0.03} & 1.38$_{\pm}$\textsubscript{0.15} & \textbf{0.29}$_{\pm}$\textsubscript{0.03} & +71.57\% \\
				& RMSE$_{\downarrow}$ 
				& 3.28$_{\pm}$\textsubscript{0.08} & 1.41$_{\pm}$\textsubscript{0.71} & 3.28$_{\pm}$\textsubscript{0.08} & 3.41$_{\pm}$\textsubscript{0.02} & 2.83$_{\pm}$\textsubscript{0.18} & \underline{1.28}$_{\pm}$\textsubscript{0.13} & 3.28$_{\pm}$\textsubscript{0.08} & 1.80$_{\pm}$\textsubscript{0.22} & \textbf{0.42}$_{\pm}$\textsubscript{0.02} & +67.19\% \\
				\bottomrule
			\end{tabular}
		}
		\vspace{1pt}
		\par
		\footnotesize
		Note: The best results are highlighted in \textbf{bold}, and the second best are \underline{underlined}. $\Delta$ denotes the relative performance gain of LFormer over the best competitor. The arrows $\uparrow$ and $\downarrow$ indicate whether a higher or lower value is better.
	\end{minipage}
\end{table*}
	
\subsubsection{Evaluation Metrics \label{sec: metrics}}
For classification tasks (LTS-C, PTS-C), we report Top-1 Accuracy. For LTS-R, we report Mean Absolute Error (MAE) and Root Mean Squared Error (RMSE); for PTS-R, we report Mean Squared Error (MSE). For TS-I and TS-E, we report MAE and RMSE on \texttt{2DS}$_\alpha$, and MSE on \texttt{PHY} and \texttt{USH}.

\subsubsection{Implementation Details \label{sec: implementation}}
We adopt a shared base configuration across the majority of benchmarks: $L=2$ Liquid Mixers, model dimension $d=64$, SwiGLU expansion $d_{ff}=176$, and $H=4$ attention heads. This default setup is used for all LTS-R tasks and four of the six LTS-C datasets (\texttt{EC}, \texttt{EW}, \texttt{HB}, \texttt{MI}). Task-specific adaptations are applied where required by the data modality or objective: \texttt{SCP1} and \texttt{SCP2} use $d=128$, $d_{ff}=352$, $H=1$, and $L=1$; PTS-R employs a convolutional embedder with $d=128$ and $d_{ff}=352$ to process $24{\times}24$ pixel inputs; PTS-C stacks $L=4$ layers for fine-grained per-point classification; TS-I/E on \texttt{2DS}$_\alpha$ adopts $d=32$, $L=1$, and a continuous time embedding following~\cite{DBLP:conf/nips/ChenRWFSL23};
\texttt{USH} and \texttt{PHY} use an MLP embedder with two hidden layers. Complete per-task specifications are provided in Section S6 of the Supplementary Material.

The optimizer is Adam with a learning rate of $10^{-3}$ (exceptions: $10^{-2}$ for PTS-R, TS-I/E on \texttt{2DS}$_\alpha$). Early stopping is applied when the validation metric does not improve for 20 consecutive epochs.To promote fair comparison, we calibrate baselines via grid search with the identical early stop strategy, or directly report published results when the evaluation protocol exactly matches ours. All experiments are run with five fixed random seeds $\{42, 43, 44, 45, 46\}$ and we report mean and standard deviation. All execution pipelines are deployed on a standardized workstation hardware environment equipped with an Intel Core i9-13900K CPU, 128 GB RAM, and a single NVIDIA RTX 4090 GPU.
	
\subsection{Temporal Modeling Capability Evaluation (RQ 1)\label{sec: main_results}}
We evaluate LFormer against state-of-the-art baselines along three core temporal modeling dimensions as follows.
\subsubsection{Long-range Dependency Modeling (LTS-C and LTS-R) \label{sec: long_term}}
As summarized in Table~\ref{tab:ltsc_accuracy}, LFormer achieves the highest average accuracy (71.3\%) and the top average rank (1.33) on the LTS-C benchmark, delivering the top performance on four out of six UEA datasets. On \texttt{EW} and \texttt{SCP1}, it ranks second behind RFormer and S5, respectively, indicating that signature-based or linear state-space inductive biases can still offer competitive advantages in certain classification regimes. Notably, LFormer secures a top-two finish across all evaluated datasets, whereas no competing baseline achieves a top-two ranking on more than three. On the extended-horizon \texttt{EW} dataset (17,984 steps), LFormer retains robust optimization stability and computational tractability, whereas the canonical Transformer incurs out-of-memory (OOM) exceptions. This result empirically confirms the scalability benefit of LGA's linear-complexity design. 

The empirical advantage of LFormer extends to the LTS-R benchmark. As detailed in Table~\ref{tab:ltser_results}, LFormer consistently outperforms all baselines across three evaluated BIDMC datasets in terms of both MAE and RMSE. Compared to the strongest baseline, LFormer achieves relative MAE reductions of 47.72\% for \texttt{HR}, 20.43\% for \texttt{RR}, and 71.57\% for \texttt{SpO}$_2$.
	
These evaluations address the first dimension of RQ1 concerning long-range dependency modeling. The prominent performance gains demonstrate the capacity of LGA's matrix-valued associative memory to capture extended temporal dependencies, supported by the sequence-level normalization that stabilizes optimization over thousands of time steps in full-sequence encoding scenarios.

\begin{table*}[t]
	\centering
	\caption{Time Series Interpolation and Extrapolation MSE (\%, mean$_{\pm}$\textsubscript{std}) on \texttt{PHY} and \texttt{USH} Datasets}
	\label{tab:physionet_ushcn_results}
	\begin{minipage}{\textwidth}
		\resizebox{\linewidth}{!}{
			\begin{tabular}{l c c cccccccccc r}
				\toprule
				\multirow{2}[2]{*}{\textbf{Task}} & \multirow{2}[2]{*}{\textbf{Dataset}} & \multirow{2}[2]{*}{\textbf{Metric}} & \multicolumn{10}{c}{\textbf{Method}} & \multirow{2}[2]{*}{$\Delta_{\uparrow}$} \\
				\cmidrule(lr){4-13}
				& & & mTAND$^\ddagger$~\cite{shukla2021multitime}
				& RKN-$\Delta_t^\ddagger$~\cite{DBLP:conf/icml/BeckerPGZTN19}
				& GRU-$\Delta_t^\ddagger$~\cite{DBLP:conf/emnlp/ChoMGBBSB14}
				& GRU-D$^\ddagger$~\cite{2018Recurrent}
				& NODE$^\ddagger$~\cite{DBLP:conf/nips/ChenRBD18}
				& R-ODE$^\ddagger$~\cite{DBLP:conf/nips/RubanovaCD19} 
				& GRU-ODE-B$^\ddagger$~\cite{DBLP:conf/nips/BrouwerSAM19}
				& CRU$^\ddagger$~\cite{DBLP:conf/icml/SchirmerELR22}
				& ACSSM$^\ddagger$~\cite{park2025amortized}
				& LFormer (ours) & \\
				\midrule
				\multirow{4}{*}{TS-I} 
				& \texttt{PHY}
				& MSE$_{\downarrow}$  & 0.208$_{\pm}$\textsubscript{0.025} & 0.186$_{\pm}$\textsubscript{0.030} & 0.271$_{\pm}$\textsubscript{0.057} & 0.338$_{\pm}$\textsubscript{0.027} & 0.212$_{\pm}$\textsubscript{0.027} & 0.236$_{\pm}$\textsubscript{0.009} & 0.521$_{\pm}$\textsubscript{0.038} & 0.182$_{\pm}$\textsubscript{0.091} & \underline{0.116}$_{\pm}$\textsubscript{0.011} & \textbf{0.040}$_{\pm}$\textsubscript{0.009} & +65.34\% \\
				\cmidrule(lr){2-14}
				& \texttt{USH}
				& MSE$_{\downarrow}$  & 1.766$_{\pm}$\textsubscript{0.009} & 0.009$_{\pm}$\textsubscript{0.002} & 0.090$_{\pm}$\textsubscript{0.059} & 0.944$_{\pm}$\textsubscript{0.011} & 1.798$_{\pm}$\textsubscript{0.009} & 0.831$_{\pm}$\textsubscript{0.008} & 0.841$_{\pm}$\textsubscript{0.142} & 0.016$_{\pm}$\textsubscript{0.006} & \underline{0.006}$_{\pm}$\textsubscript{0.001} & \textbf{0.001}$_{\pm}$\textsubscript{0.000} & +83.33\% \\
				\midrule
				\multirow{4}{*}{TS-E} 
				& \texttt{PHY}
				& MSE$_{\downarrow}$  & \textbf{0.340}$_{\pm}$\textsubscript{0.020} & 0.703$_{\pm}$\textsubscript{0.050} & 0.870$_{\pm}$\textsubscript{0.077} & 0.873$_{\pm}$\textsubscript{0.071} & 0.725$_{\pm}$\textsubscript{0.072} & 0.467$_{\pm}$\textsubscript{0.006} & 0.798$_{\pm}$\textsubscript{0.071} & 0.629$_{\pm}$\textsubscript{0.093} & 0.627$_{\pm}$\textsubscript{0.019} & \underline{0.372}$_{\pm}$\textsubscript{0.002} & -9.41\% \\
				\cmidrule(lr){2-14}
				& \texttt{USH} 
				& MSE$_{\downarrow}$  & 2.360$_{\pm}$\textsubscript{0.038} & 1.491$_{\pm}$\textsubscript{0.272} & 2.081$_{\pm}$\textsubscript{0.054} & 1.718$_{\pm}$\textsubscript{0.015} & 2.034$_{\pm}$\textsubscript{0.005} & 1.955$_{\pm}$\textsubscript{0.466} & 5.437$_{\pm}$\textsubscript{1.020} & 1.273$_{\pm}$\textsubscript{0.066} & \underline{0.941}$_{\pm}$\textsubscript{0.014} & \textbf{0.840}$_{\pm}$\textsubscript{0.018} & +10.73\% \\
				\bottomrule
			\end{tabular}
		}
		\vspace{1pt}
		\par
		\footnotesize
		Note: The best results are highlighted in \textbf{bold}, and the second best are \underline{underlined}. $\Delta$ denotes the relative performance gain of LFormer over the best-performing baseline. The arrows $\uparrow$ and $\downarrow$ indicate whether a higher or lower value is better. Results marked with $^\ddagger$ are taken from~\cite{park2025amortized}.
	\end{minipage}
\end{table*}

\begin{figure*}[t]
	\centering
	\begin{subfigure}{0.48\linewidth}
		\centering
		\includegraphics[width=\linewidth]{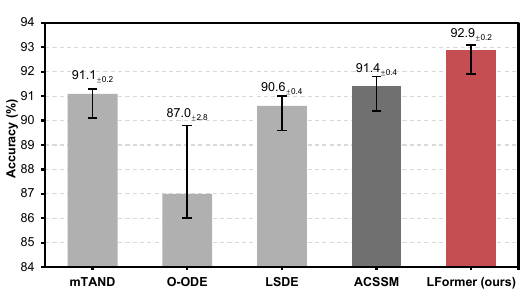}
		\caption{PTS-C Accuracy on the \texttt{HA} dataset}
		\label{fig:per_step_ha}
	\end{subfigure}
	\hfill
	\begin{subfigure}{0.48\linewidth}
		\centering
		\includegraphics[width=\linewidth]{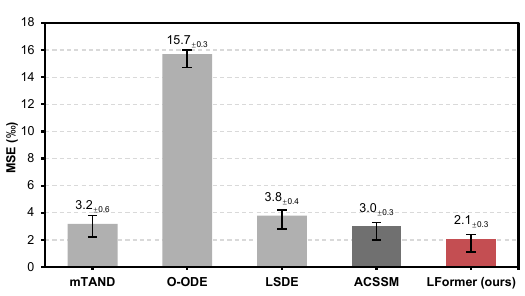}
		\caption{PTS-R MSE on the \texttt{PDL} dataset}
		\label{fig:per_step_pdl}
	\end{subfigure}
	\caption{Comparison of per-point time series classification and regression performance. Baseline results are reported from~\cite{park2025amortized}.}
	\label{fig:per_step_results}
\end{figure*}

\subsubsection{Fine-grained Dynamics Tracking (PTS-C and PTS-R) \label{sec: per_step}}
Figure~\ref{fig:per_step_results} reports the point-wise state tracking performance under the non-uniformly sampled benchmarks. LFormer delivers a classification accuracy of 92.9\% on the \texttt{HA} dataset (PTS-C) and an MSE of 2.1\textperthousand\ on the \texttt{PDL} dataset (PTS-R). 

On the \texttt{HA} dataset, LFormer yields a modest 1.6\% accuracy improvement over the state-of-the-art ACSSM, indicating that both methods achieve comparable performance when categorical decision boundaries dominate over high-fidelity continuous-time integration. 

On the \texttt{PDL} dataset, which demands dense tracking of continuous pendulum angles from strongly perturbed image sequences, LFormer achieves a substantial 31.2\% relative MSE reduction over ACSSM. Since both architectures employ the identical convolutional embedder, the pronounced performance gap isolates the algorithmic gain from visual feature extraction variations, attributing it to the underlying state transition mechanism. Crucially, our ablation analysis (Table~\ref{tab:ablation_comprehensive}) reveals that fixing $\mu=0.5$ (recovering the standard trapezoidal rule) outperforms the learnable variant. This indicates that under sparse and noisy conditions, unconstrained interpolation can be sensitive to local noise fluctuations. Conversely, the symmetric smoothing prior functions as an inductive low-pass constraint that suppresses noise propagation. Consequently, these gains stem not from any single component, but from the interplay of LGA's matrix memory, cumulative gate stabilization, and task-appropriate smoothing priors.

\begin{table*}[t]
	\centering
	\caption{Time Series Interpolation and Extrapolation MAE and RMSE (\%, mean$_{\pm}$\textsubscript{std}) on Three \texttt{2DS$_\alpha$} Datasets}
	\label{tab:spiral_results}
	\begin{minipage}{\textwidth}
		\resizebox{\linewidth}{!}{
			\begin{tabular}{l c r ccccccc c}
				\toprule
				\multirow{2}[2]{*}{\textbf{Task}} & \multirow{2}[2]{*}{{$\alpha$}} & \multirow{2}[2]{*}{\textbf{Metric}} & \multicolumn{7}{c}{\textbf{Method}} & \multirow{2}[2]{*}{$\Delta_{\uparrow}$} \\
				\cmidrule(lr){4-10}
				& & & Transformer$^\S$~\cite{DBLP:conf/nips/VaswaniSPUJGKP17} 
				& NODE$^\S$~\cite{DBLP:conf/nips/ChenRBD18}
				& R-ODE$^\S$~\cite{DBLP:conf/nips/ChenRBD18}
				& O-ODE$^\S$~\cite{DBLP:conf/nips/RubanovaCD19}
				& NCDE$^\S$~\cite{DBLP:conf/nips/KidgerMFL20}
				& ContiFormer$^\S$~\cite{DBLP:conf/nips/ChenRWFSL23}
				& LFormer (ours) & \\
				\midrule
				\multirow{6}{*}{TS-I} 
				& \multirow{2}{*}{2} 
				& MAE$_{\downarrow}$  & 2.33$_{\pm}$\textsubscript{0.60} & 2.49$_{\pm}$\textsubscript{0.30} & 2.88$_{\pm}$\textsubscript{0.54} & 4.12$_{\pm}$\textsubscript{1.00} & 2.96$_{\pm}$\textsubscript{0.23} & \underline{0.41}$_{\pm}$\textsubscript{0.07} & \textbf{0.40}$_{\pm}$\textsubscript{0.06} & +2.44\% \\
				& & RMSE$_{\downarrow}$ & 3.75$_{\pm}$\textsubscript{0.50} & 2.10$_{\pm}$\textsubscript{0.21} & 3.04$_{\pm}$\textsubscript{0.48} & 4.49$_{\pm}$\textsubscript{1.26} & 2.92$_{\pm}$\textsubscript{0.32} & \underline{0.45}$_{\pm}$\textsubscript{0.11} & \textbf{0.43}$_{\pm}$\textsubscript{0.09} & +4.44\% \\
				\cmidrule(lr){2-11}
				& \multirow{2}{*}{0.2} 
				& MAE$_{\downarrow}$  & 1.75$_{\pm}$\textsubscript{0.18} & 1.58$_{\pm}$\textsubscript{0.12} & 3.16$_{\pm}$\textsubscript{0.43} & 4.73$_{\pm}$\textsubscript{1.21} & 4.58$_{\pm}$\textsubscript{0.93} & \underline{0.42}$_{\pm}$\textsubscript{0.01} & \textbf{0.33}$_{\pm}$\textsubscript{0.02} & +21.43\% \\
				& & RMSE$_{\downarrow}$ & 2.30$_{\pm}$\textsubscript{0.51} & 1.88$_{\pm}$\textsubscript{0.09} & 3.34$_{\pm}$\textsubscript{0.10} & 4.82$_{\pm}$\textsubscript{1.97} & 4.60$_{\pm}$\textsubscript{1.08} & \underline{0.42}$_{\pm}$\textsubscript{0.04} & \textbf{0.34}$_{\pm}$\textsubscript{0.02} & +19.05\% \\
				\cmidrule(lr){2-11}
				& \multirow{2}{*}{0.02} 
				& MAE$_{\downarrow}$  & 1.42$_{\pm}$\textsubscript{0.14} & 1.90$_{\pm}$\textsubscript{0.17} & 1.95$_{\pm}$\textsubscript{0.26} & 4.25$_{\pm}$\textsubscript{1.48} & 5.12$_{\pm}$\textsubscript{0.47} & \underline{0.53}$_{\pm}$\textsubscript{0.07} & \textbf{0.33}$_{\pm}$\textsubscript{0.09} & +37.74\% \\
				& & RMSE$_{\downarrow}$ & 1.37$_{\pm}$\textsubscript{0.15} & 1.90$_{\pm}$\textsubscript{0.17} & 2.09$_{\pm}$\textsubscript{0.22} & 4.53$_{\pm}$\textsubscript{1.34} & 4.80$_{\pm}$\textsubscript{0.50} & \underline{0.50}$_{\pm}$\textsubscript{0.06} & \textbf{0.31}$_{\pm}$\textsubscript{0.06} & +38.00\% \\
				\midrule
				\multirow{6}{*}{TS-E} 
				& \multirow{2}{*}{2} 
				& MAE$_{\downarrow}$  & 4.48$_{\pm}$\textsubscript{0.82} & 4.84$_{\pm}$\textsubscript{0.43} & 4.46$_{\pm}$\textsubscript{0.60} & 4.29$_{\pm}$\textsubscript{1.34} & 3.32$_{\pm}$\textsubscript{0.85} & \underline{0.69}$_{\pm}$\textsubscript{0.07} & \textbf{0.53}$_{\pm}$\textsubscript{0.06} & +23.19\% \\
				& & RMSE$_{\downarrow}$ & 5.36$_{\pm}$\textsubscript{0.97} & 6.02$_{\pm}$\textsubscript{0.58} & 5.70$_{\pm}$\textsubscript{0.57} & 4.68$_{\pm}$\textsubscript{1.70} & 3.44$_{\pm}$\textsubscript{1.06} & \underline{0.75}$_{\pm}$\textsubscript{0.07} & \textbf{0.52}$_{\pm}$\textsubscript{0.06} & +30.67\% \\
				\cmidrule(lr){2-11}
				& \multirow{2}{*}{0.2} 
				& MAE$_{\downarrow}$  & 2.79$_{\pm}$\textsubscript{1.11} & 3.35$_{\pm}$\textsubscript{0.43} & 4.02$_{\pm}$\textsubscript{0.59} & 4.72$_{\pm}$\textsubscript{1.96} & 3.37$_{\pm}$\textsubscript{0.78} & \underline{0.69}$_{\pm}$\textsubscript{0.09} & \textbf{0.44}$_{\pm}$\textsubscript{0.04} & +36.23\% \\
				& & RMSE$_{\downarrow}$ & 2.96$_{\pm}$\textsubscript{1.46} & 4.38$_{\pm}$\textsubscript{0.35} & 4.78$_{\pm}$\textsubscript{0.15} & 5.03$_{\pm}$\textsubscript{1.97} & 3.81$_{\pm}$\textsubscript{1.02} & \underline{0.74}$_{\pm}$\textsubscript{0.10} & \textbf{0.45}$_{\pm}$\textsubscript{0.06} & +39.19\% \\
				\cmidrule(lr){2-11}
				& \multirow{2}{*}{0.02} 
				& MAE$_{\downarrow}$  & 1.49$_{\pm}$\textsubscript{0.12} & 2.02$_{\pm}$\textsubscript{0.07} & 1.52$_{\pm}$\textsubscript{0.09} & 2.83$_{\pm}$\textsubscript{0.88} & 2.38$_{\pm}$\textsubscript{0.04} & \underline{0.65}$_{\pm}$\textsubscript{0.08} & \textbf{0.34}$_{\pm}$\textsubscript{0.08} & +47.69\% \\
				& & RMSE$_{\downarrow}$ & 1.37$_{\pm}$\textsubscript{0.11} & 2.07$_{\pm}$\textsubscript{0.02} & 1.59$_{\pm}$\textsubscript{0.05} & 2.69$_{\pm}$\textsubscript{0.82} & 2.24$_{\pm}$\textsubscript{0.03} & \underline{0.64}$_{\pm}$\textsubscript{0.10} & \textbf{0.32}$_{\pm}$\textsubscript{0.06} & +50.00\% \\
				\bottomrule
			\end{tabular}
		}
		\vspace{1pt}
		\par
		\footnotesize
		Note: The best results are highlighted in \textbf{bold}, and the second best are \underline{underlined}. $\Delta$ denotes the relative performance gain of LFormer over the best-performing baseline. The arrows $\uparrow$ and $\downarrow$ indicate whether a higher or lower value is better. Results marked with $^\S$ are taken from~\cite{DBLP:conf/nips/ChenRWFSL23} under identical evaluation protocols.
	\end{minipage}
\end{table*}

\begin{figure*}[t]
	\centering
	\includegraphics[width=\linewidth]{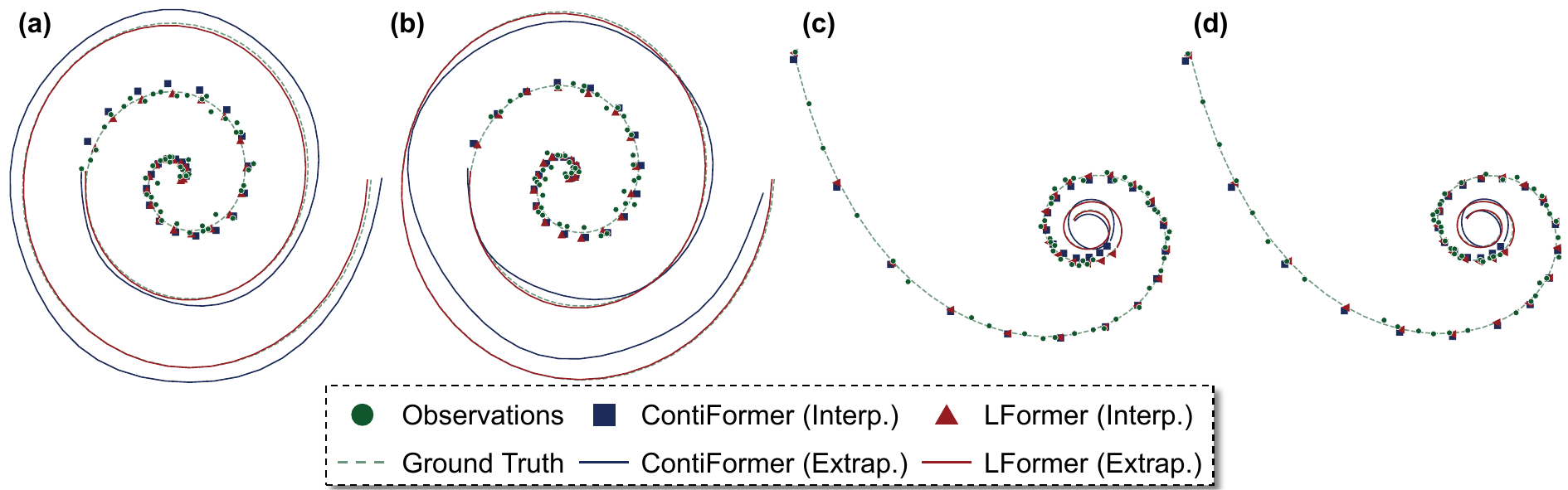}
	\caption{Qualitative trajectory reconstruction comparison on the \texttt{2DS$_{\alpha=0.02}$} benchmark. Panels (a)-(b) and panels (c)-(d) show standard Archimedean spirals and nonlinear asymptotic spirals, respectively. The initial temporal half contains 30 randomly sampled irregular observations (green dots) for interpolation, while the terminal half is left unobserved for extrapolation capacity. LFormer (red) closely tracks the continuous ground-truth trajectory (dashed green) during the unobserved extrapolation phase, whereas ContiFormer (blue) exhibits structural divergence.}
	\label{fig:spiral_case_study}
\end{figure*}
	
\subsubsection{Trajectory Reconstruction (TS-I and TS-E) \label{sec: continuous_time}}
We evaluate trajectory reconstruction on two scenarios: real-world irregular records (\texttt{PHY} and \texttt{USH}) and controlled synthetic spirals (\texttt{2DS}$_\alpha$) under varying signal-to-noise (SNR) ratios.

\paragraph{Real-world Irregular Data} As reported in Table~\ref{tab:physionet_ushcn_results}, LFormer achieves substantial gains on TS-I for both \texttt{PHY} and \texttt{USH}, reducing MSE by 65.3\% and 83.3\% respectively over the strongest baseline (ACSSM). These margins indicate that coupling state transitions with observed time intervals provides an effective continuous-time inductive bias for interpolation. Within this non-causal setting, LGA leverages bidirectional temporal gaps to modulate the associative memory, circumventing the drift and error accumulation that solver-dependent methods can suffer under sparse conditions.

The TS-E results reveal an architectural boundary: LFormer reduces MSE by 10.73\% over ACSSM on \texttt{USH}, yet trails mTAND by 9.41\% on \texttt{PHY}. This divergence reflects that LGA's recurrent memory excels on the smooth dynamics of \texttt{USH}, whereas mTAND's query-based attention pooling is better suited to the non-stationary signals of \texttt{PHY}. 

The ablation results support this: on \texttt{USH}, removing the decoupled multi-head gating or MHLGA incurs the largest performance drop (MSE rises to 0.890 and 0.884; Table~\ref{tab:ablation_comprehensive}). This highlights that head-wise specialization across distinct temporal scales is central to capturing the multi-frequency structure of climate variables.

In contrast, on \texttt{PHY}, all ablated variants stay within a narrow MSE range (0.370--0.379), with several configurations matching or slightly surpassing the full model (0.372). This insensitivity indicates that the extrapolation bottleneck of LFormer on \texttt{PHY} stems from a system-level constraint. When handling highly non-stationary signals, extrapolation demands acute sensitivity to localized boundary fluctuations. mTAND directly queries sparse past evidence at arbitrary future timestamps, avoiding the forward drift inherent in the recurrent state-transition of LGA.

This trade-off yields a concrete direction for future architectures: integrating LGA with explicit multi-scale timestamp query networks. Such a hybrid approach may effectively reconcile global trajectory consistency (evidenced by TS-E results on \texttt{USH} and real-world TS-I benchmarks) and acute localized temporal awareness in unobserved regimes.

\paragraph{Synthetic Irregular Trajectories}
As detailed in Table~\ref{tab:spiral_results}, LFormer achieves the best MAE and RMSE on all three \texttt{2DS$_\alpha$} variants in both TS-I and TS-E. The margin over ContiFormer, the strongest baseline, widens as $\alpha$ decreases: from marginal 2.44\% (TS-I) MAE reductions at $\alpha=2$ to pronounced 37.7\% (TS-I) and 47.7\% (TS-E) MAE reductions at $\alpha=0.02$. Note that both models employ the same continuous time embedding on this benchmark~\cite{DBLP:conf/nips/ChenRWFSL23}; the widening gap therefore reflects differences in the underlying state transition mechanisms.

On the \texttt{2DS}$_\alpha$ benchmark, $\alpha$ controls the geometric variance of generated trajectories under a constant observational noise. At $\alpha=2$, the signal dominates the noise, and all models can reliably recover the underlying trajectory. As $\alpha$ decreases to $0.02$, the signal weakens and the trajectories become noise-dominated. Under these low SNR conditions, solver-dependent methods can suffer from high-frequency overshooting when fitting sparse and noisy observations (as shown in Fig.~\ref{fig:spiral_case_study}).

We attribute this to the bounded liquid gate: as a convex combination within $(0,1)$, the formulation limits the propagation of high-frequency noise across time steps. Rather than acting as a spectral filter, this formulation introduces an inductive smoothness prior under strong noise, consistent with the low-pass characteristics of exponential moving averages.

\begin{table*}[t]
	\centering
	\caption{Comprehensive Ablation Study across Real-world and Synthetic Benchmarks (mean$_{\pm}$\textsubscript{std})}
	\label{tab:ablation_comprehensive}
	\begin{minipage}{\textwidth}
		\resizebox{\linewidth}{!}{
			\begin{tabular}{l c r c ccc ccc c}
				\toprule
				\multirow{2}[2]{*}{\textbf{Task}} 
				& \multirow{2}[2]{*}{\textbf{Dataset}} 
				& \multirow{2}[2]{*}{\textbf{Metric}} 
				& \textbf{Full} 
				& \multicolumn{3}{c}{\textbf{Macroscopic Architecture}} 
				& \multicolumn{3}{c}{\textbf{Liquid Gate Dynamics}} 
				& \multicolumn{1}{c}{\textbf{Stabilization}} \\
				\cmidrule(lr){4-4} \cmidrule(lr){5-7} \cmidrule(lr){8-10} \cmidrule(lr){11-11}
				& & 
				& LFormer
				& w/o MHLGA 
				& w/o SwiGLU 
				& w/o Output Gate
				& w/o Decoupled $\mu$ 
				& w/o Learnable $\mu$ 
				& w/o Input-aware $\mathbf{g}$ 
				& w/o Normalized $\mathbf{g}$ \\
				\midrule
				
				\multirow{3}{*}{LTS-R} 
				& \texttt{HR} & \multirow{3}{*}{RMSE$_\downarrow$} 
				& 1.52$_{\pm}$\textsubscript{0.07} & 1.93$_{\pm}$\textsubscript{0.20} & 1.55$_{\pm}$\textsubscript{0.15} & 1.57$_{\pm}$\textsubscript{0.20} & \textbf{1.50}$_{\pm}$\textsubscript{0.11} & 1.60$_{\pm}$\textsubscript{0.17} & 1.68$_{\pm}$\textsubscript{0.10} & 1.52$_{\pm}$\textsubscript{0.06} \\
				& \texttt{RR} & 
				& \textbf{0.97}$_{\pm}$\textsubscript{0.02} & 1.16$_{\pm}$\textsubscript{0.04} & 1.04$_{\pm}$\textsubscript{0.05} & 1.06$_{\pm}$\textsubscript{0.09} & 1.06$_{\pm}$\textsubscript{0.04} & 1.05$_{\pm}$\textsubscript{0.07} & 1.19$_{\pm}$\textsubscript{0.02} & 1.14$_{\pm}$\textsubscript{0.06} \\
				& \texttt{SpO$_2$} & 
				& 0.42$_{\pm}$\textsubscript{0.02} & 0.48$_{\pm}$\textsubscript{0.02} & \textbf{0.41}$_{\pm}$\textsubscript{0.01} & 0.42$_{\pm}$\textsubscript{0.03} & 0.44$_{\pm}$\textsubscript{0.03} & 0.42$_{\pm}$\textsubscript{0.01} & 0.48$_{\pm}$\textsubscript{0.05} & 0.45$_{\pm}$\textsubscript{0.02} \\
				\cmidrule(lr){5-11}
				\multicolumn{3}{l}{\emph{Average $\Delta$ vs. Full Model}}
				& - & -20.28\% & -2.27\% & -4.19\% & -4.24\% & -4.50\% & -15.83\% & -8.22\% \\
				\midrule
				
				\multirow{6}{*}{LTS-C} 
				& \texttt{EC} & \multirow{6}{*}{Acc.$_\uparrow$ (\%)} 
				& \textbf{45.30}$_{\pm}$\textsubscript{3.50} & 39.70$_{\pm}$\textsubscript{2.60} & 41.80$_{\pm}$\textsubscript{5.40} & 42.00$_{\pm}$\textsubscript{5.20} & 40.30$_{\pm}$\textsubscript{4.40} & 40.30$_{\pm}$\textsubscript{4.40} & 41.00$_{\pm}$\textsubscript{4.10} & 39.70$_{\pm}$\textsubscript{4.60} \\
				& \texttt{EW} & 
				& \textbf{89.40}$_{\pm}$\textsubscript{3.20} & 83.30$_{\pm}$\textsubscript{1.80} & 87.20$_{\pm}$\textsubscript{2.20} & 86.70$_{\pm}$\textsubscript{4.10} & 88.90$_{\pm}$\textsubscript{4.60} & 88.90$_{\pm}$\textsubscript{4.60} & 88.90$_{\pm}$\textsubscript{1.80} & 88.30$_{\pm}$\textsubscript{4.40} \\
				& \texttt{HB} & 
				& 81.60$_{\pm}$\textsubscript{1.60} & 80.30$_{\pm}$\textsubscript{1.60} & 80.00$_{\pm}$\textsubscript{1.60} & 81.00$_{\pm}$\textsubscript{2.10} & 80.30$_{\pm}$\textsubscript{0.60} & 80.30$_{\pm}$\textsubscript{0.60} & 80.30$_{\pm}$\textsubscript{3.30} & \textbf{82.30}$_{\pm}$\textsubscript{2.30} \\
				& \texttt{MI} & 
				& \textbf{65.30}$_{\pm}$\textsubscript{7.00} & 63.50$_{\pm}$\textsubscript{5.80} & 61.40$_{\pm}$\textsubscript{6.30} & 59.60$_{\pm}$\textsubscript{6.40} & 60.40$_{\pm}$\textsubscript{6.00} & 60.40$_{\pm}$\textsubscript{6.00} & 61.80$_{\pm}$\textsubscript{5.50} & 60.00$_{\pm}$\textsubscript{4.10} \\
				& \texttt{SCP1} & 
				& 84.50$_{\pm}$\textsubscript{1.70} & \textbf{89.20}$_{\pm}$\textsubscript{4.00} & 84.90$_{\pm}$\textsubscript{3.70} & 85.20$_{\pm}$\textsubscript{3.50} & 84.00$_{\pm}$\textsubscript{3.80} & 84.50$_{\pm}$\textsubscript{3.70} & 86.40$_{\pm}$\textsubscript{3.50} & 86.40$_{\pm}$\textsubscript{3.50} \\
				& \texttt{SCP2} & 
				& 61.80$_{\pm}$\textsubscript{3.00} & 63.20$_{\pm}$\textsubscript{3.30} & 62.80$_{\pm}$\textsubscript{2.60} & \textbf{66.00}$_{\pm}$\textsubscript{2.40} & 62.10$_{\pm}$\textsubscript{2.90} & 61.80$_{\pm}$\textsubscript{3.40} & 63.50$_{\pm}$\textsubscript{3.60} & 61.80$_{\pm}$\textsubscript{3.40} \\
				\cmidrule(lr){5-11}
				\multicolumn{3}{l}{\emph{Average $\Delta$ vs. Full Model}}
				&- & -2.62\% & -2.67\% & -2.02\% & -3.47\% & -3.45\% & -2.00\% & -3.10\% \\
				\midrule
				
				\multirow{5}{*}{TS-I} 
				& \texttt{PHY} & \multirow{2}{*}{MSE$_\downarrow$ (\%)} 
				& 0.040$_{\pm}$\textsubscript{0.015} & 0.070$_{\pm}$\textsubscript{0.028} & \textbf{0.033}$_{\pm}$\textsubscript{0.005} & 0.042$_{\pm}$\textsubscript{0.003} & 0.044$_{\pm}$\textsubscript{0.004} & 0.047$_{\pm}$\textsubscript{0.018} & 0.047$_{\pm}$\textsubscript{0.007} & 0.132$_{\pm}$\textsubscript{0.053} \\
				& \texttt{USH} & 
				& \textbf{0.001}$_{\pm}$\textsubscript{0.000} & 0.003$_{\pm}$\textsubscript{0.002} & \textbf{0.001}$_{\pm}$\textsubscript{0.000} & \textbf{0.001}$_{\pm}$\textsubscript{0.000} & \textbf{0.001}$_{\pm}$\textsubscript{0.000} & \textbf{0.001}$_{\pm}$\textsubscript{0.000} & \textbf{0.001}$_{\pm}$\textsubscript{0.000} & inf$_{\pm}$\textsubscript{nan} \\
				& \texttt{2DS}$_{\alpha=2}$ & \multirow{3}{*}{RMSE$_\downarrow$ (\%)} 
				& 0.43$_{\pm}$\textsubscript{0.09} & 0.48$_{\pm}$\textsubscript{0.07} & 0.50$_{\pm}$\textsubscript{0.19} & 0.45$_{\pm}$\textsubscript{0.07} & \textbf{0.42}$_{\pm}$\textsubscript{0.09} & 0.46$_{\pm}$\textsubscript{0.10} & 0.52$_{\pm}$\textsubscript{0.18} & 0.45$_{\pm}$\textsubscript{0.02} \\
				& \texttt{2DS}$_{\alpha=0.2}$ & 
				& \textbf{0.34}$_{\pm}$\textsubscript{0.02} & 0.45$_{\pm}$\textsubscript{0.09} & 0.38$_{\pm}$\textsubscript{0.05} & 0.38$_{\pm}$\textsubscript{0.04} & 0.39$_{\pm}$\textsubscript{0.07} & 0.37$_{\pm}$\textsubscript{0.05} & 0.39$_{\pm}$\textsubscript{0.10} & 0.39$_{\pm}$\textsubscript{0.07} \\
				& \texttt{2DS}$_{\alpha=0.02}$ & 
				& 0.31$_{\pm}$\textsubscript{0.06} & 0.42$_{\pm}$\textsubscript{0.04} & 0.37$_{\pm}$\textsubscript{0.08} & 0.40$_{\pm}$\textsubscript{0.06} & 0.31$_{\pm}$\textsubscript{0.04} & \textbf{0.30}$_{\pm}$\textsubscript{0.04} & 0.32$_{\pm}$\textsubscript{0.06} & 0.34$_{\pm}$\textsubscript{0.04} \\
				\cmidrule(lr){5-11}
				\multicolumn{3}{l}{\emph{Average $\Delta$ vs. Full Model}}
				& - & -41.22\% & -11.75\% & -11.41\% & -5.43\% & -6.18\% & -12.30\% & -32.88\% \\
				\midrule
				
				\multirow{5}{*}{TS-E} 
				& \texttt{PHY} & \multirow{2}{*}{MSE$_\downarrow$ (\%)} 
				& 0.372$_{\pm}$\textsubscript{0.002} & 0.379$_{\pm}$\textsubscript{0.002} & \textbf{0.370}$_{\pm}$\textsubscript{0.003} & 0.375$_{\pm}$\textsubscript{0.005} & \textbf{0.370}$_{\pm}$\textsubscript{0.005} & 0.374$_{\pm}$\textsubscript{0.006} & 0.373$_{\pm}$\textsubscript{0.004} & \textbf{0.370}$_{\pm}$\textsubscript{0.002} \\
				& \texttt{USH} & 
				& 0.840$_{\pm}$\textsubscript{0.018} & 0.884$_{\pm}$\textsubscript{0.007} & \textbf{0.835}$_{\pm}$\textsubscript{0.010} & 0.836$_{\pm}$\textsubscript{0.005} & 0.890$_{\pm}$\textsubscript{0.116} & 0.837$_{\pm}$\textsubscript{0.014} & 0.842$_{\pm}$\textsubscript{0.004} & inf$_{\pm}$\textsubscript{nan}  \\
				& \texttt{2DS}$_{\alpha=2}$ & \multirow{3}{*}{RMSE$_\downarrow$ (\%)} 
				& 0.52$_{\pm}$\textsubscript{0.06} & 0.56$_{\pm}$\textsubscript{0.08} & 0.51$_{\pm}$\textsubscript{0.08} & 0.55$_{\pm}$\textsubscript{0.10} & \textbf{0.49}$_{\pm}$\textsubscript{0.05} & 0.52$_{\pm}$\textsubscript{0.08} & 0.57$_{\pm}$\textsubscript{0.08} & 0.51$_{\pm}$\textsubscript{0.08} \\
				& \texttt{2DS}$_{\alpha=0.2}$ & 
				& 0.45$_{\pm}$\textsubscript{0.06} & 0.50$_{\pm}$\textsubscript{0.02} & 0.47$_{\pm}$\textsubscript{0.04} & 0.49$_{\pm}$\textsubscript{0.09} & 0.47$_{\pm}$\textsubscript{0.04} & 0.47$_{\pm}$\textsubscript{0.07} & \textbf{0.44}$_{\pm}$\textsubscript{0.07} & 0.46$_{\pm}$\textsubscript{0.06} \\
				& \texttt{2DS}$_{\alpha=0.02}$ & 
				& \textbf{0.32}$_{\pm}$\textsubscript{0.06} & 0.43$_{\pm}$\textsubscript{0.01} & 0.43$_{\pm}$\textsubscript{0.10} & 0.50$_{\pm}$\textsubscript{0.21} & 0.33$_{\pm}$\textsubscript{0.05} & 0.33$_{\pm}$\textsubscript{0.04} & 0.35$_{\pm}$\textsubscript{0.06} & 0.37$_{\pm}$\textsubscript{0.05} \\
				\cmidrule(lr){5-11}
				\multicolumn{3}{l}{\emph{Average $\Delta$ vs. Full Model}}
				& - & -11.64\% & -7.14\% & -14.65\% & -1.33\% & -1.68\% & -3.38\% & -6.44\% \\
				\midrule
				
				PTS-C & \texttt{HA} & Acc.$_\uparrow$ (\%) 
				& \textbf{92.87}$_{\pm}$\textsubscript{0.16} & 89.94$_{\pm}$\textsubscript{0.76} & 90.87$_{\pm}$\textsubscript{0.49} & 91.72$_{\pm}$\textsubscript{0.35} & 91.83$_{\pm}$\textsubscript{0.17} & 91.81$_{\pm}$\textsubscript{0.17} & 91.80$_{\pm}$\textsubscript{0.38} & 20.49$_{\pm}$\textsubscript{0.00} \\
				\cmidrule(lr){5-11}
				\multicolumn{3}{l}{\emph{$\Delta$ vs. Full Model}}
				& - & -3.15\% & -2.15\% & -1.24\% & -1.12\% & -1.14\% & -1.15\% & Collapse \\
				\midrule
				
				PTS-R & \texttt{PDL} & MSE$_\downarrow$ (\textperthousand) 
				& 2.05$_{\pm}$\textsubscript{0.25} 
				& inf$_{\pm}$\textsubscript{nan} 
				& inf$_{\pm}$\textsubscript{nan} 
				& inf$_{\pm}$\textsubscript{nan} 
				& inf$_{\pm}$\textsubscript{nan} 
				& \textbf{1.56}$_{\pm}$\textsubscript{0.36} 
				& 2.85$_{\pm}$\textsubscript{0.26} 
				& inf$_{\pm}$\textsubscript{nan}  \\
				\cmidrule(lr){5-11}
				\multicolumn{3}{l}{\emph{$\Delta$ vs. Full Model}}
				& - 
				& -
				& -
				& -
				& -
				& +23.90\% 
				& -39.02\% 
				& - \\
				\bottomrule
			\end{tabular}
		}
		\vspace{1pt}
		\par
		\footnotesize
		Note: \texttt{2DS} denotes the Spiral 2D dataset evaluated under different $\alpha$ values. The arrows $\uparrow$ and $\downarrow$ indicate whether a higher or lower value is better. \textbf{$\Delta$ vs. Full Model} denotes the average relative percentage difference in performance compared to the full LFormer model. \textbf{inf$_{\pm}$\textsubscript{nan}}  indicates the training process collapsed due to ``NaN'' or ``Inf'' values. \textbf{Collapse} denotes performance degradation by orders of magnitude.
	\end{minipage}
\end{table*}

\subsection{Component Analysis \label{sec: ablation}} 
To evaluate the contributions of individual components in LFormer, we conduct an ablation study across all benchmark datasets. Taking the complete LFormer as the baseline (\emph{Full}), variants are organized into three functional groups:
\begin{itemize}
	\item \textbf{Macroscopic Architecture} evaluates structural configurations via three variants
	\begin{itemize}
		\item \emph{w/o MHLGA} substitutes the MHLGA module with standard multi-head self-attention~\cite{DBLP:conf/nips/VaswaniSPUJGKP17}.
		\item \emph{w/o SwiGLU} replaces the SwiGLU channel mixer with a standard Feed-Forward Network~\cite{DBLP:conf/nips/VaswaniSPUJGKP17}.
		\item \emph{w/o Output Gate} removes the output gate.
	\end{itemize}
		
	\item \textbf{Liquid Gate Dynamics} dissect continuous-time transition equations via three configurations: 
	\begin{itemize}
		\item \emph{w/o Decoupled $\mu$} shares a global $\mu$ across all multi-head subspaces instead of learning head-specific coefficients independently.
		\item  \emph{w/o Learnable $\mu$} fixes the learnable interpolation weight as $0.5$, recovering the trapezoidal rule.
		\item \emph{w/o Input-aware $\mathbf{g}$} replaces input-driven gating with a deterministic time-decay.
	\end{itemize}
		
	\item \textbf{Stabilization} assesses optimization stability via \emph{w/o Normalized $\mathbf{g}$}, which ablates the sequence-level normalization from the cumulative gate computation.
	\end{itemize}

Empirical evaluation metrics for these structural configurations are compiled in Table~\ref{tab:ablation_comprehensive}, with specialized architectural analysis detailed below.

\subsubsection{Macroscopic Architecture}
\begin{itemize}
	\item \textbf{MHLGA is critical for regression tasks under irregular sampling.} On classification tasks, \emph{w/o MHLGA} and \emph{w/o SwiGLU} result in comparable, minor degradation (LTS-C: -2.62\% vs.\ -2.67\%; PTS-C: -3.15\% vs.\ -2.15\%). This suggests coarse-grained categorical decisions are relatively robust to the specific choices of temporal encoders or channel mixers. On regression tasks, this gap widens substantially: \emph{w/o MHLGA} degrades performance far more severely than \emph{w/o SwiGLU} on both LTS-R (-20.28\% vs.\ -2.27\%) and TS-I (-41.22\% vs.\ -11.75\%) tasks. Furthermore, on the highly sparse \texttt{PDL} dataset, all macroscopic variants completely collapse, underscoring the necessity of the full architectural design.
	
	\item \textbf{Output gating supports extrapolation and sparse-regime stability.} Removing the output gate causes the most severe drop among macroscopic variants on TS-E (-14.65\%) and leads to training collapse on \texttt{PDL}. Conversely, its impact remains moderate on classification and dense regression tasks (LTS-R: -4.19\%; LTS-C: -2.02\%). This aligns with the output gate's function as a nonlinear filter for retrieved memory. In sparse or extrapolative scenarios where retrieved states are unreliable, explicitly suppressing noisy channels is crucial; under dense sampling, the underlying liquid gating mechanism already provides sufficient implicit filtering.
\end{itemize}
	
\subsubsection{Liquid Gate Dynamics}
\begin{itemize}
	\item \textbf{Decoupled $\mu$ enables subspace diversity.} \emph{w/o Decoupled $\mu$} degrades performance consistently across all tasks and causes training collapse on PTS-R. This highlights the importance of head-specific interpolation for capturing multi-scale temporal dynamics.
	
	\item \textbf{Learnable $\mu$ balances adaptability and stability.} Fixing $\mu$ (\emph{w/o Learnable $\mu$}) degrades performance across most tasks, yet yields a 23.90\% MSE improvement on PTS-R. This suggests that under highly irregular, sparse observations, unconstrained learned interpolation can overfit to local noise, whereas the symmetric trapezoidal prior suppresses error accumulation at the cost of local flexibility.
	
	\item \textbf{Deterministic temporal decay fails across complex vector fields.} Replacing input-driven modulation with a static time-decay (\emph{w/o Input-aware $\mathbf{g}$}) causes severe degradation on regression tasks (LTS-R: -15.83\%; TS-I: -12.30\%; PTS-R: -39.02\%), whereas classification performance is largely unaffected (LTS-C: -2.00\%; PTS-C: -1.15\%). This indicates that input-aware gating is critical for distinguishing temporal structure from noise when modeling continuous-time dynamics.
\end{itemize}

\begin{figure*}[t]
	\centering
	\begin{subfigure}{0.48\linewidth}
		\centering
		\includegraphics[width=\linewidth]{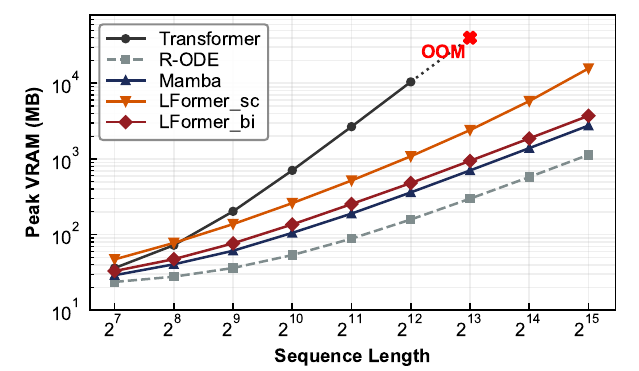}
		\caption{Peak GPU Memory Usage}
		\label{fig:complexity_vram}
	\end{subfigure}
	\hfill
	\begin{subfigure}{0.48\linewidth}
		\centering
		\includegraphics[width=\linewidth]{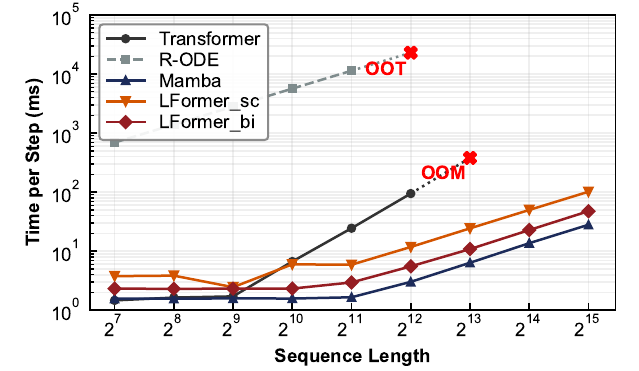}
		\caption{Per-Step Latency}
		\label{fig:complexity_time}
	\end{subfigure}
	\caption{Empirical complexity scaling. (a) Peak GPU memory usage and (b) training step latency as functions of sequence length. OOM (Out-Of-Memory) indicates that execution exceeded available GPU memory; OOT (Out-Of-Time) indicates that a single step exceeded the measurement time limit.}
	\label{fig:complexity_scaling}
\end{figure*}
	
\begin{figure}[t]
	\centering
	\includegraphics[width=\linewidth]{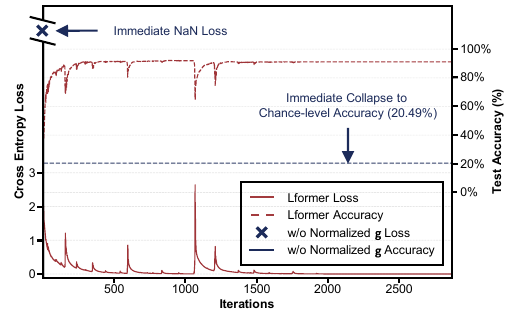}
	\caption{Optimization dynamics during the stabilization ablation on the PTS-C benchmark. The left and right vertical axes track training loss convergence (solid curves) and out-of-sample test accuracy (dashed curves), respectively.}
	\label{fig:ptsc_ablation_dynamics}
\end{figure}

\subsubsection{Numerical Stabilization}
Removing sequence-level normalization  degrades average performance across LTS-R (-8.22\%), LTS-C (-3.10\%), TS-I (-32.88\%), and TS-E (-6.44\%), with individual exceptions on \texttt{HB} and \texttt{PHY} (TS-E) where it proves unnecessary or mildly beneficial. This omission causes training collapse on PTS-C, PTS-R, and on \texttt{USH} in both TS-I and TS-E. Figure~\ref{fig:ptsc_ablation_dynamics} illustrates the training dynamics on PTS-C: \emph{w/o Normalized $\mathbf{g}$} suffers immediate gradient explosion and flatlines near the random-guessing level, while LFormer maintains stable optimization and smooth convergence.

The failure stems from the unconstrained cumulative sum in the liquid gate. Without normalization, the prefix sum over non-negative decay terms grows monotonically with sequence length, and the subsequent exponentiation amplifies this into numerical overflow or vanishing gradients. By bounding the cumulative gate, sequence-level normalization prevents these extremes. This mechanism is therefore essential for stable training on long or irregularly sampled sequences, while on shorter or regularly sampled tasks its role can be partially compensated by other components.

\subsection{Computational Efficiency (RQ3) \label{sec: efficiency}}
To address RQ3, we evaluate LFormer across three dimensions: (1) empirical complexity scaling with sequence length; (2) hardware utilization and inference throughput; and (3) the accuracy-efficiency Pareto trade-off.
	
\subsubsection{Empirical Complexity Scaling}
To quantify asymptotic complexity, we evaluate two variants of the parallel LGA operator: a causal variant LFormer$_\text{sc}$ that leverages a prefix scan (Eq.~\eqref{eq: lga_causal_linear_form}), and a non-causal bidirectional variant LFormer$_\text{bi}$ that uses full-sequence matrix multiplication (Eq.~\eqref{eq: lga_parallel_matrix_form}). We benchmark both variants against the standard Transformer~\cite{DBLP:conf/nips/VaswaniSPUJGKP17}, Mamba~\cite{DBLP:journals/corr/abs-2312-00752}, and R-ODE~\cite{DBLP:conf/nips/RubanovaCD19}. Synthetic trajectories are generated with a fixed batch size of $8$ and a feature dimension of $12$, while the sequence length scales exponentially from $2^6$ to $2^{15}$ steps. For fair comparison, we align hyperparameters across all models: hidden dimension $64$, depth $2$, and $4$ attention heads for attention-based variants.
	
Fig.~\ref{fig:complexity_vram} reports the peak VRAM consumption, defined as the maximum GPU memory allocated during a complete training iteration. The Transformer exhibits quadratic memory growth and exceeds GPU capacity at $2^{13}$ steps (OOM). Both LFormer$_\text{sc}$ and LFormer$_\text{bi}$ exhibit linear memory scaling across all tested lengths, where LFormer$_\text{sc}$ incurs higher memory overhead due to intermediate state maintenance during the prefix scan. Both LFormer variants show a linear spatial scaling profile, avoiding the quadratic bottleneck of global attention. Fig.~\ref{fig:complexity_time} reports the execution latency, measured as the average time over 10 consecutive training forward-backward steps. Due to sequential numerical integration, R-ODE incurs substantial overhead, exceeding $10^4$~ms at sequence length $2^{11}$. We mark this as out-of-time (OOT) and omit longer sequences. Both LFormer variants show a linear temporal scaling profile. Despite its native PyTorch implementation, LFormer closely matches Mamba which is accelerated by the hardware-optimized selective scan.

These empirical observations confirm that LFormer exhibits linear scaling complexity across spatial and temporal dimensions under both the non-causal and causal settings.

\begin{figure*}[t]
	\centering
	\begin{subfigure}{0.48\linewidth}
		\centering
		\includegraphics[width=\linewidth]{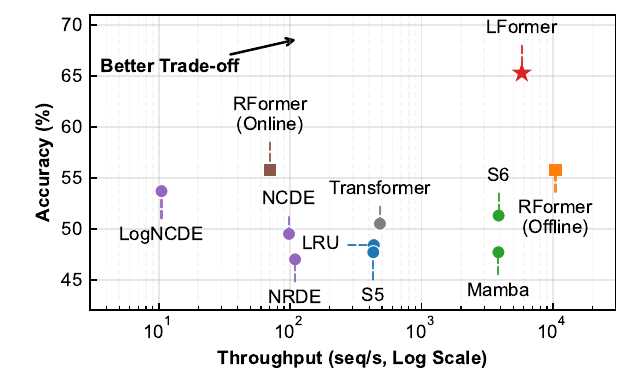}
		\caption{Pareto Frontier on the \texttt{MI} Dataset}
		\label{fig:pareto_mi}
	\end{subfigure}
	\hfill
	\begin{subfigure}{0.48\linewidth}
		\centering
		\includegraphics[width=\linewidth]{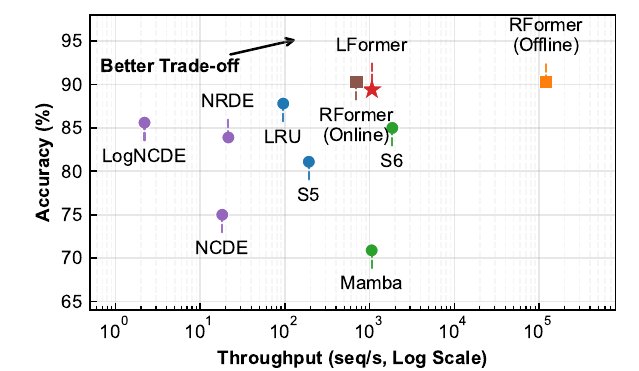}
		\caption{Pareto Frontier on the \texttt{EW} Dataset}
		\label{fig:pareto_ew}
	\end{subfigure}
	\caption{The efficiency-accuracy Pareto frontier on representative LTS-C datasets. The x-axis shows inference throughput (sequences per second) on a logarithmic scale; the y-axis shows classification accuracy.}
	\label{fig:pareto_frontier}
\end{figure*}

\begin{table}[t]
	\centering
	\caption{Parameter Count and Inference Throughput}
	\label{tab:throughput}
	\begin{minipage}{0.48\textwidth}
			\renewcommand{\arraystretch}{1.2} 
			\setlength{\tabcolsep}{6.0pt}      
			\footnotesize                     
			
			{\centering
					\begin{tabular}{l r r r}
							\toprule
							\textbf{Method} 
							& \textbf{$\mathbf{|\theta|}$  $\downarrow$} 
							& \textbf{$\boldsymbol{\mathcal{T}}$  $\uparrow$} 
							& \textbf{$\boldsymbol{\Delta}$  $\uparrow$} \\
							\midrule
							R-ODE~\cite{DBLP:conf/nips/RubanovaCD19}         
							& \textbf{34.56} 
							& 2.46 
							& $1.0\times$ \\
							
							Transformer~\cite{DBLP:conf/nips/VaswaniSPUJGKP17}   
							& 101.38 
							& 257.95 
							& $\sim$105$\times$ \\
							
							Mamba~\cite{DBLP:journals/corr/abs-2312-00752}           
							& \underline{66.69} 
							& \textbf{8,974.65} 
							& \textbf{$\sim$3,648$\times$} \\
							
							\textbf{LFormer (ours)}  
							& 103.86 
							& \underline{4,652.90} 
							& \underline{$\sim$1,891$\times$} \\
							\bottomrule
						\end{tabular}\par} 
			\vspace{6pt} 
			\par
			\footnotesize
			\footnotesize\noindent Note: The best results are highlighted in \textbf{bold}, and the second best are \underline{underlined}. The arrows $\uparrow$ and $\downarrow$ indicate whether a higher or lower value is better. $|\theta|$ denotes the number of trainable parameters (K), $\mathcal{T}$ represents the inference throughput (sequence per second), and $\Delta$ indicates the speedup relative to the R-ODE baseline.
		\end{minipage}
\end{table}

\subsubsection{Throughput and Hardware Efficiency}
We further evaluate the practical hardware efficiency of bidirectional LFormer by measuring parameter count and inference throughput under identical configurations. Synthetic sequences are generated with a batch size of $16$, a feature dimension of $12$, and a sequence length of $4,096$ steps.
	
Table~\ref{tab:throughput} summarizes the throughput and parameter alignment across the evaluated architectures. LFormer matches the parameter footprint of the standard Transformer while achieving substantially higher throughput, supporting the efficacy of its linear formulation. Although LFormer trails Mamba in absolute inference speed, this gap partly reflects Mamba's use of hardware-optimized parallel scan kernels; LFormer relies on standard PyTorch operators without custom CUDA kernels. Consequently, these statistics highlight the substantial potential of LFormer for future operator-level optimization.
		
\subsubsection{Accuracy vs. Efficiency Trade-off}
To quantify the accuracy-efficiency trade-off, we map empirical Pareto frontiers on the \texttt{MI} and \texttt{EW} datasets. All baselines are instantiated using optimal configurations from their respective literature. For RFormer~\cite{10.5555/3737916.3741287}, we evaluate two configurations: an online variant that computes signatures during inference, and an offline variant that uses pre-computed cached signatures.
		
Fig.~\ref{fig:pareto_frontier} plots the classification accuracy against logarithmic inference throughput. An ideal architecture occupies the top-right quadrant, maximizing both accuracy and throughput. The visualization reveals several insights:
\begin{itemize}
	\item \textbf{Solver-dependent Models' Bottlenecks:} NCDE, NRDE, and LogNCDE occupy the lowest efficiency tier. Their reliance on numerical solvers necessitates iterative vector-field evaluations per step, imposing a prohibitive temporal bottleneck that limits scalability.
	
	\item \textbf{Transformer's Scalability Limits:} The standard Transformer achieves moderate throughput via parallelized attention. However, its quadratic complexity limits scalability over extended sequences, resulting in OOM exceptions on the \texttt{EW} dataset.
	
	\item \textbf{Linear-Time RNNs' Suboptimality:} While LRU and S5 enhance efficiency via diagonalized recurrent matrices, they deliver lower performance, limiting their position on the accuracy-efficiency frontier.
	
	\item \textbf{RFormer's Real-time Constraint:} The peak throughput of the offline RFormer variant is enhanced by pre-computed signature transformations, which reduce the effective sequence lengths to 200 (\texttt{MI}) and 10 (\texttt{EW}). This preprocessing step assumes offline access to the full sequence, and the online variant shows lower throughput.
	
	\item \textbf{LFormer's Pareto Optimality:} LFormer (bidirectional) traces the leading Pareto frontier on both datasets. It achieves the highest accuracy on \texttt{MI} and ranks second on \texttt{EW} dataset, with negligible throughput degradation compared to the fastest models. 
\end{itemize}
		
These observations position LFormer as an efficient continuous-time framework that combines competitive temporal modeling capacities with high-throughput execution in non-causal full-sequence settings.

\section{Limitations and Future Work \label{sec: limitations}}
The current LGA formulation has several limitations. First, the sequence-level normalization bounds the total decay budget for stable training, but it assumes offline full-sequence access; a prefix-normalized variant would be required for online or causal inference. Second, the learnable endpoint interpolation improves flexibility on most benchmarks, but can overfit under highly sparse and noisy conditions, as evidenced by the PTS-R task where a fixed $\mu=0.5$ outperforms the learned variant. Third, the constrained extrapolation performance on non-stationary signals reveals the architectural boundary of LGA. These limitations suggest several directions for future work including streaming-compatible normalization, task-adaptive endpoint interpolation and integrating LGA with time-aware query networks.
		
\section{Conclusion \label{sec: conclusion}}
This paper presented Liquid Gated Attention (LGA), a solver-free operator that embeds observed temporal intervals into gated linear attention. By unifying a continuous-time gating mechanism derived from liquid time-constant dynamics with fast-weight associative memory and a sequence-level normalization, LGA maintains linear spatio-temporal complexity and empirically stable training over extended horizons. Built upon LGA, LFormer provides a modular backbone for continuous-time representation learning. Across six tasks and sixteen benchmarks, LFormer delivers strong results across three core temporal modeling capabilities: long-range dependency modeling, fine-grained state tracking, and trajectory reconstruction from sparse and noisy observations. 

Ultimately, LGA demonstrates how continuous-time dynamical principles can be embedded into parallel sequence architectures, positioning LFormer as an efficient backbone for time series representation learning.


\appendices
\section{Derivation of the Closed-form Integral Solution for 1D LTC}
We solve the linear non-autonomous ODE that governs a decoupled scalar LTC state $s(t) \in \mathbb{R}$ under a scalar exogenous input $x(t) \in \mathbb{R}$:
\begin{equation}
	\frac{\mathrm{d}s(t)}{\mathrm{d}t} = -\bigl[w_\tau + f\bigl(x(t)\bigr)\bigr]s(t) + b_\tau f\bigl(x(t)\bigr),
	\label{app:eq:raw_1d_ltc}
\end{equation}
where $w_\tau, b_\tau \in \mathbb{R}$ represent learnable scalar parameters. 

Following the structural symmetry assumption established in~\cite{DBLP:journals/natmi/HasaniLALRTTR22}, we introduce a balancing inverse time-constant term $w_\tau$ into the bias compartment to symmetrize the algebraic structure:
\begin{equation}
	\frac{\mathrm{d}s(t)}{\mathrm{d}t} = -\bigl[w_\tau+f(x(t))\bigr]s(t) + \bigl[w_\tau+f(x(t))\bigr]b_\tau.
	\label{app:eq:symmetric_ode}
\end{equation}
The effect of this structural balancing term on the solution has been analyzed in~\cite{DBLP:journals/natmi/HasaniLALRTTR22}, which demonstrated its negligible impact on the approximation fidelity.

By defining the time-variant system rate coefficient as $p(t) \triangleq w_\tau+f\bigl(x(t)\bigr)$, we map Eq.~\eqref{app:eq:symmetric_ode} onto the canonical form of a first-order linear non-homogeneous ODE:
\begin{equation}
	\frac{\mathrm{d}s(t)}{\mathrm{d}t} + p(t)s(t) = p(t)b_\tau.
	\label{app:eq:canonical_ode}
\end{equation}

To derive the analytical trajectory of Eq.~\eqref{app:eq:canonical_ode}, we define the standard non-autonomous integrating factor $\mu(t)$ as:
\begin{equation}
	\mu(t) \triangleq \exp\left[\int_0^t p(u)\,\mathrm{d}u\right].
	\label{app:eq:integrating_factor}
\end{equation}

From the fundamental theorem of calculus, the derivative of the integrating factor is:
\begin{equation}
	\frac{\mathrm{d}\mu(t)}{\mathrm{d}t} = \mu(t)\frac{\mathrm{d}}{\mathrm{d}t}\left[\int_0^t p(u)\,\mathrm{d}u\right] = \mu(t)p(t).
\end{equation}

Multiplying both sides of the canonical ODE in Eq.~\eqref{app:eq:canonical_ode} by $\mu(t)$ yields:
\begin{equation}
	\mu(t)\frac{\mathrm{d}s(t)}{\mathrm{d}t} + \underbrace{\mu(t)p(t)s(t)}_{s(t)\frac{\mathrm{d}\mu(t)}{\mathrm{d}t}} = \mu(t)p(t)b_\tau.
\end{equation}

Recognizing the left-hand side as the exact derivative of the product $\mu(t)s(t)$, we condense the equation to:
\begin{equation}
	\frac{\mathrm{d}}{\mathrm{d}t}\bigl[\mu(t)s(t)\bigr] = \mu(t)p(t)b_\tau.
	\label{app:eq:condensed_ode}
\end{equation}

Performing definite integration on Eq.~\eqref{app:eq:condensed_ode} from the initial boundary $0$ to the target timestamp $t$, and noting that $\mu(0) = \exp(0) = 1$, we obtain:
\begin{equation}
	\mu(t)s(t) - s(0) = b_\tau\int_{0}^{t}\mu(v)p(v)\,\mathrm{d}v,
	\label{app:eq:integrated_state}
\end{equation}
where $s(0) \in \mathbb{R}$ denotes the initial state condition. 

Given that the exponential integrating factor $\mu(t)$ is strictly positive, multiplying both sides by $\mu^{-1}(t)$ isolates the continuous state trajectory, revealing a nested integral expression:
\begin{equation}
	\begin{aligned}
		s(t) &= s(0)\mu^{-1}(t) \\
		&\quad + b_\tau\int_{0}^{t}\mu^{-1}(t)\mu(v)p(v)\,\mathrm{d}v \\[3pt]
		&= s(0)\exp\left[-\int_0^t p(u)\,\mathrm{d}u\right] \\
		&\quad + b_\tau \underbrace{\int_{0}^{t}
		\exp\left[-\int_v^t p(u)\,\mathrm{d}u\right]
		p(v)\,\mathrm{d}v}_{\text{Nested Integral Term } n(t)}.
		\label{app:eq:nested_solution}
	\end{aligned}
\end{equation}

To simplify the nested integral $n(t)$, we apply integration by substitution. Define the auxiliary function $a(v)$ as:
\begin{equation}
	a(v) \triangleq \int_v^t p(u)\,\mathrm{d}u.
	\label{app:eq:auxiliary_function}
\end{equation}
Differentiating with respect to $v$ gives $\mathrm{d}a(v) = -p(v)\,\mathrm{d}v$. The integration limits transform as:
\begin{equation}
	a(v) = 
	\begin{cases}
		0, & \text{when } v = t, \\
		\int_0^t p(u)\,\mathrm{d}u, & \text{when } v = 0.
	\end{cases}
\end{equation}
Substituting these relations into $n(t)$ yields:
\begin{equation}
	\begin{aligned}
		n(t)
		&= \int_{a(0)}^{0} \exp[-a] \bigl(-\mathrm{d}a\bigr) \\
		&= \int_{0}^{a(0)} \exp[-a] \,\mathrm{d}a \\
		&= \Bigl[-\exp[-a]\Bigr]_{0}^{a(0)} \\
		&= 1 - \exp\bigl[-a(0)\bigr].
	\end{aligned}
		\label{app:eq:nested_evaluation}
\end{equation}
Substituting the result of Eq.~\eqref{app:eq:nested_evaluation} into  Eq.~\eqref{app:eq:nested_solution} reduces the state trajectory to:
\begin{equation}
	s(t) = \bigl[s(0)-b_\tau\bigr]\exp\bigl[-a(0)\bigr] + b_\tau.
	\label{app:eq:simplified_trajectory}
\end{equation}

Recalling that $a(0) = \int_0^t p(u)\,\mathrm{d}u$ and $p(t) = w_\tau + f\bigl(x(t)\bigr)$, we obtain the closed-form integral solution:
\begin{equation}
	s(t) = \bigl(s(0)-b_\tau\bigr) \exp\left[-w_\tau t - \int_{0}^{t}f\bigl(x(u)\bigr)\,\mathrm{d}u\right] + b_\tau.
	\label{app:eq:final_closed_form}
\end{equation}

\section{Error Analysis of the Learnable Endpoint Interpolation}
To evaluate the local truncation error of the proposed interpolation, we first establish the requisite mathematical smoothness. Assuming the underlying trajectory $x(u)$ is smooth, the integrand $\tilde{x}(u) = \sigma(\mathbf{x}(u)\mathbf{w}_f + b_f) \in (0,1)$ is twice continuously differentiable, $\tilde{x} \in C^2[t_{n-1}, t_n]$, over the localized time segment with step size $\delta_n = t_n - t_{n-1}$.

A Taylor expansion of the endpoint value $\tilde{x}(t_n)$ centered at $t_{n-1}$ yields:
\begin{equation}
	\tilde{x}(t_n) = \tilde{x}(t_{n-1}) + \delta_n \tilde{x}'\left(t_{n-1}\right) + \frac{\delta_n^2}{2} \tilde{x}''\left(t_{n-1}\right) + \mathcal{O}\left(\delta_n^3\right).
	\label{app:eq:supp_taylor_x}
\end{equation}

The exact definite integral can be expressed via Taylor expansion as:
\begin{equation}
	\begin{aligned}
	\int_{t_{n-1}}^{t_n} \tilde{x}(u)\,\mathrm{d}u
	&= \delta_n \tilde{x}\left(t_{n-1}\right) \\
	&\quad + \frac{\delta_n^2}{2}\tilde{x}'\left(t_{n-1}\right) \\
	&\quad + \frac{\delta_n^3}{6}\tilde{x}''\left(t_{n-1}\right)
	+ \mathcal{O}\left(\delta_n^4\right).
	\end{aligned}
	\label{app:eq:supp_exact_integral}
\end{equation}

To bypass sequential numerical integration, LGA approximates this integral via a learnable endpoint interpolation, parameterized by a bounded coefficient $\mu = \sigma(\theta) \in (0,1)$:
\begin{equation}
	I_{\text{approx}} \triangleq \delta_n \left[\mu \tilde{x}\left(t_{n-1}\right) + \left(1-\mu\right) \tilde{x}\left(t_n\right)\right].
	\label{app:eq:supp_operator_A}
\end{equation}

Substituting the expansion of $\tilde{x}(t_n)$ from Eq.~\eqref{app:eq:supp_taylor_x} into the numerical surrogate in Eq.~\eqref{app:eq:supp_operator_A} yields:
\begin{equation}
	\begin{aligned}
		I_{\text{approx}}
		&= \delta_n\Bigl\{\mu\tilde{x}\left(t_{n-1}\right) \\
		&\quad +(1-\mu)\Bigl[\tilde{x}\left(t_{n-1}\right)
		+\delta_n\tilde{x}'\left(t_{n-1}\right) \\
		&\qquad +\frac{\delta_n^2}{2}\tilde{x}''\left(t_{n-1}\right)
		+\mathcal{O}\left(\delta_n^3\right)\Bigr]\Bigr\} \\[3pt]
		&= \delta_n\tilde{x}\left(t_{n-1}\right)
		+\delta_n^2(1-\mu)\tilde{x}'\left(t_{n-1}\right) \\
		&\quad +\frac{\delta_n^3}{2}(1-\mu)
		\tilde{x}''\left(t_{n-1}\right)
		+\mathcal{O}\left(\delta_n^4\right).
	\end{aligned}
	\label{app:eq:supp_expanded_A}
\end{equation}

The local truncation error $E$ is defined as the algebraic difference between the exact integral (Eq.~\eqref{app:eq:supp_exact_integral}) and the numerical surrogate (Eq.~\eqref{app:eq:supp_expanded_A}):
\begin{equation}
	\begin{aligned}
		E &= \int_{t_{n-1}}^{t_n} \tilde{x}(u)\,\mathrm{d}u - I_{\text{approx}} \\[4pt]
		&= \left[\frac{1}{2}-(1-\mu)\right]
		\delta_n^2\tilde{x}'\left(t_{n-1}\right) \\
		&\quad + \left[\frac{1}{6}-\frac{1-\mu}{2}\right]
		\delta_n^3\tilde{x}''\left(t_{n-1}\right)
		+\mathcal{O}\left(\delta_n^4\right) \\[4pt]
		&= \left(\mu-\frac{1}{2}\right)
		\delta_n^2\tilde{x}'\left(t_{n-1}\right) \\
		&\quad + \left(\frac{\mu}{2}-\frac{1}{3}\right)
		\delta_n^3\tilde{x}''\left(t_{n-1}\right)
		+\mathcal{O}\left(\delta_n^4\right).
	\end{aligned}
	\label{app:eq:supp_error_algebra}
\end{equation}

Equation~\eqref{app:eq:supp_error_algebra} characterizes the error under two distinct parameterization regimes:

\begin{itemize}
	\item \textbf{General case ($\mu \neq 0.5$):} The error is dominated by the $\delta_n^2$ term, giving a local truncation error of $\mathcal{O}(\delta_n^2)$. The magnitude of the leading error is proportional to $|\mu - 0.5|$, so that moderate deviations from the midpoint do not induce unbounded error amplification. This property allows the model to adapt the interpolation to data-dependent temporal dynamics while preserving numerical stability.
	
	\item \textbf{Symmetric case ($\mu = 0.5$):} The coefficient of the leading $\delta_n^2$ term vanishes, and the error algebraically reduces to:
	\begin{equation}
		E = -\frac{1}{12} \delta_n^3 \tilde{x}''\left(t_{n-1}\right) + \mathcal{O}\left(\delta_n^4\right) = \mathcal{O}\left(\delta_n^3\right).
		\label{app:eq:supp_trapezoidal_limit}
	\end{equation}
	Under this condition, the interpolation recovers the classical trapezoidal rule with second-order local accuracy. For well-behaved integrands, accumulating this local error over a finite interval yields a global error of $\mathcal{O}(\delta^2)$.
\end{itemize}

\section{Numerical Bounds and Gradient Stability of Sequence-level Normalization}
Let $\mathbf{u} = \mathbf{\Delta} \odot \overline{\mathbf{x}}$ denote the unnormalized decay vector, where $u_i = \delta_i \overline{x}_i$ for $i \in \{1, \dots, n\}$. In practice, the temporal intervals are non-negative and the interpolated trajectory satisfies $\overline{x}_i \in (0,1)$. Consequently, the elements are non-negative: $u_i \ge 0$ for all $i$. To prevent unbounded accumulation in the cumulative gate, we apply the sequence-level normalization:
\begin{equation}
	\hat{u}_i = \frac{u_i}{\sum_{k=1}^{n} u_k + \epsilon},
	\label{app:eq:supp_norm_element}
\end{equation}
where $\epsilon > 0$ is a small stability constant that prevents division by zero. Since $u_i \ge 0$ and $\epsilon > 0$, we have $\hat{u}_i \ge 0$ for all $i$.

Let $s_j \triangleq \sum_{i=1}^{j} \hat{u}_i$ denote the $j$-th element of the cumulative sum vector $\mathbf{s} = \mathrm{CumSum}(\hat{\mathbf{u}})$. Since $\hat{u}_i \ge 0$, the sequence $\{s_j\}_{j=1}^n$ is monotonically non-decreasing. The terminal value $s_n$ is:
\begin{equation}
	s_n = \sum_{i=1}^{n} \hat{u}_i = \frac{\sum_{i=1}^{n} u_i}{\sum_{k=1}^{n} u_k + \epsilon}.
	\label{app:eq:supp_terminal_sum}
\end{equation}
where the denominator strictly exceeds the numerator, yielding $s_n < 1$. Combined with the lower bound $s_j \ge 0$, we obtain $s_j \in [0, 1)$ for all $j\in\left\{1, \dots, n\right\}$. Thus, the stabilized cumulative gate $\dot{g}_{\text{stable}, j} = \exp(-s_j) \in (e^{-1}, 1]$ for all $j\in\left\{1, \dots, n\right\}$.

\vspace{1em}
\noindent\textbf{Remark on temporal semantics.} The normalized gate is not algebraically equivalent to the LTC continuous-time transition. Instead, it functions as a regularized surrogate that bounds the total decay budget while preserving the relative temporal allocation across the sequence. This design is suitable for offline full-sequence representation learning; for online streaming or causal inference, a prefix-normalized variant is required to avoid dependence on future observations.

\vspace{1em}
\noindent\textbf{Remark on gradient stability.} Beyond bounding the forward-pass values, the sequence-level normalization alters the asymptotic behavior of the gradients during back propagation through time (BPTT). Consider the gradient of the cumulative gate $\dot{g}_{\text{stable}, k} = \exp(-s_k)$ with respect to the unnormalized decay term $u_i$ for $i \le k$. 
By the chain rule:
\begin{equation}
	\frac{\partial \dot{g}_{\text{stable}, k}}{\partial u_i} 
	= -\dot{g}_{\text{stable}, k} \cdot \frac{\partial s_k}{\partial u_i}
	= -\dot{g}_{\text{stable}, k} \cdot \sum_{j=1}^{k} \frac{\partial \hat{u}_j}{\partial u_i}.
\end{equation}
The partial derivative of the normalized element $\hat{u}_j$ with respect to $u_i$ is evaluated as:
\begin{equation}
	\frac{\partial \hat{u}_j}{\partial u_i} = 
	\begin{cases}
		\displaystyle \frac{\sum_{\ell \neq i} u_\ell + \epsilon}{(\sum_{\ell=1}^{n} u_\ell + \epsilon)^2}, & j = i, \\[14pt]
		\displaystyle -\frac{u_j}{(\sum_{\ell=1}^{n} u_\ell + \epsilon)^2}, & j \neq i.
	\end{cases}
\end{equation}
Summing these derivatives over $j = 1, \dots, k$ simplifies to:
\begin{equation}
	\sum_{j=1}^{k} \frac{\partial \hat{u}_j}{\partial u_i}
	= \frac{\sum_{\ell=1}^{n} u_\ell + \epsilon - \sum_{j=1}^{k} u_j}{(\sum_{\ell=1}^{n} u_\ell + \epsilon)^2}
	\ge \frac{\epsilon}{(\sum_{\ell=1}^{n} u_\ell + \epsilon)^2} > 0,
\end{equation}
where the inequality holds because $\sum_{j=1}^{k} u_j \le \sum_{\ell=1}^{n} u_\ell$. 
Substituting this bound back into the chain rule, we obtain the absolute gradient magnitude:
\begin{equation}
	\left| \frac{\partial \dot{g}_{\text{stable}, k}}{\partial u_i} \right|
	= \dot{g}_{\text{stable}, k} \cdot \sum_{j=1}^{k} \frac{\partial \hat{u}_j}{\partial u_i}
	\ge e^{-1} \cdot \frac{\epsilon}{(\sum_{\ell=1}^{n} u_\ell + \epsilon)^2}.
\end{equation}

This quantitative result reveals a crucial property. Let $S \triangleq \sum_{\ell=1}^{n} u_\ell$ denote the total sum of unnormalized decay terms along the sequence. While the gradient signal does inversely scale with the total sum of the sequence, the attenuation is strictly \textbf{polynomial} $\mathcal{O}(S^{-2})$. Such polynomial scaling is manageable and easily mitigated by modern adaptive optimizers. In contrast, for the unnormalized variant, the gradient evaluates to $\frac{\partial \dot{g}_k}{\partial u_i} = -\dot{g}_k$, which decays \textbf{exponentially} $\mathcal{O}(e^{-S})$ as the cumulative sum grows. Thus, the normalization mathematically mitigates the exponential vanishing gradient problem, enabling robust optimization across extremely long temporal horizons.

\section{Dataset Details and Preprocessing Protocols}
\subsection{UEA Multivariate Time Series Archive}
We select six benchmarks from the UEA multivariate time series archive~\cite{DBLP:journals/datamine/BagnallLBLK17}: EigenWorms (\texttt{EW}), EthanolConcentration (\texttt{EC}), Heartbeat (\texttt{HB}), MotorImagery (\texttt{MI}), SelfRegulationSCP1 (\texttt{SCP1}), 
and SelfRegulationSCP2 (\texttt{SCP2}). Following the selection criteria in~\cite{DBLP:conf/icml/0001MQCLL24}, we adopt benchmarks with more than 200 instances to ensure reliable evaluation. The selected benchmarks span a broad range of sequence lengths (405 to 17,984 steps), enabling assessment of long-range modeling capacity. Duplicate trajectories are removed following the preprocessing protocols in~\cite{DBLP:conf/icml/0001MQCLL24, DBLP:conf/icml/MorrillSKF21}. The remaining sequences are randomly partitioned into training ($70\%$), validation ($15\%$), and testing ($15\%$). Observation timestamps are normalized to $[0, 1]$.

\subsection{BIDMC Physiological Database}
We use three subsets from the BIDMC database~\cite{DBLP:journals/tbe/PimentelJCBWTC17}: heart rate (\texttt{HR}), respiratory rate (\texttt{RR}), and blood oxygen saturation (\texttt{SpO}$_2$). The task predicts a single continuous scalar (the vital sign value) from the full trajectory. The input consists of dual-channel physiological waveforms sampled at 125~Hz. A 32-second sliding window produces dense sequences spanning 4,000 time steps. Each sequence is labeled with the corresponding heart rate, respiratory rate, or blood oxygen saturation level. The remaining sequences are randomly partitioned into training ($70\%$), validation ($15\%$), and testing ($15\%$). Observation timestamps are normalized to $[0, 1]$.

\subsection{Human Activity Benchmark}
The Human Activity (\texttt{HA}) benchmark~\cite{DBLP:conf/nips/RubanovaCD19} comprises 12-dimensional time-series recordings acquired from four wearable sensors across five subjects. The objective centers on dense, point-wise classification, mapping each temporal observation to one of seven distinct physical activity categories. Adhering to the protocol in~\cite{DBLP:conf/nips/RubanovaCD19}, the raw trajectories are processed into 6,554 distinct sequence instances. While each canonical sequence contains 211 steps, a subset of 50 temporal observations is randomly sampled per instance to create irregularly sampled sequences. Timestamps are normalized by a scaling factor of $1/211$, and the dataset is partitioned into 4,194 training, 1,049 validation, and 1,311 testing sequences.

\subsection{Pendulum Tracking Benchmark}
The Pendulum (\texttt{PDL}) dataset~\cite{DBLP:conf/icml/BeckerPGZTN19} consists of sequences containing $24 \times 24$ pixel images, where the dense target at each temporal index corresponds to the angular coordinate $\left(\sin \theta, \cos \theta\right)$. Consistent with~\cite{DBLP:conf/icml/SchirmerELR22}, 4,000 sequences are synthesized. Each baseline trajectory spans 100 uniformly spaced steps, from which 50 frames are randomly sub-sampled to construct an irregular temporal grid. Spatially and temporally correlated noise is introduced to the pixel grids following~\cite{DBLP:conf/icml/BeckerPGZTN19}. Observation timestamps are scaled by a factor of $0.1$ following~\cite{park2025amortized}. The data configuration comprises 2,000 training, 1,000 validation, and 1,000 testing instances.

\subsection{Synthetic 2D Spiral Trajectories}
Synthetic 2D spiral trajectories ($\texttt{2DS}_\alpha$)~\cite{DBLP:conf/nips/ChenRWFSL23} are generated to evaluate trajectory interpolation (TS-I) and extrapolation (TS-E). Each individual trajectory is stochastically assigned a clockwise or counter-clockwise orientation, governed by Archimedean parametric equations. The first geometric variation is formulated as:
\begin{equation}
	\begin{aligned}
		x(t) &= \left(a + bt\right) \cos(t), \\
		y(t) &= \left(a + bt\right) \sin(t),
	\end{aligned}
\end{equation}
and the alternative formulation is defined via:
\begin{equation}
	\begin{aligned}
		x(t) &= \left(a + \frac{50b}{e - t}\right) \cos(e - t), \\
		y(t) &= \left(a + \frac{50b}{e - t}\right) \sin(e - t),
	\end{aligned}
\end{equation}
where $a$ and $b$ represent latent structural parameters drawn from normal distributions: $a \sim \mathcal{N}\left(0, \alpha\right)$ and $b \sim \mathcal{N}\left(0.3, \alpha\right)$. Three distinct benchmarks are established by varying the variance scale parameter $\alpha \in \left\{2, 0.2, 0.02\right\}$. Additive Gaussian noise $\mathcal{N}\left(0, 0.1\right)$ is added to the training trajectories. Each complete trajectory contains 150 uniformly spaced time steps. To simulate partial observations with irregular sampling, the model receives 30 points randomly drawn from the first half of the trajectory ($t \in [1, 75]$). The task requires interpolating the unobserved positions within this interval and extrapolating the full trajectory over the entirely unobserved second half ($t \in [76, 150]$). For each value of $\alpha$, we generate 300 trajectories, partitioned into 200 training and 100 testing instances.

\subsection{USHCN Benchmark}
The United States Historical Climatology Network (\texttt{USH}) dataset~\cite{osti_1394920} comprises daily measurements from 1,218 weather stations across the United States, covering five variables: precipitation, snowfall, snow depth, and daily minimum and maximum temperature. Following the preprocessing pipeline of~\cite{park2025amortized}, we retain a subset of stations with continuous records spanning a four-year period starting from 1990. Stations whose observations end before 1994 or begin after 1990 are excluded, along with measurements flagged with unreliable quality markers. Each variable is standardized independently to zero mean and unit variance, and extreme values exceeding four standard deviations from the mean are treated as missing. To introduce irregular sampling, we randomly subsample 50\% of the time points and independently drop 20\% of the remaining observations, following the protocol of~\cite{park2025amortized}. Only stations retaining at least 730 observations after subsampling are kept. The processed dataset contains 1,192 stations, which are randomly partitioned into training (60\%), validation (20\%), and testing (20\%) sets at the station level. Observation timestamps are encoded as integer days elapsed since January 1, 1950, and normalized to $[0, 1]$.

\subsection{PhysioNet Benchmark}
The PhysioNet Challenge 2012 dataset (\texttt{PHY})~\cite{Silva2012PredictingIM} contains 8,000 multivariate clinical time series collected from intensive care unit (ICU) patients during the first 48 hours after admission. Each record comprises 41 measurements, of which we discard 4 static features (age, gender, height, and ICU type) and retain the remaining 37 time-varying clinical variables, following the preprocessing protocol of~\cite{DBLP:conf/nips/RubanovaCD19}. Each variable is independently normalized to $[0, 1]$ via min-max scaling, with statistics computed separately for each data split. The processed records are randomly partitioned into training (60\%), validation (20\%), and testing (20\%) sets at the patient level.

\section{Detailed Baseline Specifications}
This section provides the detailed description for all baseline architectures used in the experimental evaluation.

\subsection{Discrete-Time Baselines}
\begin{itemize}
	\item GRU~\cite{DBLP:conf/emnlp/ChoMGBBSB14} models sequential dependencies through discrete gated recurrent units over uniform step grids. It compresses past information via interleaved reset and update gates, serving as a baseline for discrete recurrence. GRU-$\Delta_t$ is a variant of GRU which takes additional time intervals as input.
	
	\item Transformer~\cite{DBLP:conf/nips/VaswaniSPUJGKP17} employs global dot-product self-attention mechanisms over static discrete tokens. It eliminates sequential dependencies via fully parallel matrix operations, evaluating standard discrete relational modeling without explicit temporal distance metrics.
	
	\item LRU~\cite{10.5555/3618408.3619518} linearizes and diagonalizes recurrent layers into decoupled complex-valued hidden states to mitigate optimization bottlenecks over long sequences.
	
	\item Mamba (S6)~\cite{DBLP:journals/corr/abs-2312-00752} introduces input-dependent, selective linear state-space layers parallelized via a hardware-aware associative scan algorithm. In this setup, Mamba operates over uniform discrete steps, with selection step sizes driven by local token features rather than physical temporal distances.
	
	\item GRU-D~\cite{2018Recurrent} incorporates the missing mask and time intervals into the GRU architecture, enabling it to capture long-term temporal dependencies while leveraging informative missingness for improved prediction.
	
	\item RKN-$\Delta_t$~\cite{DBLP:conf/icml/BeckerPGZTN19} factorizes the latent state representation to scalar Kalman updates. It employs local linear dynamics for state propagation and provides explicit uncertainty estimates alongside predictions.
\end{itemize}

\subsection{Solver-Dependent Continuous-Time Baselines}
\begin{itemize}
	\item NODE~\cite{DBLP:conf/nips/ChenRBD18} parameterizes the latent state vector-field derivative via a neural network, mapping state trajectories through continuous integration solved by adaptive-step numerical ODE solvers. It utilizes the adjoint sensitivity method to achieve constant-memory backpropagation.
	
	\item The Latent ODE framework (comprising R-ODE~\cite{DBLP:conf/nips/RubanovaCD19} and O-ODE~\cite{DBLP:conf/nips/RubanovaCD19}) proposes a continuous-time generative framework for irregularly sampled time series. It utilizes neural ODE variants driven by either a recurrent recognition network (R-ODE) or an observation-aware continuous encoder (O-ODE) to infer latent state trajectories.
	
	\item LSDE~\cite{DBLP:conf/nips/ZengGK23} formulates variational Bayesian inference over continuous latent trajectories governed by stochastic differential equations. It performs path tracking over homogeneous vector fields through a simplified Kullback-Leibler divergence and parameterized continuous diffusion gradients.
	
	\item NCDE~\cite{DBLP:conf/nips/KidgerMFL20} extends the neural differential paradigm to controlled differential equations, addressing limitations related to initial conditions. It constructs continuous-time paths from irregularly sampled inputs via cubic spline interpolation to drive the hidden state evolution.
	
	\item NRDE~\cite{DBLP:conf/icml/MorrillSKF21} combines neural controlled dynamics with rough path theory to process high-frequency, irregularly sampled signals. It maps localized signal intervals into bounded algebraic log-signatures, compressing high-frequency variations into summary statistics to accelerate continuous-time scaling.
	
	\item LogNCDE~\cite{DBLP:conf/icml/0001MQCLL24} integrates the Log-ODE numerical approximation scheme into the neural rough path paradigm for stiff controlled differential equations, targeting long-range dependencies over extended integration horizons.
	
	\item ContiFormer~\cite{DBLP:conf/nips/ChenRWFSL23} combines the continuous-time representation of neural ODEs with self-attention. It extends the attention formulation into the continuous-time domain to parameterize input-dependent vector fields.
	
	\item CRU~\cite{DBLP:conf/icml/SchirmerELR22} evolves hidden state according to a linear stochastic differential equation, yielding temporal continuity between hidden states and a gating mechanism that integrates noisy irregular observations.
	
	\item GRU-ODE-B~\cite{DBLP:conf/nips/BrouwerSAM19} combines NODE with a Bayesian update network to handle observations that are irregular in both time and feature dimensions, encoding a continuity prior for the latent process.
\end{itemize}

\subsection{Solver-Free Continuous-Time Baselines}
\begin{itemize}
	\item CfC~\cite{DBLP:journals/natmi/HasaniLALRTTR22} derives a bounded, analytic closed-form integral solution to approximate the dynamics of liquid time-constant networks. It bypasses the numerical integration overhead, enabling efficient continuous-time modeling.
	
	\item RFormer~\cite{10.5555/3737916.3741287} enhances self-attention with multi-view algebraic path signatures. It constructs non-sequential continuous representations designed to bypass sequential integration overhead across long temporal trajectories.
	
	\item mTAND~\cite{shukla2021multitime} leverages a continuous multi-time attention embedding kernel to map irregular, variable-length observation coordinates into fixed-dimensional continuous-time representations, preserving non-uniform interval metrics without tracking hidden derivatives.
	
	\item S5~\cite{smith2023simplified} formulates structured linear state-space layers over non-uniform grids, driving state evolution analytically via bilinear (trapezoidal) transformations. It leverages a time-varying parallel associative scan to directly process physical time intervals under full parallelization.
	
	\item ACSSM~\cite{park2025amortized} leverages multi-marginal Doob's $h$-transform mechanics and stochastic optimal control to construct a simulation-free latent dynamics framework. It integrates parallelized variational inference into a linear architecture, enabling parallel optimization over continuous trajectories.
\end{itemize}

\section{Detailed Hyperparameter and Implementation Configurations}
\label{app:sec:appx_implementation}

This section provides the complete architectural specifications and hyperparameter settings for LFormer.

Before detailing the network architectures, we adopt the following notational conventions. \texttt{Linear($d_{\text{in}}$, $d_{\text{out}}$)} denotes a fully-connected layer with $d_{\text{in}}$ input and $d_{\text{out}}$ output dimensions. \texttt{Conv($c_{\text{in}}$, $c_{\text{out}}$, $k$, $s$, $p$)} denotes a 2D convolutional layer with $c_{\text{in}}$ input channels, $c_{\text{out}}$ output channels, kernel size $k$, stride $s$, and padding $p$. \texttt{MaxPool($k$, $s$)} denotes a 2D max-pooling operation with kernel size $k$ and stride $s$.

Across all benchmarks, each Liquid Mixer layer consists of a multi-head liquid gated attention (MHLGA) module followed by a SwiGLU channel mixer, organized in a pre-normalization residual structure. LFormer stacks $L$ such layers, with hidden dimension $d$, $H$ attention heads, and SwiGLU expansion dimension $d_{ff}$, denoted as \texttt{$L$ $\times$ [MHLGA($d$, $H$) $\rightarrow$ SwiGLU($d_{ff}$)]}.

All tasks are optimized using Adam with $\beta_1 = 0.9$, $\beta_2 = 0.999$, no weight decay, and a batch size chosen to maximize GPU memory utilization. Early stopping is applied when the validation metric does not improve for 20 consecutive epochs. The learning rate for each benchmark is listed alongside its architecture below.
\begin{itemize}
	\item \textbf{LTS-R Benchmark:}
	\begin{itemize}
		\item Embedder: \texttt{Linear(2, 64) $\rightarrow$ ReLU() $\rightarrow$ LayerNorm()}
		\item Liquid Mixer: \texttt{2 $\times$ [MHLGA(64, 4) $\rightarrow$ SwiGLU(176)]}
		\item Predictor: \texttt{Linear(64, 1)}
		\item Learning Rate: \texttt{0.001}
	\end{itemize}
	
	\item \textbf{LTS-C Benchmark:} For \texttt{EC}, \texttt{EW}, \texttt{HB}, and \texttt{MI}, we adopt the standard configuration ($d=64$, $H=4$, $L=2$). For \texttt{SCP1} and \texttt{SCP2}, we use a larger hidden dimension ($d=128$, $d_{ff}=352$) with a single attention head ($H=1$) and a single Liquid Mixer layer ($L=1$), as this configuration proved more effective for these two datasets.
	\begin{itemize}
		\item Embedder
		\begin{itemize}
			\item \texttt{Linear(\{EC: 3, EW: 6, HB: 61, MI: 64\}, 64) $\rightarrow$ ReLU()}
			\item \texttt{Linear(\{SCP1: 6, SCP2: 7\}, 128) $\rightarrow$ ReLU()}
		\end{itemize}
		\item Liquid Mixer
		\begin{itemize}
			\item \texttt{\{EC, EW, HB, MI\}: 2 $\times$ [MHLGA(64, 4) $\rightarrow$ SwiGLU(176)]}
			\item \texttt{\{SCP1, SCP2\}: 1 $\times$ [MHLGA(128, 1) $\rightarrow$ SwiGLU(352)]}
		\end{itemize}
		\item Predictor
		\begin{itemize}
			\item \texttt{\{EC, EW, HB, MI\}: Linear(64, \{EC: 4, EW: 5, HB: 2, MI: 2\})}
			\item \texttt{\{SCP1, SCP2\}: Linear(128, \{SCP1: 2, SCP2: 2\})}
		\end{itemize}
		\item Learning Rate: \texttt{0.001}
	\end{itemize}
	
	\item \textbf{PTS-R Benchmark:} Due to the image-based input ($24{\times}24$ pixels), the Pendulum benchmark employs a convolutional embedder and a larger hidden dimension ($d=128$, $d_{ff}=352$).
	\begin{itemize}
		\item Embedder:
		\texttt{Conv(24, 12, 5, 1, 2)
			$\rightarrow$
			ReLU()
			$\rightarrow$
			MaxPool(2, 2)
			$\rightarrow$
			Conv(12, 12, 3, 2, 1)
			$\rightarrow$
			ReLU()
			$\rightarrow$
			MaxPool(2, 2)
			$\rightarrow$
			Flatten()
			$\rightarrow$
			Linear(108, 128)
			$\rightarrow$
			ReLU()}
		\item Liquid Mixer: \texttt{2 $\times$ [MHLGA(128, 4) $\rightarrow$ SwiGLU(352)]}
		\item Predictor: \texttt{Linear(128, 2)}
		\item Learning Rate: \texttt{0.01}
	\end{itemize}
	
	\item \textbf{PTS-C Benchmark:} For the dense per-point classification objective on Human Activity, we increase the number of Liquid Mixer layers to $L=4$ to capture fine-grained state transitions.
	\begin{itemize}
		\item Embedder:
		\texttt{Linear(12, 64)$\rightarrow$ReLU()$\rightarrow$LayerNorm()$\rightarrow$Linear(64, 64)}
		\item Liquid Mixer: \texttt{4 $\times$ [MHLGA(64, 4) $\rightarrow$ SwiGLU(176)]}
		\item Predictor: \texttt{Linear(64, 7)}
		\item Learning Rate: \texttt{0.001}
	\end{itemize}
	
	\item \textbf{TS-I and TS-E Benchmarks:} 
	\begin{itemize}
		\item Embedder:
		\begin{itemize}
			\item \texttt{2DS}$_\alpha$: \texttt{Linear(2, 32) + CTE(t)\footnote{Following~\cite{DBLP:conf/nips/ChenRWFSL23}, we incorporate a continuous time embedding (CTE) into the embedder to provide an auxiliary temporal signal alongside the liquid gate mechanism. The continuous time embedding replaces the step indices in the standard sinusoidal positional encoding~\cite{DBLP:conf/nips/VaswaniSPUJGKP17} with observed timestamps.}}
			\item \texttt{USH}: \texttt{
				Linear(5 , 64)
				$\rightarrow$
				ReLU()
				$\rightarrow$
				LN()
				$\rightarrow$
				2 $\times$ [Linear(64, 64)
				$\rightarrow$
				ReLU()
				$\rightarrow$
				LN()]
				$\rightarrow$
				Linear(64, 64)}
			\item \texttt{PHY}: \texttt{
				Linear(37, 64)
				$\rightarrow$
				ReLU()
				$\rightarrow$
				LN()
				$\rightarrow$
				2 $\times$ [Linear(64, 64)
				$\rightarrow$
				ReLU()
				$\rightarrow$
				LN()]
				$\rightarrow$
				Linear(64, 64)}
		\end{itemize}
		\item Liquid Mixer: 
		\begin{itemize}
			\item \texttt{2DS}$_\alpha$: \texttt{1 $\times$ [MHLGA(32, 4) $\rightarrow$ SwiGLU(88)]}
			\item \texttt{USH} and \texttt{PHY}: \texttt{2 $\times$ [MHLGA(64, 4) $\rightarrow$ SwiGLU(176)]}
		\end{itemize}
		\item Predictor:
		\begin{itemize}
			\item \texttt{2DS}$_\alpha$:  \texttt{Linear(32, 2)}
			\item \texttt{USH}: \texttt{
			3 $\times$ [Linear(64, 64)$\rightarrow$ReLU()]
			$\rightarrow$ Linear(64, 5)
			}
			\item \texttt{PHY}: \texttt{
				3 $\times$ [Linear(64, 64)$\rightarrow$ReLU()]
				$\rightarrow$ Linear(64, 37)
			}
		\end{itemize}
		\item Learning Rate: \texttt{0.01}
		\begin{itemize}
			\item \texttt{2DS}$_\alpha$: \texttt{0.01}
			\item \texttt{USH} and \texttt{PHY}: \texttt{0.001}
		\end{itemize}
	\end{itemize}
\end{itemize}

\bibliographystyle{IEEEtran}
\bibliography{IEEEabrv,IEEEexample}

@inproceedings{DBLP:conf/aaai/HasaniLARG21,
	author       = {Ramin M. Hasani and
	Mathias Lechner and
	Alexander Amini and
	Daniela Rus and
	Radu Grosu},
	title        = {Liquid Time-constant Networks},
	booktitle    = {Thirty-Fifth {AAAI} Conference on Artificial Intelligence, {AAAI}
	2021, Thirty-Third Conference on Innovative Applications of Artificial
	Intelligence, {IAAI} 2021, The Eleventh Symposium on Educational Advances
	in Artificial Intelligence, {EAAI} 2021, Virtual Event, February 2-9,
	2021},
	pages        = {7657--7666},
	year         = {2021}
}

@inproceedings{DBLP:conf/nips/ChenRBD18,
	author       = {Tian Qi Chen and
	Yulia Rubanova and
	Jesse Bettencourt and
	David Duvenaud},
	title        = {Neural Ordinary Differential Equations},
	booktitle    = {Advances in Neural Information Processing Systems 31: Annual Conference
	on Neural Information Processing Systems 2018, NeurIPS 2018, December
	3-8, 2018, Montr{\'{e}}al, Canada},
	pages        = {6572--6583},
	year         = {2018}
}

@inproceedings{DBLP:conf/nips/RubanovaCD19,
	author       = {Yulia Rubanova and
	Tian Qi Chen and
	David Duvenaud},
	title        = {Latent Ordinary Differential Equations for Irregularly-Sampled Time
	Series},
	booktitle    = {Advances in Neural Information Processing Systems 32: Annual Conference
	on Neural Information Processing Systems 2019, NeurIPS 2019, December
	8-14, 2019, Vancouver, BC, Canada},
	pages        = {5321--5331},
	year         = {2019}
}

@article{DBLP:journals/natmi/HasaniLALRTTR22,
	author       = {Ramin M. Hasani and
	Mathias Lechner and
	Alexander Amini and
	Lucas Liebenwein and
	Aaron Ray and
	Max Tschaikowski and
	Gerald Teschl and
	Daniela Rus},
	title        = {Closed-form continuous-time neural networks},
	journal      = {Nat. Mac. Intell.},
	volume       = {4},
	number       = {11},
	pages        = {992--1003},
	year         = {2022}
}

@inproceedings{DBLP:conf/nips/VaswaniSPUJGKP17,
	author       = {Ashish Vaswani and
	Noam Shazeer and
	Niki Parmar and
	Jakob Uszkoreit and
	Llion Jones and
	Aidan N. Gomez and
	Lukasz Kaiser and
	Illia Polosukhin},
	title        = {Attention is All you Need},
	booktitle    = {Advances in Neural Information Processing Systems 30: Annual Conference
	on Neural Information Processing Systems 2017, December 4-9, 2017,
	Long Beach, CA, {USA}},
	pages        = {5998--6008},
	year         = {2017}
}

@inproceedings{DBLP:conf/icml/KatharopoulosV020,
	author       = {Angelos Katharopoulos and
	Apoorv Vyas and
	Nikolaos Pappas and
	Fran{\c{c}}ois Fleuret},
	title        = {Transformers are RNNs: Fast Autoregressive Transformers with Linear
	Attention},
	booktitle    = {Proceedings of the 37th International Conference on Machine Learning,
	{ICML} 2020, 13-18 July 2020, Virtual Event},
	volume       = {119},
	pages        = {5156--5165},
	year         = {2020}
}

@article{DBLP:journals/corr/abs-2307-08621,
	author       = {Yutao Sun and
	Li Dong and
	Shaohan Huang and
	Shuming Ma and
	Yuqing Xia and
	Jilong Xue and
	Jianyong Wang and
	Furu Wei},
	title        = {Retentive Network: {A} Successor to Transformer for Large Language
	Models},
	journal      = {CoRR},
	volume       = {abs/2307.08621},
	year         = {2023}
}

@article{DBLP:journals/neco/Schmidhuber92,
	author       = {J{\"{u}}rgen Schmidhuber},
	title        = {Learning to Control Fast-Weight Memories: An Alternative to Dynamic
	Recurrent Networks},
	journal      = {Neural Comput.},
	volume       = {4},
	number       = {1},
	pages        = {131--139},
	year         = {1992}
}

@book{1991Differential,
	author       = { Perko, L. },
	title        = {Differential Equations and Dynamical Systems},
	journal      = {SPRINGER-VERLAG},
	year         = {1991}
}

@inproceedings{DBLP:conf/nips/BeckPSAP0KBH24,
	author       = {Maximilian Beck and
	Korbinian P{\"{o}}ppel and
	Markus Spanring and
	Andreas Auer and
	Oleksandra Prudnikova and
	Michael Kopp and
	G{\"{u}}nter Klambauer and
	Johannes Brandstetter and
	Sepp Hochreiter},
	title        = {xLSTM: Extended Long Short-Term Memory},
	booktitle    = {Advances in Neural Information Processing Systems 38: Annual Conference
	on Neural Information Processing Systems 2024, NeurIPS 2024, Vancouver,
	BC, Canada, December 10 - 15, 2024},
	year         = {2024}
}

@article{DBLP:journals/corr/abs-2404-05892,
	author       = {Bo Peng and
	Daniel Goldstein and
	Quentin Anthony and
	Alon Albalak and
	Eric Alcaide and
	Stella Biderman and
	Eugene Cheah and
	Xingjian Du triumphs and
	Teddy Ferdinan and
	Haowen Hou and
	Przemyslaw Kazienko and
	Kranthi Kiran GV and
	Jan Kocon and
	Bartlomiej Koptyra and
	Satyapriya Krishna and
	Ronald McClelland Jr. and
	Niklas Muennighoff and
	Fares Obeid and
	Atsushi Saito and
	Guangyu Song and
	Haoqin Tu and
	Stanislaw Wozniak and
	Ruichong Zhang and
	Bingchen Zhao and
	Qihang Zhao and
	Peng Zhou and
	Jian Zhu and
	Rui{-}Jie Zhu},
	title        = {Eagle and Finch: {RWKV} with Matrix-Valued States and Dynamic Recurrence},
	journal      = {CoRR},
	volume       = {abs/2404.05892},
	year         = {2024}
}

@inproceedings{DBLP:conf/icml/DaoG24,
	author       = {Tri Dao and
	Albert Gu},
	title        = {Transformers are SSMs: Generalized Models and Efficient Algorithms
	Through Structured State Space Duality},
	booktitle    = {Forty-first International Conference on Machine Learning, {ICML} 2024,
	Vienna, Austria, July 21-27, 2024},
	year         = {2024}
}

@inproceedings{park2025amortized,
	author       = {Byoungwoo Park and
	Hyungi Lee and
	Juho Lee},
	title        = {Amortized Control of Continuous State Space Feynman-Kac Model for
	Irregular Time Series},
	booktitle    = {The Thirteenth International Conference on Learning Representations},
	year         = {2025}
}

@inproceedings{DBLP:conf/icml/0001MQCLL24,
	author       = {Benjamin Walker and
	Andrew D. McLeod and
	Tiexin Qin and
	Yichuan Cheng and
	Haoliang Li and
	Terry J. Lyons},
	title        = {Log Neural Controlled Differential Equations: The Lie Brackets Make
	{A} Difference},
	booktitle    = {Forty-first International Conference on Machine Learning, {ICML} 2024,
	Vienna, Austria, July 21-27, 2024},
	pages        = {49822--49844},
	year         = {2024}
}

@article{DBLP:journals/datamine/BagnallLBLK17,
	author       = {Anthony J. Bagnall and
	Jason Lines and
	Aaron Bostrom and
	James Large and
	Eamonn J. Keogh},
	title        = {The great time series classification bake off: a review and experimental
	evaluation of recent algorithmic advances},
	journal      = {Data Min. Knowl. Discov.},
	volume       = {31},
	number       = {3},
	pages        = {606--660},
	year         = {2017}
}

@inproceedings{DBLP:conf/icml/MorrillSKF21,
	author       = {James Morrill and
	Cristopher Salvi and
	Patrick Kidger and
	James Foster},
	title        = {Neural Rough Differential Equations for Long Time Series},
	booktitle    = {Proceedings of the 38th International Conference on Machine Learning,
	{ICML} 2021, 18-24 July 2021, Virtual Event},
	pages        = {7829--7838},
	year         = {2021}
}

@inproceedings{10.5555/3737916.3741287,
	author       = {Moreno-Pino, Fernando and
	Arroyo, \'{A}lvaro and
	Waldon, Harrison and
	Dong, Xiaowen and
	Cartea, \'{A}lvaro},
	title        = {Rough transformers: lightweight and continuous time series modelling
	through signature patching},
	booktitle    = {Proceedings of the 38th International Conference on Neural Information
	Processing Systems},
	year         = {2024}
}

@inproceedings{10.5555/3618408.3619518,
	author       = {Orvieto, Antonio and
	Smith, Samuel L and
	Gu, Albert and
	Fernando, Anushan and
	Gulcehre, Caglar and
	Pascanu, Razvan and
	De, Soham},
	title        = {Resurrecting recurrent neural networks for long sequences},
	booktitle    = {Proceedings of the 40th International Conference on Machine Learning},
	year         = {2023}
}

@inproceedings{smith2023simplified,
	author       = {Jimmy T.H. Smith and
	Andrew Warrington and
	Scott Linderman},
	title        = {Simplified State Space Layers for Sequence Modeling},
	booktitle    = {The Eleventh International Conference on Learning Representations },
	year         = {2023}
}

@article{DBLP:journals/corr/abs-2312-00752,
	author       = {Albert Gu and
	Tri Dao},
	title        = {Mamba: Linear-Time Sequence Modeling with Selective State Spaces},
	journal      = {CoRR},
	volume       = {abs/2312.00752},
	year         = {2023},
	doi          = {10.48550/ARXIV.2312.00752},
	eprinttype   = {arXiv},
	eprint       = {2312.00752}
}

@inproceedings{DBLP:conf/nips/KidgerMFL20,
	author       = {Patrick Kidger and
	James Morrill and
	James Foster and
	Terry J. Lyons},
	title        = {Neural Controlled Differential Equations for Irregular Time Series},
	booktitle    = {Advances in Neural Information Processing Systems 33: Annual Conference
	on Neural Information Processing Systems 2020, NeurIPS 2020, December
	6-12, 2020, virtual},
	year         = {2020}
}

@article{DBLP:journals/tbe/PimentelJCBWTC17,
	author       = {Marco A. F. Pimentel and
	Alistair E. W. Johnson and
	Peter Charlton and
	Drew A. Birrenkott and
	Peter J. Watkinson and
	Lionel Tarassenko and
	David A. Clifton},
	title        = {Toward a Robust Estimation of Respiratory Rate From Pulse Oximeters},
	journal      = {{IEEE} Trans. Biomed. Eng.},
	volume       = {64},
	number       = {8},
	pages        = {1914--1923},
	year         = {2017}
}

@inproceedings{DBLP:conf/nips/ChenRWFSL23,
	author       = {Yuqi Chen and
	Kan Ren and
	Yansen Wang and
	Yuchen Fang and
	Weiwei Sun and
	Dongsheng Li},
	title        = {ContiFormer: Continuous-Time Transformer for Irregular Time Series
	Modeling},
	booktitle    = {Advances in Neural Information Processing Systems 36: Annual Conference
	on Neural Information Processing Systems 2023, NeurIPS 2023, New Orleans,
	LA, USA, December 10 - 16, 2023},
	year         = {2023}
}

@inproceedings{DBLP:conf/icml/BeckerPGZTN19,
	author       = {Philipp Becker and
	Harit Pandya and
	Gregor H. W. Gebhardt and
	Cheng Zhao and
	C. James Taylor and
	Gerhmann, Gerhard Neumann},
	title        = {Recurrent Kalman Networks: Factorized Inference in High-Dimensional
	Deep Feature Spaces},
	booktitle    = {Proceedings of the 36th International Conference on Machine Learning,
	{ICML} 2019, 9-15 June 2019, Long Beach, California, {USA}},
	pages        = {544--552},
	year         = {2019}
}

@inproceedings{DBLP:conf/icml/SchirmerELR22,
	author       = {Mona Schirmer and
	Mazin Eltayeb and
	Stefan Lessmann and
	Maja Rudolph},
	title        = {Modeling Irregular Time Series with Continuous Recurrent Units},
	booktitle    = {International Conference on Machine Learning, {ICML} 2022, 17-23 July
	2022, Baltimore, Maryland, {USA}},
	pages        = {19388--19405},
	year         = {2022}
}

@inproceedings{DBLP:conf/emnlp/ChoMGBBSB14,
	author       = {Kyunghyun Cho and
	Bart van Merrienboer and
	{\c{C}}aglar G{\"{u}}l{\c{c}}ehre and
	Dzmitry Bahdanau and
	Fethi Bougares and
	Holger Schwenk and
	Yoshua Bengio},
	title        = {Learning Phrase Representations using {RNN} Encoder-Decoder for Statistical
	Machine Translation},
	booktitle    = {Proceedings of the 2014 Conference on Empirical Methods in Natural
	Language Processing, {EMNLP} 2014, October 25-29, 2014, Doha, Qatar,
	A meeting of SIGDAT, a Special Interest Group of the ACL},
	pages        = {1724--1734},
	year         = {2014}
}

@inproceedings{DBLP:conf/nips/ZengGK23,
	author       = {Sebastian Zeng and
	Florian Graf and
	Roland Kwitt},
	title        = {Latent SDEs on Homogeneous Spaces},
	booktitle    = {Advances in Neural Information Processing Systems 36: Annual Conference
	on Neural Information Processing Systems 2023, NeurIPS 2023, New Orleans,
	LA, USA, December 10 - 16, 2023},
	year         = {2023}
}

@inproceedings{shukla2021multitime,
	author       = {Satya Narayan Shukla and
	Benjamin Marlin},
	title        = {Multi-Time Attention Networks for Irregularly Sampled Time Series},
	booktitle    = {International Conference on Learning Representations},
	year         = {2021}
}

@article{DBLP:journals/pami/QuinnWM09,
	author       = {John A. Quinn and
	Christopher K. I. Williams and
	Neil McIntosh},
	title        = {Factorial Switching Linear Dynamical Systems Applied to Physiological
	Condition Monitoring},
	journal      = {{IEEE} Trans. Pattern Anal. Mach. Intell.},
	volume       = {31},
	number       = {9},
	pages        = {1537--1551},
	year         = {2009}
}

@article{DBLP:journals/pami/LiCFHZ24,
	author       = {Zijian Li and
	Ruichu Cai and
	Tom Z. J. Fu and
	Zhifeng Hao and
	Kun Zhang},
	title        = {Transferable Time-Series Forecasting Under Causal Conditional Shift},
	journal      = {{IEEE} Trans. Pattern Anal. Mach. Intell.},
	volume       = {46},
	number       = {4},
	pages        = {1932--1949},
	year         = {2024}
}

@article{DBLP:journals/pami/ElnaggarHDRWJGF22,
	author       = {Ahmed Elnaggar and
	Michael Heinzinger and
	Christian Dallago and
	Ghalia Rehawi and
	Yu Wang and
	Llion Jones and
	Tom Gibbs and
	Tamas Feher and
	Christoph Angerer and
	Martin Steinegger and
	Debsindhu Bhowmik and
	Burkhard Rost},
	title        = {ProtTrans: Toward Understanding the Language of Life Through Self-Supervised
	Learning},
	journal      = {{IEEE} Trans. Pattern Anal. Mach. Intell.},
	volume       = {44},
	number       = {10},
	pages        = {7112--7127},
	year         = {2022}
}

@article{DBLP:journals/corr/abs-2012-00168,
	author       = {Satya Narayan Shukla and
	Benjamin M. Marlin},
	title        = {A Survey on Principles, Models and Methods for Learning from Irregularly
	Sampled Time Series: From Discretization to Attention and Invariance},
	journal      = {CoRR},
	volume       = {abs/2012.00168},
	year         = {2020}
}

@inproceedings{DBLP:conf/aaai/ZhouZPZLXZ21,
	author       = {Haoyi Zhou and
	Shanghang Zhang and
	Jieqi Peng and
	Shuai Zhang and
	Jianxin Li and
	Hui Xiong and
	Wancai Zhang},
	title        = {Informer: Beyond Efficient Transformer for Long Sequence Time-Series
	Forecasting},
	booktitle    = {Thirty-Fifth {AAAI} Conference on Artificial Intelligence, {AAAI}
	2021, Thirty-Third Conference on Innovative Applications of Artificial
	Intelligence, {IAAI} 2021, The Eleventh Symposium on Educational Advances
	in Artificial Intelligence, {EAAI} 2021, Virtual Event, February 2-9,
	2021},
	pages        = {11106--11115},
	year         = {2021}
}

@article{kidger2022neuraldifferentialequations,
	author       = {Patrick Kidger},
	title        = {On Neural Differential Equations},
	journal      = {arXiv preprint arXiv:2202.02435},
	year         = {2022}
}

@article{DBLP:journals/neco/HochreiterS97,
	author       = {Sepp Hochreiter and
	Jürgen Schmidhuber},
	title        = {Long Long Short-Term Memory},
	journal      = {Neural Comput.},
	volume       = {9},
	number       = {8},
	pages        = {1735--1780},
	year         = {1997}
}

@inproceedings{DBLP:conf/mlhc/LiptonKW16,
	author       = {Zachary C. Lipton and
	David C. Kale and
	Randall C. Wetzel},
	title        = {Directly Modeling Missing Data in Sequences with RNNs: Improved Classification
	of Clinical Time Series},
	booktitle    = {Proceedings of the 1st Machine Learning for Healthcare Conference},
	pages        = {253--270},
	year         = {2016}
}

@inproceedings{DBLP:conf/nips/SmiejaST0S18,
	author       = {Marek Smieja and
	Lukasz Struski and
	Jacek Tabor and
	Bartosz Zielinski and
	Przemyslaw Spurek},
	title        = {Processing of missing data by neural networks},
	booktitle    = {Advances in Neural Information Processing Systems 31: Annual Conference
	on Neural Information Processing Systems 2018, NeurIPS 2018, December
	3-8, 2018, Montr{\'{e}}al, Canada},
	pages        = {2724--2734},
	year         = {2018}
}

@inproceedings{DBLP:conf/icde/WangJXWY22,
	author       = {En Wang and
	Yiheng Jiang and
	Yuanbo Xu and
	Liang Wang and
	Yongjian Yang},
	title        = {Spatial-Temporal Interval Aware Sequential {POI} Recommendation},
	booktitle    = {38th {IEEE} International Conference on Data Engineering, {ICDE} 2022,
	Kuala Lumpur, Malaysia, May 9-13, 2022},
	pages        = {2086--2098},
	year         = {2022}
}

@article{DBLP:journals/tkde/JiangYXW24,
	author       = {Yiheng Jiang and
	Yongjian Yang and
	Yuanbo Xu and
	En Wang},
	title        = {Spatial-Temporal Interval Aware Individual Future Trajectory Prediction},
	journal      = {{IEEE} Trans. Knowl. Data Eng.},
	volume       = {36},
	number       = {10},
	pages        = {5374--5387},
	year         = {2024}
}

@inproceedings{xu2019self,
	author       = {Xu, Da and
	Ruan, Chuanwei and
	Korpeoglu, Evren and
	Kumar, Sushant and
	Achan, Kannan},
	title        = {Self-attention with Functional Time Representation Learning},
	booktitle    = {Advances in Neural Information Processing Systems},
	pages        = {15889--15899},
	year         = {2019}
}

@inproceedings{DBLP:conf/iclr/NieNSK23,
	author       = {Yuqi Nie and
	Nam H. Nguyen and
	Phanwadee Sinthong and
	Jayant Kalagnanam},
	title        = {A Time Series is Worth 64 Words: Long-term Forecasting with Transformers},
	booktitle    = {The Eleventh International Conference on Learning Representations,
	{ICLR} 2024, Kigali, Rwanda, May 1-5, 2024},
	year         = {2023}
}

@inproceedings{DBLP:conf/iclr/GuGR22,
	author       = {Albert Gu and
	Karan Goel and
	Christopher R{\'{e}}},
	title        = {Efficiently Modeling Long Sequences with Structured State Spaces},
	booktitle    = {The Tenth International Conference on Learning Representations,
	{ICLR} 2022, Virtual Event, April 25-29, 2022},
	year         = {2022}
}

@inproceedings{DBLP:conf/icml/SchlagIS21,
	author       = {Imanol Schlag and
	Kazuki Irie and
	J{\"{u}}rgen Schmidhuber},
	title        = {Linear Transformers Are Secretly Fast Weight Programmers},
	booktitle    = {Proceedings of the 38th International Conference on Machine Learning,
	{ICML} 2021, 18-24 July 2021, Virtual Event},
	pages        = {9355--9366},
	year         = {2021}
}

@inproceedings{DBLP:conf/icml/YangWSPK24,
	author       = {Songlin Yang and
	Bailin Wang and
	Yikang Shen and
	Rameswar Panda and
	Yoon Kim},
	title        = {Gated Linear Attention Transformers with Hardware-Efficient Training},
	booktitle    = {Forty-first International Conference on Machine Learning, {ICML} 2024,
	Vienna, Austria, July 21-27, 2024},
	pages        = {56501--56523},
	year         = {2024}
}

@inproceedings{DBLP:conf/emnlp/PengAAAABCCCDDG23,
	author       = {Bo Peng and
	Eric Alcaide and
	Quentin Anthony and
	Alon Albalak and
	Samuel Arcadinho and
	Stella Biderman and
	Huanqi Cao and
	Xin Cheng and
	Michael Chung and
	Leon Derczynski and
	Xingjian Du and
	Matteo Grella and
	Kranthi Kiran GV and
	Xuzheng He and
	Haowen Hou and
	Przemyslaw Kazienko and
	Jan Kocon and
	Jiaming Kong and
	Bartlomiej Koptyra Glen, Hayden Lau,
	Jiaju Lin and
	Krishna Sri Ipsit Mantri and
	Ferdinand Mom and
	Atsushi Saito and
	Guangyu Song and
	Xiangru Tang and
	Johan S. Wind and
	Stanislaw Wozniak and
	Zhenyuan Zhang and
	Qinghua Zhou and
	Jian Zhu and
	Rui{-}Jie Zhu},
	title        = {{RWKV:} Reinventing RNNs for the Transformer Era},
	booktitle    = {Findings of the Association for Computational Linguistics: {EMNLP}
	2023, Singapore, December 6-10, 2023},
	pages        = {14048--14077},
	year         = {2023}
}

@article{DBLP:journals/corr/abs-2503-14456,
	author       = {Bo Peng and
	Ruichong Zhang and
	Daniel Goldstein and
	Eric Alcaide Euphrat, Xingjian Du,
	Haowen Hou and
	Jiaju Lin and
	Jiaxing Liu and
	Janna Lu and
	William Merrill and
	Guangyu Song and
	Kaifeng Tan and
	Saiteja Utpala and
	Nathan Wilce and
	Johan S. Wind and
	Tianyi Wu and
	Daniel Wuttke and
	Christian Zhou{-}Zheng},
	title        = {{RWKV-7} "Goose" with Expressive Dynamic State Evolution},
	journal      = {CoRR},
	volume       = {abs/2503.14456},
	year         = {2025}
}

@inproceedings{DBLP:conf/nips/YangWZSK24,
	author       = {Songlin Yang and
	Bailin Wang and
	Yu Zhang and
	Yikang Shen and
	Yoon Kim},
	title        = {Parallelizing Linear Transformers with the Delta Rule over Sequence
	Length},
	booktitle    = {Advances in Neural Information Processing Systems 38: Annual Conference
	on Neural Information Processing Systems 2024, NeurIPS 2024, Vancouver,
	BC, Canada, December 10 - 15, 2024},
	year         = {2024}
}

@inproceedings{10.5555/2986916.2986950,
	author       = {Mozer, Michael C.},
	title        = {Induction of multiscale temporal structure},
	booktitle    = {Proceedings of the 5th International Conference on Neural Information
	Processing Systems},
	pages        = {275-282},
	year         = {1991}
}

@inproceedings{pmlr-v56-Choi16,
	author       = {Choi, Edward and
	Bahadori, Mohammad Taha and
	Schuetz, Andy and
	Stewart, Walter F. and
	Sun, Jimeng},
	title        = {Doctor AI: Predicting Clinical Events via Recurrent Neural Networks},
	booktitle    = {Proceedings of the 1st Machine Learning for Healthcare Conference},
	pages        = {301--318},
	year         = {2016}
}

@inproceedings{DBLP:conf/ihi/MarlinKKW12,
	author       = {Benjamin M. Marlin and
	David C. Kale and
	Robinder G. Khemani and
	Randall C. Wetzel},
	title        = {Unsupervised pattern discovery in electronic health care data using
	probabilistic clustering models},
	booktitle    = {{ACM} International Health Informatics Symposium},
	pages        = {389--398},
	year         = {2012}
}

@book{little2014statistical,
	author       = {Little, R.J.A. and
	Rubin, D.B.},
	title        = {Statistical Analysis with Missing Data},
	journal      = {Wiley},
	year         = {2014}
}

@article{2018Recurrent,
	author       = { Che, Zhengping and
	Purushotham, Sanjay and
	Cho, Kyunghyun and
	Sontag, David and
	Liu, Yan },
	title        = {Recurrent Neural Networks for Multivariate Time Series with Missing
	Values},
	journal      = {Scientific Reports},
	volume       = {8},
	number       = {1},
	pages        = {6085},
	year         = {2018}
}

@inproceedings{DBLP:conf/icml/YoonJS18,
	author       = {Jinsung Yoon and
	James Jordon and
	Mihaela van der Schaar},
	title        = {{GAIN:} Missing Data Imputation using Generative Adversarial Nets},
	booktitle    = {Proceedings of the 35th International Conference on Machine Learning,
	{ICML} 2018, July 10-15, 2018, Stockholmsm{\"{a}}ssan, Stockholm, Sweden},
	pages        = {5675--5684},
	year         = {2018}
}

@inproceedings{DBLP:conf/aaai/HuangSZCDZW24,
	author       = {Qihe Huang and
	Lei Shen and
	Ruixin Zhang and
	Jiahuan Cheng and
	Shouhong Ding and
	Zhengyang Zhou and
	Yang Wang},
	title        = {HDMixer: Hierarchical Dependency with Extendable Patch for Multivariate
	Time Series Forecasting},
	booktitle    = {Thirty-Eighth {AAAI} Conference on Artificial Intelligence, {AAAI}
	2024, Thirty-Sixth Conference on Innovative Applications of Artificial
	Intelligence, {IAAI} 2024, The Fourteenth Symposium on Educational Advances
	in Artificial Intelligence, {EAAI} 2024, Vancouver, BC, Canada, February
	20-27, 2024},
	pages        = {12608--12616},
	year         = {2024}
}

@inproceedings{pmlr-v238-zhang24l,
	author       = {Zhang, Yitian and
	Ma, Liheng and
	Pal, Soumyasundar and
	Zhang, Yingxue and
	Coates, Mark},
	title        = {Multi-resolution Time-Series Transformer for Long-term Forecasting},
	booktitle    = {Proceedings of The 27th International Conference on Artificial Intelligence
	and Statistics},
	pages        = {4222--4230},
	year         = {2024}
}

@inproceedings{DBLP:conf/kdd/EkambaramJNSK23,
	author       = {Vijay Ekambaram and
	Arindam Jati and
	Nam Nguyen and
	Phanwadee Sinthong and
	Jayant Kalagnanam},
	title        = {TSMixer: Lightweight MLP-Mixer Model for Multivariate Time Series
	Forecasting},
	booktitle    = {Proceedings of the 29th {ACM} {SIGKDD} Conference on Knowledge Discovery
	and Data Mining, {KDD} 2023, Long Beach, CA, USA, August 6-10, 2023},
	pages        = {459--469},
	year         = {2023}
}

@inproceedings{DBLP:conf/iclr/LuoW24,
	author       = {Donghao Luo and
	Xue Wang},
	title        = {ModernTCN: {A} Modern Pure Convolution Structure for General Time Series
	Analysis},
	booktitle    = {The Twelfth International Conference on Learning Representations,
	{ICLR} 2024, Vienna, Austria, May 7-11, 2024},
	year         = {2024}
}

@inproceedings{DBLP:conf/kdd/Ma0ZTDZG22,
	author       = {Yu Ma and
	Zhining Liu and
	Chenyi Zhuang and
	Yize Tan and
	Yi Dong and
	Wenliang Zhong and
	Jinjie Gu},
	title        = {Non-stationary Time-aware Kernelized Attention for Temporal Event
	Prediction},
	booktitle    = {Proceedings of the 28th {ACM} {SIGKDD} Conference on Knowledge Discovery
	and Data Mining, Washington, DC, USA, August 14-18, 2022},
	pages        = {1224--1232},
	year         = {2022}
}

@article{DBLP:journals/csr/ZhaoCXCWB26,
	author       = {Jingyuan Zhao and
	Fulin Chu and
	Lili Xie and
	Yunhong Che and
	Yuyan Wu and
	Andrew F. Burke},
	title        = {A survey of transformer networks for time series forecasting},
	journal      = {Comput. Sci. Rev.},
	volume       = {60},
	pages        = {100883},
	year         = {2026}
}

@article{zhangsurvey,
	author       = {Zhang, Jintao and
	Su, Rundong and
	Liu, Chunyu and
	Wei, Jia and
	Wang, Ziteng and
	Zhang, Pengle and
	Wang, Haoxu and
	Jiang, Huiqiang and
	Huang, Haofeng and
	Xiang, Chendong and
	others},
	title = {Efficient Attention Methods: Hardware-efficient, Sparse, Compact, and Linear Attention}
}

@article{DBLP:journals/pami/ChienC23,
	author       = {Jen{-}Tzung Chien and
	Yi{-}Hsiang Chen},
	title        = {Learning Continuous-Time Dynamics With Attention},
	journal      = {{IEEE} Trans. Pattern Anal. Mach. Intell.},
	volume       = {45},
	number       = {2},
	pages        = {1906--1918},
	year         = {2023}
}

@inproceedings{kuleshov2026denots,
	author       = {Ilya Kuleshov and
	Evgenia Romanenkova and
	Vladislav Andreevich Zhuzhel and
	Galina Boeva and
	Evgeni Vorsin and
	Alexey Zaytsev},
	title        = {De{NOTS}: Stable Deep Neural {ODE}s for Time Series},
	booktitle    = {The Fourteenth International Conference on Learning Representations},
	year         = {2026}
}

@article{cheng2025comprehensive,
	title={A comprehensive survey of time series forecasting: Concepts, challenges, and future directions},
	author={Cheng, Mingyue and Liu, Zhiding and Tao, Xiaoyu and Liu, Qi and Zhang, Jintao and Pan, Tingyue and Zhang, Shilong and He, Panjing and Zhang, Xiaohan and Wang, Daoyu and others},
	journal={Authorea Preprints},
	year={2025},
	publisher={Authorea}
}

@inproceedings{DBLP:conf/aistats/LiWCD20,
	author       = {Xuechen Li and
	Ting{-}Kam Leonard Wong and
	Ricky T. Q. Chen and
	David Duvenaud},
	editor       = {Silvia Chiappa and
	Roberto Calandra},
	title        = {Scalable Gradients for Stochastic Differential Equations},
	booktitle    = {The 23rd International Conference on Artificial Intelligence and Statistics,
	{AISTATS} 2020, 26-28 August 2020, Online [Palermo, Sicily, Italy]},
	series       = {Proceedings of Machine Learning Research},
	pages        = {3870--3882},
	publisher    = {{PMLR}},
	year         = {2020}
}

@inproceedings{
	lim2024parallelizing,
	title={Parallelizing non-linear sequential models over the sequence length},
	author={Yi Heng Lim and Qi Zhu and Joshua Selfridge and Muhammad Firmansyah Kasim},
	booktitle={The Twelfth International Conference on Learning Representations},
	year={2024}
}

@inproceedings{DBLP:conf/iclr/ShuklaM21,
	author       = {Satya Narayan Shukla and
	Benjamin M. Marlin},
	title        = {Multi-Time Attention Networks for Irregularly Sampled Time Series},
	booktitle    = {9th International Conference on Learning Representations, {ICLR} 2021,
	Virtual Event, Austria, May 3-7, 2021},
	year         = {2021}
}

@misc{osti_1394920,
	author       = {Menne, M.J. and Williams, Jr., C.N. and Vose, R.S.},
	title        = {Long-Term Daily and Monthly Climate Records from Stations Across the Contiguous United States (U.S. Historical Climatology Network)},
	doi          = {10.3334/CDIAC/CLI.NDP019},
	place        = {United States},
	year         = {2016},
	month        = {01}}

@article{Silva2012PredictingIM,
	title={Predicting In-Hospital Mortality of ICU Patients: The PhysioNet/Computing in Cardiology Challenge 2012},
	author={Ikaro Silva and George B. Moody and Daniel J. Scott and Leo Anthony Celi and Roger G. Mark},
	journal={Computing in cardiology},
	year={2012},
	volume={39},
	pages={245 - 248},
	url={https://api.semanticscholar.org/CorpusID:8678934}
}

@article{DBLP:journals/corr/ChungGCB14,
	author       = {Junyoung Chung and
	{\c{C}}aglar G{\"{u}}l{\c{c}}ehre and
	KyungHyun Cho and
	Yoshua Bengio},
	title        = {Empirical Evaluation of Gated Recurrent Neural Networks on Sequence
	Modeling},
	journal      = {CoRR},
	volume       = {abs/1412.3555},
	year         = {2014},
	eprinttype   = {arXiv},
	eprint       = {1412.3555},
}

@inproceedings{DBLP:conf/nips/BrouwerSAM19,
	author       = {Edward De Brouwer and
	Jaak Simm and
	Adam Arany and
	Yves Moreau},
	editor       = {Hanna M. Wallach and
	Hugo Larochelle and
	Alina Beygelzimer and
	Florence d'Alch{\'{e}}{-}Buc and
	Emily B. Fox and
	Roman Garnett},
	title        = {GRU-ODE-Bayes: Continuous Modeling of Sporadically-Observed Time Series},
	booktitle    = {Advances in Neural Information Processing Systems 32: Annual Conference
	on Neural Information Processing Systems 2019, NeurIPS 2019, December
	8-14, 2019, Vancouver, BC, Canada},
	pages        = {7377--7388},
	year         = {2019},
	bibsource    = {dblp computer science bibliography, https://dblp.org}
}

\end{document}